\documentclass[10pt, conference, compsocconf]{IEEEtran}

\usepackage[hyphens]{url}
\usepackage{hyperref}
\usepackage[pdftex]{graphicx}
\graphicspath{{figures/}} 
\usepackage{float} 
\usepackage{subcaption} 
\usepackage{tabularx}
\usepackage{booktabs} 
\usepackage{array} 
\usepackage{xltabular}  
\usepackage[numbers]{natbib}
\usepackage[table]{xcolor} 
\usepackage{amssymb,amsmath}

\begin{document}

\title{A Comprehensive Survey of Natural Language Generation Advances from the Perspective of Digital Deception}

\author{
\IEEEauthorblockN{Keenan Jones, Enes Altuncu, Virginia N. L. Franqueira,
Yichao Wang and Shujun Li}
\IEEEauthorblockA{Institute of Cyber Security for Society (iCSS)
\\School of Computing, University of Kent, Canterbury, UK\\
\{ksj5, ea483, V.Franqueira, yw300,  S.J.Li\}@kent.ac.uk
}}

\maketitle 

\begin{abstract}

In recent years there has been substantial growth in the capabilities of systems designed to generate text that mimics the fluency and coherence of human language. From this, there has been considerable research aimed at examining the potential uses of these natural language generators (NLG) towards a wide number of tasks. The increasing capabilities of powerful text generators to mimic human writing convincingly raises the potential for deception and other forms of dangerous misuse. As these systems improve, and it becomes ever harder to distinguish between human-written and machine-generated text, malicious actors could leverage these powerful NLG systems to a wide variety of ends, including the creation of fake news and misinformation, the generation of fake online product reviews, or via chatbots as means of convincing users to divulge private information. In this paper, we provide an overview of the NLG field via the identification and examination of 119 survey-like papers focused on NLG research. From these identified papers, we outline a proposed high-level taxonomy of the central concepts that constitute NLG, including the methods used to develop generalised NLG systems, the means by which these systems are evaluated, and the popular NLG tasks and subtasks that exist. In turn, we provide an overview and discussion of each of these items with respect to current research and offer an examination of the potential roles of NLG in deception and detection systems to counteract these threats. Moreover, we discuss the broader challenges of NLG, including the risks of bias that are often exhibited by existing text generation systems. This work offers a broad overview of the field of NLG with respect to its potential for misuse, aiming to provide a high-level understanding of this rapidly developing area of research.

\end{abstract}

\begin{IEEEkeywords}
Natural Language Generation, NLG, Digital Deception, Survey, Taxonomy
\end{IEEEkeywords}

\section{Introduction}
\label{sec:intro} 

The rapid growth in our abilities to generate artificial texts, leveraging new capacities for deep learning and powerful pre-trained models, such as OpenAI's GPT-3~\cite{BMRSK2020}, has meant that natural language generation (NLG) has never been more relevant as a tool for real world use.

Given the broad potential that comes with being able to generate convincing texts, NLG thus finds applications in a large variety of text-based tasks. In turn, NLG has been used in a wide range of fields, including chatbot development~\cite{motger2021}, story creation~\cite{AA2021}, and joke telling~\cite{amin2020}.

This breadth of application means that NLG has a powerful capacity for integrating itself into our everyday lives. Already, virtual assistants, like Amazon's Alexa and Apple's Siri, have found their way into the homes of hundreds of millions of users~\cite{kepuska2018}. These devices rely on  NLG-based modules responsible for generating dialogue in response to users' input in order to answer questions or complete tasks~\cite{kepuska2018}. Beyond personal assistants, NLG systems have been used in a variety of other real-world applications. These include the medical field, in which dialogue and Q\&A systems have been leveraged to aid with diagnoses and patient care~\cite{L2019,pavlopoulos2019} and education, in which a wide variety of task-based systems including story-telling tools and Q\&A chatbots have been proposed as a means of aiding student learning via their integration into virtual platforms and e-tutoring solutions~\cite{amin2020,huang2020}.

With these real-world applications, however, comes the potential for deception and misuse~\cite{tommi2019}. Malicious chatbots could masquerade as genuine people in the hope of misleading them, or tricking them into divulging personal information~\cite{ye2020}. NLG systems designed to persuade could be used by unscrupulous political leaders to profile and target vulnerable users with bespoke, automatically generated advertisements aimed at subtly skewing their perceptions of controversial issues~\cite{wang2019}. Moreover, anonymisation methods could be used to hide the authors of hateful or extremist online posts, granting them added protection from site moderators and law enforcement~\cite{tommi2019}. With every NLG system comes the potential for abuse, and as these systems improve in their abilities to produce text that passes as human this potential for abuse becomes ever more relevant~\cite{stock2016}. 

Given this broad capacity for deception, we opt to take a wide scope in our approach to fully cover the topic of NLG. This will allow us to present a complete consideration of the myriad ways in which various NLG systems, applied to various tasks, can be adapted for deception.

\subsection{Contributions}

In this survey, we offer an overview of the current research in NLG via a review of 119 literature-review (and other survey-like papers) articles in this field. From these identified papers, we outline a high-level taxonomy of the key concepts that constitute the field of NLG and provide dedicated overviews of each of these concepts. Moving beyond this, we also review surveys examining the manner in which NLG systems can be leveraged for deception and other forms of misuse, offering a discussion of the many forms of deception that NLG can take given the wide range of tasks and applications that it can be applied to. In addition, we also discuss the work that has been conducted to address these risks of deception, including the research that has been undertaken to develop detection systems to identify machine-generated text. Finally, we end by examining the broader challenges facing NLG systems beyond the risks of deception and misuse, highlighting work in the literature examining the dangers that model bias may pose to NLG systems and their intended users, before discussing the solutions that have been proposed to address these challenges. 

\section{Methodology}

To source the articles necessary for our survey, considering the breadth and depth needed to provide a sufficiently useful coverage of NLG, we opted for a venue-driven approach, selecting a number of relevant review and survey-like papers focused on artificial intelligence-based (AI) NLG. This approach allowed for the selection of a focused set of relevant papers whilst still providing a good overview of the field as a whole.

To this end, we looked at research papers published since 2019 at a number of venues known to have published NLG-related research, including those listed in the \href{https://aclanthology.org/}{ACL (Association for Computational Linguistics) Anthology}, all conferences and workshops of \href{https://aclweb.org/aclwiki/SIGGEN}{ACL's Special Interest Group on Natural Language Generation (SIGGEN)}, four additional NLG-related conferences not indexed by the ACL Anthology (\href{https://dblp.org/db/conf/ialp/}{IALP}, \href{https://www.cicling.org/}{CICLing}, \href{https://dblp.org/db/conf/pacling/}{PACLING}, \href{https://link.springer.com/conference/tsd}{TSD}), four major journals related to NLG or for publishing survey papers (\href{https://ieeexplore.ieee.org/xpl/RecentIssue.jsp?punumber=6570655}{\emph{IEEE/ACM Transactions on Audio, Speech, and Language Processing}}, \href{https://dl.acm.org/journal/tallip}{\emph{ACM Transactions on Asian and Low-Resource Language Information Processing}}, \href{https://dl.acm.org/journal/csur}{\emph{ACM Computing Surveys}}, \href{https://ieeexplore.ieee.org/xpl/RecentIssue.jsp?punumber=9739}{\emph{IEEE Communications Tutorials \& Surveys}}, and \href{https://ieeeaccess.ieee.org/}{\emph{IEEE Access}}), arXiv.org, and 14 NLG-related Chinese journals. 

For some sources, we first used a search query to identify all surveys, systematisation of knowledge (SoK) papers, systematic reviews, taxonomies, ontologies and general reviews, and then screened all returned papers manually to identify NLG-related papers. For other sources, we screened all papers published in the time period (2019-2021) to identify NLG-related survey papers. In addition, we also manually inspected \href{https://dblp.org/search/publ/bibtex?q=SoK\%24}{all SoK papers indexed by DBLP}. All initially identified papers were further inspected and encoded for exclusion or inclusion in our survey based on their relevance to NLG and its deceptive uses. The following exclusion criteria were applied: (1) papers published before 2019 were excluded, (2) papers not published in English or Chinese were excluded, (3) Papers that do not generate, or the discuss the generation of, natural language texts in English or Chinese were excluded, (4) papers mainly focusing on recommender systems were excluded. From this, 116 NLG-related survey-like papers were selected for inclusion.

All finally selected papers were used to derive the taxonomy of NLG presented in Section~\ref{sec:taxonomy} and the wider contents of this paper.

\begin{figure*}[!htb]
\centering
\includegraphics[width=\linewidth]{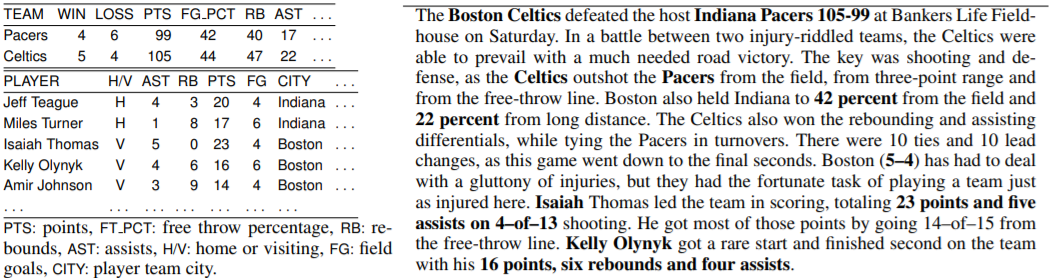}
\caption{An example of data-to-text generation; the left table shows data records provided as input, and the right part shows the corresponding generated text output. Retrieved from \cite{PDL2019}.}
\label{fig:PDL2019}
\end{figure*}

\section{Proposed Taxonomy}
\label{sec:taxonomy}

Based on the NLG-related survey papers identified during the screening process, we derived a proposed taxonomy encompassing the important topics and concepts relevant to NLG. Currently, we have defined the following high-level concepts that compose the first level of our NLG taxonomy:

\textbf{NLG Methods}: This concept encapsulates the methodological approach taken in order to construct systems capable of generalised (non-task specific) language generation. This includes the form of input utilised to generate text, the specific task that the NLG system is directed towards, the training approach taken, and the training data used to build NLG models. We discuss this concept in detail in Section~\ref{sec:TG}.

\textbf{Evaluation of NLG Methods}: This concept refers to the process by which an NLG system is evaluated to measure the quality of its outputs and its performance towards a desired task. This includes the choice of evaluator, the interactivity and internality of the evaluation approach, the metrics used to measure generated text quality, and the methodology that follows from these classes. A discussion of NLG evaluation is presented in Section~\ref{sec:evaluation}.

\textbf{AI Techniques}: AI techniques refers to the form of machine-learning technique(s) or model(s) that are leveraged for NLG. There exist a wide range of AI techniques, including neural network-based (NN) approaches such as recurrent neural networks (RNNs) and their derivatives (e.g., Long short-term memory
(LSTMs) and gated recurrent units
(GRUs)), and pre-trained transformer-based models like GPT-2/3, XLNet, and CTRL that have been applied to NLG. Further discussion of common AI techniques for NLG is presented in Section~\ref{sec:techniques}.

\textbf{Tasks}: This concept encompasses the array of tasks to which NLG systems can be developed towards. As mentioned earlier, NLG can be applied widely, finding uses in the creation of chatbots and other dialogue systems, creative generation (including story and poem creation), as well as being highly valuable in developing language translation and image-captioning systems. An overall discussion of the broad NLG tasks we identify in our survey is presented in Section~\ref{sec:tasks}, with the subsequent sections \ref{sec:Stylised_Generation} -- \ref{sec:Translation} offering more detailed discussions of the various NLG tasks we identify; discussing each task's formulation, the common approaches to designing, implementing, and evaluating NLG models for each task, and the common applications of these task-based NLG systems.

\textbf{Deception \& Detection}: In terms of NLG, attacks can take two forms: attacks on the system and attacks by the system. The former thus refers to malicious attempts to cause undesired or unintended behaviours of a given NLG system or to otherwise avoid any detection systems they may have implemented. Attacks by the system, instead, refers to the usage of NLG systems themselves for malicious purposes, such as generating hate-speech or fake news. Detection encapsulates approaches aimed at identifying AI-generated texts produced by an NLG system -- a crucial task given fears of NLG outputs being disguised as human-created texts. Detection also includes approaches to identify potential attacks against NLG systems or those launched by an NLG system. This is discussed in full in Section~\ref{sec:Deception}.

\textbf{Bias \& Other Challenges}: Finally, this concept covers the future of NLG, examining the issues facing current and future NLG approaches which remain open and questions that need answering to fix these. Beyond issues of NLG-based attacks and detection, this encapsulates other problems with current NLG approaches, such as fears of model bias produced by poorly curated datasets, issues of privacy, and issues with current approaches to generating non-English texts~\cite{bender2021}. We discuss this in Section~\ref{sec:Bias}.

At its core, NLG ``aims to produce plausible and readable text in human language from input data''~\citep{LTZW2021}. Beyond this, there exist a variety of popular NLG tasks (e.g., story generation, question answering) requiring bespoke NLG development strategies that can be adapted for a number of different applications (e.g., creative writing, conversation). In this section, we focus on different methods for the generalised generation of text, including the types of input and the underlying techniques and training approaches used in the development of NLG models. An examination of typical methods of developing systems for specific NLG tasks is then covered in sections \ref{sec:tasks} -- \ref{sec:Translation}.

\section{NLG Methods\label{sec:TG}}

\subsection{NLG Input}

NLG tasks can be \emph{uncontrolled} (also called unconditional) or \emph{controlled} (also called conditional). In the former case, the text is generated without constraints, typically based on a random noise vector~\citep{LTZW2021} or no input.
For the latter case -- controlled NLG tasks -- a variety of inputs can be used to generate different types of text-based outputs according to the following generation paradigms:

\textbf{Text-to-text}: This paradigm includes topic-based and  attribute-based NLG. It processes \emph{unstructured} textual inputs to generate new output text(s). The types of input text include~\citep{LTZW2021, PBS2020, YZLHWJJ2020}: topics, keywords, sentiment labels, stylistic attributes (e.g., politeness, formality), demographic attributes of the ``intended writer'' (e.g., gender, age), information (e.g., event, entity), and text sequencing or ordering (e.g., from paragraphs, grounded documents, webpages).

\textbf{Data-to-Text}: This paradigm processes \emph{structured} data input to generate new output text, retaining as much relevant information as possible from the structured data~\citep{LTZW2021}. It includes non-linguistic data (knowledge-based or table-based data) and can take the form of knowledge graphs, expert system knowledge bases, database of records, spreadsheets, and simulations of physical systems~\citep{LTZW2021, CB2020, PDL2019}. Figure~\ref{fig:PDL2019} shows an example of data input and its corresponding generated output.

\textbf{Multimedia-to-Text}: This paradigm uses \emph{multimedia} data input (e.g., image, video, speech) to generate output text~\citep{LTZW2021}. An example application is image captioning where the input is an image and the output is a corresponding text.


\subsection{Training of NLG Models}

The main goal of training an AI model is to reduce the gap between a desired output and a model selected output, where the gap is defined by a given objective function (otherwise known as a loss or cost function)~\cite{PBS2020}. In the context of NLG, objective functions are typically leveraged to generate more fluent, grammatical and diverse generations by attempting to find the optimum solution (e.g., the optimum next word or character in an NLG context) for a given objective function~\cite{PBS2020}. Whilst a number of loss functions are used in NLG, they generally centre around the comparison between a predicted token selected by the NLG model (given a sequence of tokens as input) and a reference token. By training the NLG model to minimise the loss between the generated output and the reference output, the model can `learn' to output good quality texts.

As most NLG solutions currently rely on deep learning techniques, it is generally necessary to provide large amounts of text data to the NLG model in order to ensure it is adequately trained to perform a given NLG task to a desired level of capability. To train these deep learning models (commonly used techniques include RNNs and LSTMs) for NLG there exist a number of common training paradigms including: \textit{supervised learning}, \textit{reinforcement learning}, and \textit{adversarial learning}~\cite{lu2018}.

\textbf{Supervised Learning}: The most common approaches to NLG utilise (pseudo) supervised learning via maximum likelihood estimation (MLE)~\cite{lu2018}. In turn, large amounts of training text are provided to the model of choice as a series of training patterns (text fragments of a given size, sampled from the training data). The model is then tasked, for each training pattern, to predict the token that follows it~\cite{lu2018}. This can, therefore, be thought of as a classification task, where each class is the equivalent of a unique token in the training dataset and the total of number of classes equals the total number of unique token (words, characters etc.)~\cite{lu2018}. Via this training, the model is thus able to model the probability of each token occurring, given a sequence as input. MLE approaches, whilst popular, are limited by their tendency to overfit to the training data. This \textit{exposure bias} means that MLE-based models are often limited in their ability to generalise beyond the training data~\cite{lu2018}, particularly as they must now generate tokens based on previously generated sequences, rather than sequences existing in the training set. This becomes particularly problematic as the length of the generated sequence increases~\cite{CB2020}. 

\textbf{Reinforcement Learning}: To try and account for the limitations in MLE, reinforcement learning (RL) approaches have been suggested which aim to optimise non-differentiable metrics of text quality. A common approach to this is PG-BLEU. This learning method leverages the popular text evaluation metric BLEU (Bilingual Evaluation Understudy), which measures the n-gram overlap between a given (generated) text and a set of reference texts~\cite{celikyilmaz2020}. In turn, PG-BLEU aims to optimise for BLEU using typical RL policy gradient algorithms like REINFORCE~\cite{CB2020}. Whilst, in theory, this could allow for the generation of more relevant texts than MLE-based approaches, the significant computational cost of computing BLEU so frequently means that this approach is seldom used in practice~\cite{lu2018}. Additionally, criticism regarding the suitability of BLEU (and other, similar NLG evaluation metrics) to measure text quality raise further questions as to the applicability of these RL-based approaches~\cite{celikyilmaz2020}. As such, these approaches are less commonly seen~\cite{lu2018}.

\textbf{Adversarial Learning}: Beyond the above, adversarial approaches to NLG have also been proposed. An early example in this space is the Professor Forcing algorithm, which uses adversarial domain adaption to reduce the distance between the training and generation of an RNN~\cite{CB2020}. This, in turn, aims to limit exposure bias and boost generation quality. Beyond this, GAN-based approaches have also been suggested, using the discriminator's gradient to improve the generator's outputs~\cite{CB2020}. A variety of GAN-based NLG approaches have been suggested, including seqGAN, maskGAN, and LeakGAN -- these are described fully in Section~\ref{sec:techniques}. Whilst GAN-based approaches can be effective, they are inhibited by a number of problems. The most common of these are issues of vanishing gradient, in which the discriminator becomes much stronger than the generator leading to minimal updates being provided~\cite{lu2018}. Issues of mode collapse are also common, in which the generator learns to sample from a small subset of tokens that receive higher evaluations from the discriminator~\cite{CB2020}. This can lead to the GAN learning only a subset of the target distribution, limiting its ability to produce more generalised and diverse texts.

A problem with these typical core approaches to NLG is the necessity for large amounts of training data~\cite{LTZW2021}. Without this, NLG models are prone to overfitting the training dataset and thus fail to generalise adequately to their desired task~\cite{LTZW2021}. Moreover, even when datasets of sufficient size are available, the computational resources required to train these models are often prohibitive, restricting the feasibility of these solutions for many developers~\cite{LTZW2021}. Additionally, these NLG models are largely task-specific in nature, only capable of generating text within the context of the training data provided. This means that developing NLG models to cover the large variety of NLG tasks and contexts requires a wide range of bespoke models, each trained on a significant amount of training data relevant to each task.

To help overcome this \emph{data scarcity}, recent approaches instead leverage massive pre-trained language models (PLM)~\cite{LTZW2021, WYLFBP2021, SQZZH2021}. Rather than training a given model to perform a specific NLG task, these language models (LM), such GPT-2/3~\cite{DCLT2018, radford2019}, are instead pre-trained using a generic unsupervised text prediction task. Examples of these tasks include \textit{Masked LM}, in which a series of sentences are presented to the model with a section of the sentence removed or `masked'~\cite{DCLT2018}. The model must then attempt to predict the missing part of the sentence. Other tasks include next sentence and next word prediction, in which a given input (e.g., a sentence) is provided to the model, and it is tasked with predicting the following word or sentence~\cite{radford2019}. By training these models to successfully conduct these `low-level' prediction tasks, the LM is able to achieve a good `understanding' of how the language in the data provided is used.

Having conducted this pre-training, the model can then be fine-tuned using small amounts of task-specific training data to conduct a given NLG task -- leveraging its understanding of language achieved at the pre-training stage and specifying it using the task-specific data. Beyond requiring smaller amounts of data to conduct a given NLG task, this also allows for high degrees of transference in which the pre-trained understanding of language achieved by the model can be adapted to a variety of NLG (and other NLP) tasks. 

Moreover, the power of the newer PLMs (e.g., GPT-3) has allowed for further approaches capable of leveraging even smaller sets of data, provided at inference time, to guide generation. These approaches include:

\textbf{Few-shot learning}: Few-shot learning relies on providing only a small number of samples to the NLG model at inference time~\cite{LTZW2021}. A subset of this is one-shot learning, in which a single sample is provided~\cite{LTZW2021}. This approach therefore leans heavily on the generalised understanding of language achieved during the PLM's pre-training. Applications of this in NLG include question answering, in which a few examples of similar questions are provided a given model, before it is prompted to answer a new (unseen) question.

\textbf{Zero-shot learning}: Moving beyond few-shot learning, zero-shot learning asks a given NLG model to respond correctly to an unseen prompt with no additional data provided~\cite{BMRSK2020, LTZW2021}. This approach thus relies even more heavily on the model's ability to generalise from its learning during pre-training. Examples of this in NLG include the writing of news articles, using only the headline of the article as a prompt, or the answering of a question with only the question itself as a prompt. This approach (alongside few-shot learning) has only become feasible in recent years, with the rapid increase in the amounts of training data used to build powerful PLMs~\cite{BMRSK2020}.

Whilst few-shot (and zero-shot) learning may be desirable, in practice there is generally a sufficient degree of difference between the pre-training domain and the task domain that the model is unable to generalise effectively from minimal data~\cite{LTZW2021}. In turn, most NLG approaches utilise domain transfer, in which larger amounts of data are used to adapt the model to the desired NLG task. This may involve simply providing larger amounts of fine-tuning data, or potentially providing additional data at the pre-training stage~\cite{LTZW2021}. This approach can allow for greater NLG performances than can be achieved using few-shot learning, whilst often still requiring less training data than other deep learning solutions.

Although the use of powerful LMs pre-trained on massive datasets has allowed for state-of-the-art capabilities in NLG, this approach to training has brought with it a number of limitations.

Currently, most NLG models are notably English-centric, with many of the common pre-trained approaches specifically only utilising English datasets (e.g., GPT-2~\cite{radford2019} and XLNet~\cite{YDYCSL2019}). Moreover, even when efforts are made to include additional languages, such as in GPT-3~\cite{BMRSK2020}, these additional languages are typically under-represented in the training data used~\cite{joshi2020}. This is particularly problematic in regard to the current reliance on web data for pre-training LMs. Whilst this approach is useful for extracting the vast amounts of data required for adequate pre-training, this approach yields clear biases towards a smaller number of over-represented languages~\cite{joshi2020}.

\begin{table*}[!htb]
\centering
\begin{tabular}{ p{0.1\textwidth} p{0.7\textwidth} p{0.1\textwidth}}
 \toprule
 \textbf{Model} & \textbf{Dataset(s)} & \textbf{Size}\\
 \midrule
 BART & BooksCorpus, English Wikipedia, CommonCrawl (filtered), OpenWebText & 160GB\\
 BERT & BooksCorpus, English Wikipedia & 16GB\\
 CTRL & Wikipedia (En, De, Es, Fr), Project Gutenberg, OpenWebText, Amazon Reviews, and several other data sources & 140GB\\
 GPT & BooksCorpus & 5GB\\
 GPT-2 & WebText & 40GB\\
 GPT-3 & Common Crawl (filtered), WebText2, Books1, Books2, Wikipedia & 570GB\\
 MASS & WMT News Crawl (En, De, Fr, Ro) & 41.5GB\\
 T5 & Colossal Clean Crawled Corpus (C4) & 750GB\\
 UniLM & BooksCorpus, English Wikipedia & 16GB\\
 XLNet & BooksCorpus, English Wikipedia, Giga5, ClueWeb 2012-B (filtered), Common Crawl (filtered) & 126GB\\
 \bottomrule
\end{tabular}
\caption{Pre-training datasets for PLMs popularly used in NLG.}
\label{table:pre-training_data}
\end{table*}

In order to rectify this, approaches have been suggested to adapt current NLG methods to better support text generation in non-English languages~\cite{siblini2019}. Some approaches seek to leverage zero-shot learning, combined with large language agnostic datasets. LMs pre-trained on a variety of languages have thus been proposed, e.g., mBERT~\cite{siblini2019}, mT5~\cite{xue2021}, XLM~\cite{lample2019}, XLM-R~\cite{conneau2020}. These have, in turn, shown a reasonable degree of promise in adapting to text generation and other NLP tasks in different languages, either using their inherent language understanding achieved through pre-training or through minimal fine-tuning using texts written in the target language~\cite{siblini2019}. Whilst this approach has achieved a reasonable degree of success, these models are still limited by the lack of language variety within their pre-training datasets~\cite{joshi2020}. Additionally, issues of accidental translation in which the generated output contains the `wrong' language are common~\cite{xue2021}. Moreover, current research in this space has been limited by a focus on multilingual pre-training using genetically related languages. It is currently less clear as to the degree to which generalisation to non-related languages is possible~\cite{joshi2020}.

Other approaches instead take a mono-lingual direction, in which the given LM is pre-trained entirely on text from the desired language (in the same way as English-only models are). This has proven to be the most effective approach, generally exceeding the capabilities of most multilingual models~\cite{bender2021}. However, this method is limited in much the same way as English-only models are, being confined to generation in a single language. Moreover, as sufficient data for many languages is not currently available, reliance on this approach would thus exclude the usage of NLG for many of the world's languages~\cite{bender2021}.

Issues of bias in the current (typically) web-based training data for NLG have been noted beyond the lack of multilingual support~\cite{bender2021}. A major concern is the degree to which the often atypical and extreme nature of web content from certain platforms could have negative effects on the behaviour of NLG models trained using them~\cite{bender2021}. This, in turn, could lead to NLG models that mistakenly internalise biases present in their (pre) training data, which could cause undesirable outputs hostile against certain protected groups, as well as causing potential predispositions towards violent or offensive language and hate-speech~\cite{bender2021}. Whilst many of the filtering processes taken during the data collection phase aim to limit this, the vast amounts of training data currently used for PLM-based NLG means that sufficiently curating datasets to limit these harms and biases is currently an open problem~\cite{bender2021}. Further discussion of the issues of bias facing current NLG appraoches are discussed in Section~\ref{sec:Bias}.

\subsection{Training Datasets for NLG}

As popular neural approaches to NLG require large amounts of high quality text data in order to be effective, there has been a large amount of effort dedicated to curating large datasets for NLG. This has become increasingly important with the rise of powerful PLMs (mentioned above), which require sufficiently large datasets to generalise effectively. A table indicating the datasets used by several of the most popular PLMs can be found in Table~\ref{table:pre-training_data}. In this section we detail some of the training datasets commonly used in NLG.

\textbf{Common Crawl}: The Common Crawl dataset is a massive collection of petabytes of web data currently hosted by Amazon (\url{https://commoncrawl.org/the-data/}). This dataset contains raw web page data, metadata extracts, and text extracts scraped over a period of 12 years. Due to its size, Common Crawl has been popularly used in the training and pre-training of many NLG models, including GPT-2/3~\cite{BMRSK2020,radford2019}, T5~\cite{RSRLNMZLL2020}, and XLNet~\cite{YDYCSL2019}. However, as the dataset is so large and contains within it a great deal of low quality text~\cite{BMRSK2020}, current NLG approaches conduct additional filtering to extract subsets of the data more suited to the NLG or general language modelling task at hand~\cite{BMRSK2020,RSRLNMZLL2020}.

\textbf{WebText}: Developed for use in the pre-training of GPT-2, WebText offers a subset of 8 million web pages from the Common Crawl dataset containing web pages that have been previously filtered or created by humans~\cite{radford2019}. To do this, all outbound links from the web forum site Reddit with 3 or more karma were extracted. This ensured that the data included in WebText had received some degree of favourable response by humans and was likely of reasonable quality. Any Wikipedia articles included in this set of articles were then removed since they are commonly used in other NLG datasets.   

\textbf{OpenWebText}: OpenWebText (\url{https://skylion007.github.io/OpenWebTextCorpus/}) is an open-source recreation of the WebText dataset. It contains 38GB of text data extracted from more than 8 million URLs shared on Reddit with at least three upvotes.

\textbf{Colossal Clean Crawled Corpus (C4)}: Another subset of the Common Crawl Corpus, C4 was created for use in pre-training the T5 LM~\cite{RSRLNMZLL2020}. To create C4, the authors filtered out all non-English texts, whilst also removing short texts, obscene texts, duplicates, and non-natural language text.

\textbf{BooksCorpus}: The BooksCorpus dataset contains a collection of 11,038 unpublished books scraped from the web~\cite{zhu2015}. These books are all at least 20K words in length, and include a variety of genres including fantasy, romance, and science fiction.

\textbf{ClueWeb12}: Created by the Lemur Project, the ClueWeb12 dataset contains more than 7 million web pages scraped between 2010 and 2012 (\url{https://lemurproject.org/clueweb12/}).

\textbf{Giga5}: The English Gigaword Fifth Edition (Giga5) is a dataset of English newswire text data extracted from seven sources including the Los Angeles Times/Washington Post Newswire Service, the Washington Post/Bloomberg Newswire Service, and the Xinhua News Agency, English Service (\url{https://catalog.ldc.upenn.edu/LDC2011T07}). The data collection process was conducted over a number of years, beginning with the first edition in 2003 and ending with the fifth edition in 2010.

\textbf{WMT News Crawl}: WMT News Crawl (\url{http://data.statmt.org/news-crawl/}) dataset contains 1.5 billion lines of monolingual text from 59 languages, extracted from online newspapers. It was released for the Workshop on Statistical Machine Translation (WMT) series of shared tasks.

\textbf{Wikipedia}: Likely due to its size and availability (\url{https://en.wikipedia.org/wiki/Wikipedia:Database_download}), Wikipedia data has been commonly used in NLG~\cite{DCLT2018}. In turn, many powerful models have leveraged subsets of the complete list of articles on Wikipedia, typically filtering it by the languages relevant to the desired language modelling/NLG task~\cite{DCLT2018, KMVXS2019}. This data is typically also filtered to only include the actual text content of the articles themselves, with non-prose-based text such as tables, lists, and headers commonly being excluded~\cite{DCLT2018}.

\section{Evaluation of NLG Methods\label{sec:evaluation}}

Beyond the construction of a given NLG system, a further challenge in generating synthetic texts is presented by the manner in which these AI-generated texts can be successfully evaluated. Owing to the open-ended nature of many NLG tasks, the role of creativity in the text generation process, and the natural ambiguity of language, conducting successful evaluation is an ongoing challenge within the field of NLG~\cite{celikyilmaz2020}. 

Currently, there exist many different approaches to evaluating the outputs of a given text generator~\cite{celikyilmaz2020,hamalainen2021}. In turn, we systematise these approaches using a series of different categories: the evaluator, the level of interactivity used, the internality of evaluation, the measures used for evaluation, and the overall methodological approach taken. Finally, we end by reviewing some of the existing standardised evaluation methods and tools that are currently in use, the best practices suggested in previous research, and the challenges still posed by a lack of standardisation in this area.

\subsection{Evaluators}

Central to the evaluation of NLG systems is the role of the evaluator. That is, the agent responsible for evaluating the given NLG system. In turn, we identify two overarching types of evaluator that are used in NLG evaluation: \textbf{human-based} and automated \textbf{machine-based}.

The human evaluator refers to the use of human agents as the judges of a given NLG system~\cite{celikyilmaz2020}. Human evaluators are typically called upon to perform one of three evaluation tasks: (1) they may be asked to rate a sample of generated outputs from a given NLG model in a stand-alone fashion~\cite{hamalainen2021}, (2) they may be tasked with comparing or ranking outputs from a series of NLG models~\cite{celikyilmaz2020}, or (3) they may be tasked with conducting some form of modified Turing test, using their abilities to distinguish between generated texts and human-created texts~\cite{garbacea2019}. 

Additionally, human evaluators can then be sub-dived into \textit{expert} and \textit{non-expert} evaluators~\cite{van-der-lee2019}. Typically, NLG evaluations are conducted using either a small number of expert evaluators or a larger number of non-expert evaluators, though the precise numbers commonly used vary significantly from study to study -- with typically 1--4 expert evaluators used and anywhere from 10--60 non-expert evaluators being common~\cite{van-der-lee2019}. Whilst most approaches utilise either expert or non-expert evaluators, there is the potential for the usage of both to be of value as research indicates that different insights can be gleaned from human evaluators with varying levels of expertise~\cite{van-der-lee2019}. 

Humans evaluators are commonly used and are generally considered the `gold-standard' for NLG evaluation. This is due to their superior capabilities of language comprehension, their strong abilities towards context-based evaluation (relative to automated evaluators), and the added depths of insight that they can provide~\cite{celikyilmaz2020,howcroft2020,schoch2020}.

However, it is also worth noting that there are limitations in the current usage of human evaluators. Firstly, human evaluation can be time-consuming and inefficient to conduct~\cite{amidei2019,howcroft2020}. Moreover, the use of human evaluators requires some form of recruitment process which may be prohibitively expensive for some NLG projects~\cite{celikyilmaz2020}. 
Additionally, there can be issues of consistency when using human evaluators. As human evaluators must typically rely on their subjective judgement, criticisms of the consistency and replicability of projects evaluated using human evaluators are common~\cite{celikyilmaz2020,schoch2020}.

Parallel to the usage of human evaluators is automated evaluation. In this approach, some form of automated methods are utilised to provide evaluation of the NLG system~\cite{finch2020}. Automated evaluation, in turn, can then be categorised into two further sub-domains: \textit{untrained automatic evaluation} and \textit{machine-learned evaluation}~\cite{celikyilmaz2020}.

Untrained automatic evaluators, the form of automated evaluator most commonly utilised in NLG studies, rely on the use of one, or a series of, objective metrics to evaluate a given NLG system~\cite{finch2020}. These metrics require no pre-training and can be used to provide efficient evaluations of large amounts of generated data. 

However, there are issues regarding the abilities of these automatic approaches to successfully replicate human judgement -- with low correlations often being found between these metrics and human evaluations~\cite{finch2020}. Moreover, automatic approaches are often limited in their ability to provide more holistic measures of automated text quality, which can be particularly limiting in scenarios in which a given NLG system is applied to more creative tasks such as story generation~\cite{celikyilmaz2020}.

A more recent approach to automated evaluation concerns the use of machine learning-based evaluators~\cite{celikyilmaz2020}. Rather than leveraging predetermined measures of evaluating NLG systems, this approach attempts to train machine learning (ML) models to act as pseudo-human evalautors. In turn, this aims to account for the limitations of automatic evalautors in capturing holistic qualities of generated texts and in correlating with human judgement. Currently, however, there has been less research into the creation of these evaluators and they remain relatively underused~\cite{celikyilmaz2020}.

\subsection{Interactivity}

Interactivity refers to the the manner in which the evaluator engages with the generated artefacts they are tasked with evaluating. In turn, this leads to two forms of interactivity: \textbf{static}, and \textbf{interactive}~\cite{finch2020}.

In static evaluation, the evaluator is simply presented with samples of the generated output of a given NLG model~\cite{finch2020}. The evaluator can then proceed to offer judgements on the sample provided. This approach is by far the most common in NLG evaluation and is task agnostic -- meaning that it can be conducted with any form of NLG system directed towards any of the numerous NLG tasks possible -- and can be used for both human and automated evaluation. 

Whilst static evaluation has the advantage of being broadly applicable (often being the only applicable approach to NLG evaluation) it can be limited when evaluating dynamic NLG systems. This is particularly true in the case of chatbots and other dialogue systems, where the offline evaluation of their text outputs in isolation can often lack relevance in terms of evaluating the true capabilities and behaviours of the dynamic NLG model~\cite{finch2020}.

In turn, interactive evaluation is offered as a solution to this. Interactive evaluation is conducted via direct interactions with the NLG system, as if the evaluator was a user~\cite{finch2020}. The evaluator is thus able to directly interact with the NLG system and provide their evaluations in response to the NLG's abilities within its desired role~\cite{finch2020}.

\begin{figure*}[!ht]
\centering
\includegraphics[width=0.92\linewidth]{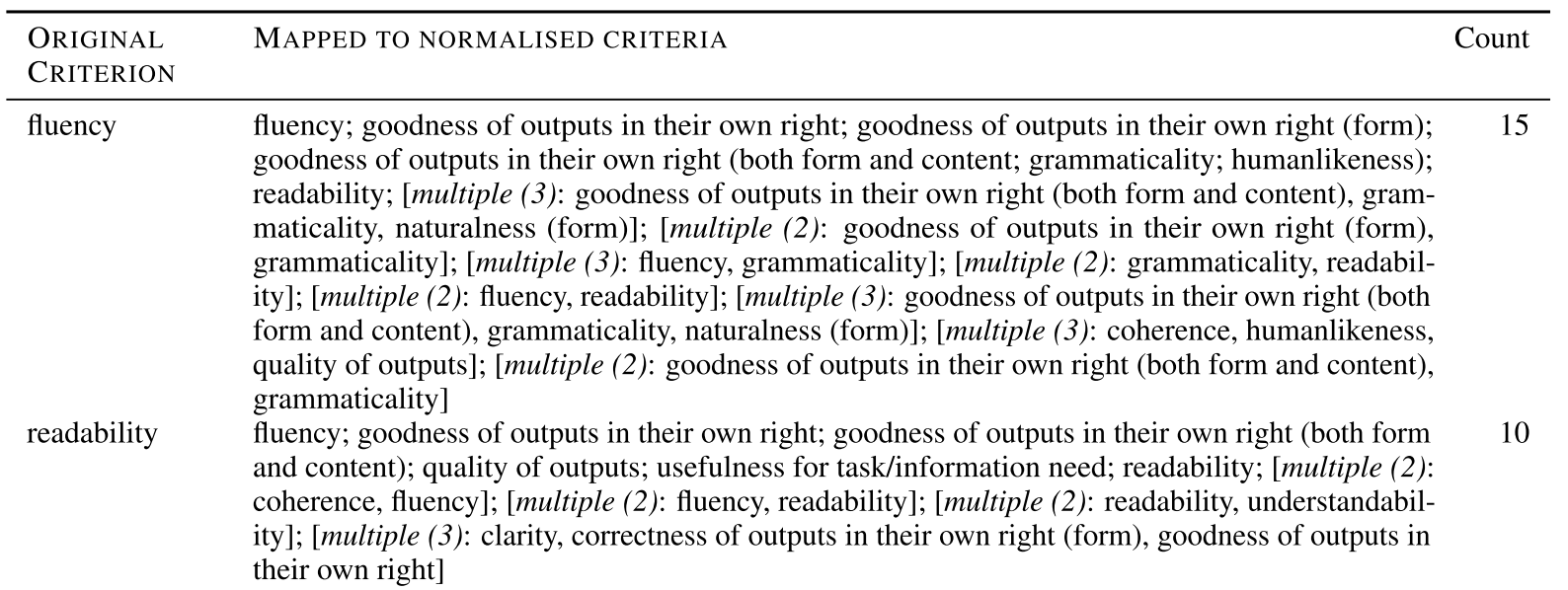}
\caption{Example of the original quality criterion names used in the literature and the variety of normalised criteria these names are actually referring. This highlights the lack of common definitions in the current usage of quality criteria. Retrieved from \cite{howcroft2020}.}
\label{fig:quality_definitions}
\end{figure*}

Although interactive evaluation can provide greater value in understanding the capabilities of a dynamic NLG system it is also more complex to implement, typically requiring more time and evaluator effort than static evaluation. Moreover, the added complexity of interactive evaluation means that the use of automated evaluators is considerably more difficult to implement. This, in turn, typically necessitates the use of human evaluators~\cite{finch2020}.

\subsection{Internality}

Internality refers to the area that the evaluator should emphasise when making judgements of an NLG system. Specifically, internality can take one of two distinct attributes: \textbf{intrinsic} and \textbf{extrinsic}~\cite{belz2020,celikyilmaz2020,van-der-lee2019}.

With an intrinsic approach to NLG evaluation, the goal is to evaluate the proposed NLG system via a direct assessment of its generated outputs~\cite{celikyilmaz2020}. With extrinsic evaluation, the aim is instead to assess the NLG system based on how successfully it achieves an intended goal or downstream task~\cite{celikyilmaz2020}. For instance, an extrinsic evaluation of an NLG system designed to generate advertisements for cars may be conducted by examining how well these adverts increase car sales, whereas an intrinsic evaluation would focus on assessing the content of the advertisements themselves.

In general, intrinsic methods are most commonly used in NLG studies ~\cite{celikyilmaz2020} -- with one review identifying only 3 (3\%) papers that utilise extrinsic evaluation~\cite{van-der-lee2019}. Due to its focus on evaluating the NLG system in a downstream process, extrinsic evaluation often requires a more long-term and costly evaluation process~\cite{celikyilmaz2020}. Additionally, current NLG research is typically focused on smaller subtasks lacking the clear real-world application required to perform extrinsic evaluation~\cite{van-der-lee2019}. However, extrinsic evaluation is typically considered a more relevant and tangible form of evaluation, and as NLG systems become more integrated into real-world applications an increase in this form of evaluation is likely.

\subsection{Evaluation Measures}
\label{label:eval_measures}

Emerging from the choices of the evaluator and its level of interactivity and internality are the measures that are to be used to score the `quality' of a given NLG system or model. These measures, in turn, are generally divided by the evaluator used into measures for human evaluator-based evaluation and measures for machine-based evaluation~\cite{celikyilmaz2020}. We present the most common measures used in NLG evaluation for each of these.

Human-based evaluation measures are typically based on examining the direct scores provided by a series of human evaluators on a set of generated texts~\cite{howcroft2020}. In turn, this can be divided into two parts: the \textit{quality criterion} and the \textit{evaluation mode}~\cite{belz2020,howcroft2020}. These two parts combine to allow a human evaluator to provide their assessment of the relevant aspects of an NLG system's quality. 

Quality criterion describes the aspect of the NLG system's outputs that the human evaluator is attempting to measure~\cite{belz2020,hamalainen2021,howcroft2020}. Human-based evaluation methods will often measure multiple quality criteria, where each criterion relates to a desirable component that should appear in a quality AI-generated output.

There are a large range of quality criteria in current usage that can be sub-divided taxonomically in a variety of ways. In \cite{howcroft2020}, the authors identify the high level quality concepts of \textit{measures of correctness} and \textit{measures of goodness} that define the current quality criterion used in NLG studies. Further sub-classes then ask the evaluator to consider the quality of the text either in its own right, relative to a reference external to the NLG system, or relative to the inputs provided to the NLG system~\cite{howcroft2020}.

A key limitation in current studies is that the quality criteria measured are essentially innumerable and often very different from one study to the next~\cite{belz2020,howcroft2020}. A lack of a common vocabulary for these quality criteria is also a key limiting factor in replicability and comparison, with papers frequently using the same quality criteria name but with varying definitions (as shown in Fig.~\ref{fig:quality_definitions})~\cite{belz2020,briakou2021,hamalainen2021,howcroft2020,schoch2020,van-der-lee2019}. 

Some of the most commonly used quality criterion are:

\textbf{Fluency}: This refers to the degree to which a generated text, or set of generated texts, mimics the intended language it is written in~\cite{hamalainen2021}. A broad criterion, this can include considerations of correct grammar and syntax, spelling, and style. Whilst most commonly used in machine translation, fluency can be easily applied to any NLG task in which fluent writing is desired. This has lead to it becoming one of the most commonly used dimensions in existing studies~\cite{van-der-lee2019}. Additional aspects of fluency, including tone and formality, are also of importance to style transfer tasks~\cite{briakou2021}. Due to its broad nature, however, the lack of precise definition for fluency within the NLG literature is problematic~\cite{belz2020}.
    
\textbf{Usefulness}: Usefulness is a criterion focused around the degree to which the generated text is valuable for a given task or information need~\cite{howcroft2020}.
    
\textbf{Factuality}: Factuality examines the degree to which the generated text is logically coherent, and the degree to which its statements are true~\cite{celikyilmaz2020}. The first aspect of factuality is broadly useful for NLG evaluation, whilst the second aspect is of particular value to tasks such as news generation, where accurate reporting in the generated text is desired.
    
\textbf{Naturalness/Typicality}: Naturalness (also called typicality) asks the evaluator to assess how `typical' a given generated text is, or how often they'd expect to see a text like this~\cite{hamalainen2021,van-der-lee2019}. This is usually considered in terms of how likely a natural speaker would produce the given text, and can be measured in terms of both the content and form of the output~\cite{howcroft2020}.

\textbf{Grammaticality}: This criterion asks the evaluator to measure the extent to which the generated text is free of grammatical errors~\cite{howcroft2020}.

Evaluation mode, in turn, describes the approach by which the human evaluator provides a measure of how successful the NLG system has been in capturing a given quality criterion (or criteria)~\cite{belz2020}.

Evaluation modes also have their share of limitations, often hindered by subjective criteria that makes interpretation and comparison of scores recorded by multiple evaluators difficult~\cite{hamalainen2021}. This is particularly problematic when comparing studies~\cite{amidei2019}. Additionally, there is also a clear lack of consensus as to the most effective evaluation modes to use in human evaluation of NLG systems, with a wide variety of scoring techniques commonly being used in the literature~\cite{briakou2021,van-der-lee2019}.

Common evaluation modes include:

\textbf{Preference}: This evaluation mode involves the evaluator selecting their preferred text, or texts, from a set of texts~\cite{van-der-lee2019}. These texts are typically a collection of generated outputs from a series of models or a combination of AI-generated and human-created texts~\cite{van-der-lee2019}.

\begin{figure}[H]
\centering
\includegraphics[width=0.95\linewidth]{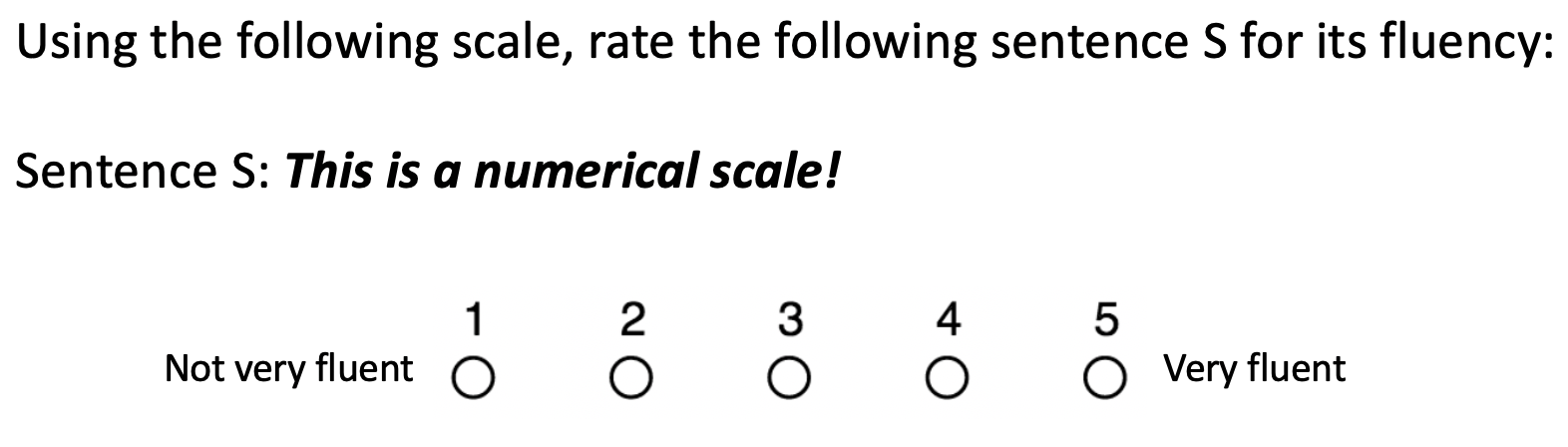}
\caption{Example of a numerical scale.}
\label{fig:numerical_scale}
\end{figure}

\textbf{Numerical Scale}: One of the most commonly used evaluation modes, numerical scales ask the evaluator to rate the quality of a set of generated text(s) on a sliding scale (e.g., from 1 -- 5 as shown in Fig.~\ref{fig:numerical_scale})~\cite{amidei2019}. This allows for a more fine-grained measurement of the quality of the generated outputs compared to binary scoring~\cite{celikyilmaz2020}.

\begin{figure}[H]
\centering
\includegraphics[width=0.95\linewidth]{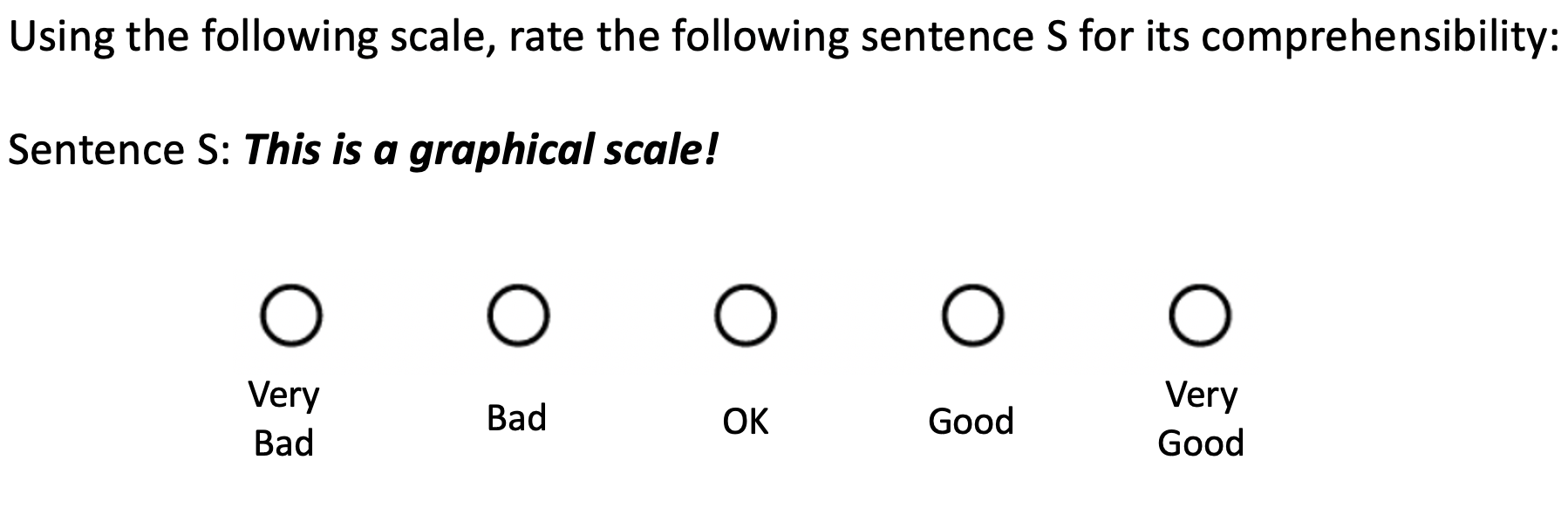}
\caption{Example of a graphical scale.}
\label{fig:graphical_scale}
\end{figure}

\textbf{Graphical Scale}: Similar to numerical scales, graphical scales utilise words or phrases (as opposed to numbers) for each rating value (See Fig.~\ref{fig:graphical_scale} for an example)~\cite{amidei2019}.

\begin{figure}[H]
\centering
\includegraphics[width=0.95\linewidth]{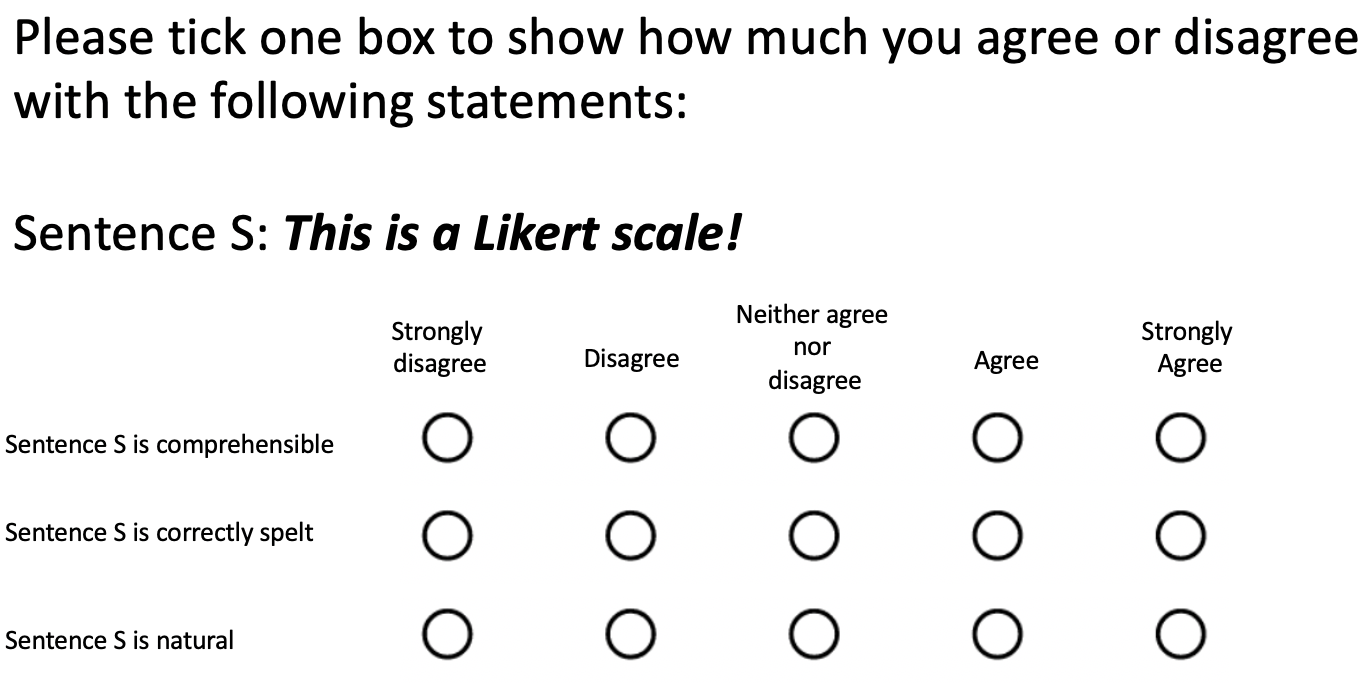}
\caption{Example of a Likert scale.}
\label{fig:likert_scale}
\end{figure}

\textbf{Likert Scale}: Likert Scales are an aggregate scale comprised of multiple graphical scales called Likert Items~\cite{amidei2019} (see Fig.~\ref{fig:likert_scale} for an example). This allows for a survey-style approach that record overall evaluator impressions based on responses to multiple dimensions of a set of generated texts. Due to the lack of consistent intervals between scale items, however, many researchers argue that Likert Scale results must only be considered in aggregate, though there is much disagreement regarding this~\cite{amidei2019}. This confusion makes evaluator--evaluator and study--study comparisons using Likert Scale evaluation particularly difficult (though all of the evaluation modes presented here suffer from this to some extent). Despite this, Likert Scales remain popular in NLG evaluation~\cite{amidei2019}.

\textbf{Ranking}: To overcome the limitations of the above approaches, ranking has been used as a means of human-based evaluation. Rather than scoring the generated texts, the evaluator instead ranks the generated texts according to their quality~\cite{hamalainen2021}. This has the advantage of allowing for better inter-evaluator comparison, but only provides a relative appreciation of the NLG model or its outputs, rather than a true measure of its output quality~\cite{celikyilmaz2020}. Ranking can also be limited by its complexity, with large numbers of comparisons becoming prohibitively complex.

Extending the use of direct measures of quality, some studies leverage existing inter-annotator agreement metrics to better account for the level of unity in the scores provided by a set of human evaluators~\cite{celikyilmaz2020}. These metrics measure the amount of agreement between evaluators and provide indications of the extent to which evaluators concur with the suitability or quality of the NLG model they are evaluating~\cite{amidei2019Agree}. 
It is worth noting that the use of inter-annotator metrics in NLG evaluation is generally less common than the sole use of individual metrics (appearing in only 12.5\% of papers examined in \cite{van-der-lee2019}), and that even when they are used, the agreement scores are typically lower than what would be considered `acceptable'~\cite{amidei2019Agree,briakou2021,celikyilmaz2020,hamalainen2021,van-der-lee2019}.

Furthermore, there are questions regarding the suitability of inter-annotator agreement as a sole metric for evaluating consistency in NLG evaluations~\cite{amidei2019Agree}. Due to the ambiguous and varied nature of language, there are questions regarding whether a metric focusing on strict agreement -- as is the case with inter-annotator agreement metrics -- is suitable given the potential for different (but valid) interpretations of the same texts~\cite{amidei2019Agree}.

The most commonly used inter-annotator metrics are:

\textbf{Percent Agreement}: Percent agreement is the most straightforward means of measuring agreement between two independent evaluators. It simply reports the percent of cases in which the two evaluators agreed with each other~\cite{celikyilmaz2020}. Whilst popularly used in NLG evaluation, percent agreement fails to account for the possibility that agreement between evaluators may occur by chance~\cite{amidei2019Agree}. This is particularly problematic when utilising individual metrics with fewer scoring options such as binary scoring. Percent Agreement is given as

\[P_a = \frac{\sum_{i=0}^{|X|} a_i}{|X|} \]

where $X$ is a set of generated texts for which evaluators assign a score to each text $x_i$, and $a_i$ is the characteristic function denoting agreement in the scores for $x_i$. Hence, $a_i=1$ if the evaluators assign the same score and $a_i=0$ if not.

\textbf{Cohen's~$\kappa$}: Improving on percent agreement, Cohen's~$\kappa$ is able to account for the possibility of agreement occurring by chance in the annotations of two evaluators~\cite{cohen1960}. To achieve this, Cohen's~$\kappa$ defines the probability of two evaluators, $e_1$ and $e_2$, agreeing by chance as

\[P_c = \sum_{s \in S}^{} P(s|e_1) * P(s|e_2)\]

where $S$ is the set of all possible scores for texts in $X$. The conditional probabilities $P(s|e_i)$ are estimated via the frequency with which the given evaluator assigned each of the possible scores in $S$. Combining percent agreement $P_a$ with the probability of agreement by chance $P_c$, Cohen's~$\kappa$ is defined as

\[\kappa = \frac{P_a - P_c}{1 - P_c}.\]
    
\textbf{Fleiss'~$\kappa$}: Fleiss'~$\kappa$ improves on Cohen's~$\kappa$ by measuring the agreement of more than a single pair of evaluators by considering all pairwise inter-annotator agreements~\cite{fleiss1971}. To this, $a_i$, the agreement of scores for two evaluators for a given generated text, is redefined as

\[a_i = \frac{\sum_{s \in S}^{} \text{\# pairs scoring $x_i$ as $s$}}{\text{\# evaluator pairs}}. \]

The probability of agreement by chance, $P_c$ is also redefined by estimating the probability of a given score by the frequency of that score across all evaluators. This is defined as

\[P_c = \sum_{s \in S}^{} r_s^2\]

where $r_s$ is the proportion of evaluators that assigned a given score $s$. Fleiss'~$\kappa$ is thus defined using the same definition of Cohen's~$\kappa$, combining the definition of $P_a$ used for percent agreement alongside the redefined $P_c$.
    
\textbf{Krippendorff's $\alpha$}: Reevaluating the approaches above to consider the likelihood of disagreement, Krippendorff's $\alpha$, as with Fleiss'~$\kappa$, can be used to evaluate multiple annotators whilst accounting for agreements that occurred by chance~\cite{krippendorff1970}. Moving beyond Fleiss'~$\kappa$, however, Krippendorff's $\alpha$ is also capable of handling missing values, where Fleiss'~$\kappa$ and Cohen's~$\kappa$ cannot~\cite{krippendorff1970}. To define Krippendorff's $\alpha$, we first find the probability of disagreement using

\[P_d = \sum_{m=0}^{|S|} \sum_{n=0}^{|S|}w_{m,n} \sum_{i=0}^{|X|} \frac{\text{\# pairs scoring $x_i$: $(s_m,s_n)$}}{\text{\# of evaluator pairs}}\]

where $(s_m, s_n)$ indicates one possible score pair, and $w_{m,n}$ the weight used to adjust the degree of penalisation for a given disagreement. The probability of agreement by chance is also redefined, using $r_{m,n}$ to represent the proportion of all evaluation pairs that assign the scores $s_m$ and $s_n$. Given this, the probability of agreement by chance is defined as

\[P_c = \sum_{m=0}^{|S|} \sum_{n=0}^{|S|}w_{m,n}r_{m,n}.\]

Given the probability of disagreement $P_d$, and the redefined probability of agreement by chance $P_c$, Krippendorff’s $\alpha$ is calculated using

\[\alpha = 1 - \frac{P_d}{P_c}.\]

Whilst a range of automated evaluation measures and metrics exist, including more sophisticated approaches built around trained ML models, most automated evaluation of NLG leverage untrained automated evaluation metrics~\cite{celikyilmaz2020}. These automated metrics offer an objective, easy-to-implement method for measuring the quality of generated texts~\cite{finch2020}, centred around the comparison of a set of generated texts to a gold-standard set of (generally human-created) reference texts~\cite{celikyilmaz2020}. The assumption is that the closer the generated texts are to the reference texts, the better. Whilst a large number of automated evaluation measures exist, the most commonly used in NLG studies are n-gram overlap metrics~\cite{celikyilmaz2020,finch2020,van-miltenburg2020}. 

N-gram overlap metrics are designed to measure the degree of similarity in the n-grams present in the generated texts when compared to a set of reference texts~\cite{garbacea2019}. These approaches typically leverage word-based n-grams, though other n-gram approaches (e.g., character n-grams) can be used. The assumption is that the larger the overlap in n-grams between the generated text and the reference texts, the higher the quality of the generated text.

\begin{figure}[!htb]
\centering
\includegraphics[width=0.95\linewidth]{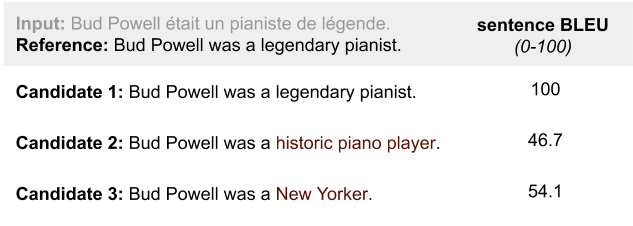}
\caption{Example of BLEU for three generated candidates. Retrieved from \cite{sellam2020}}
\label{fig:bleu}
\end{figure}

Some of the most commonly used n-gram overlap metrics are:  

\textbf{BLEU}: The Bilingual Evaluation Understudy (BLEU), is one of the oldest and most commonly used n-gram overlap metrics~\cite{papineni2002}. Originally intended for evaluating machine translation tasks, BLEU has seen further use in other generation tasks including style transfer, story generation, and question generation~\cite{celikyilmaz2020}. BLEU works by comparing the overlap in the n-grams of a candidate (generated) text and the n-grams of a set of reference texts using the weighted geometric mean of modified n-gram precision scores~\cite{papineni2002}. N-gram precision scores are calculated by measuring the fraction of n-grams appearing in the generated text that appear in any of the reference texts. An example of BLEU can be found in Fig.~\ref{fig:bleu}. In this figure, note that candidate 2 is ranked lower than 3, despite more closely matching the meaning of the reference. This highlights a key limitation of n-gram-based metrics, which can overemphasise surface level lexical similarities.

\textbf{ROUGE}: The Recall-Oriented Understudy for Gisted Evaluation (ROUGE) works in much the same way as BLEU, but focuses on measuring recall rather than precision~\cite{lin2004}. In other words, ROUGE measures the fraction of n-grams in the references texts that appear in the generated candidate text. ROUGE itself is a broad class that describes a set of variants. Most commonly used are the ROUGE-N variants, where $N$ is the size of n-gram to be evaluated (e.g., ROUGE-1 evaluates unigram overlaps)~\cite{celikyilmaz2020}. Another common variant is ROUGE-L, which evaluates the longest sequence of shared tokens in both the generated and the reference texts~\cite{lin2004}. ROUGE is generally considered to yield more interpretable scores than BLEU~\cite{celikyilmaz2020}.
    
\textbf{METEOR}: The Metric for Evaluation of Translation with Explicit ORdering (METEOR) attempts to improve on some of BLEU's weaknesses by utilising the weighted F-score using unigrams, where recall is weighted more heavily than precision as this has been found to yield higher correlations with human judgement~\cite{lavie2007}. Moreover, METEOR also incorporates a penalty function that penalises incorrect unigram order~\cite{lavie2007}.
    
\textbf{CIDEr}: The Consensus-based Image Description Evaluation (CIDEr) utilises a consensus-based protocol to measure the similarity of a generated sentence to that of a set of human-created reference sentences using TF-IDF weighted n-gram frequencies~\cite{vedantam2015}. Originally intended for evaluating generated image captions, CIDEr has also been used in the evaluation of other NLG tasks including online review generation~\cite{garbacea2019}.

\begin{figure*}[!ht]
\centering
\includegraphics[width=0.98\linewidth]{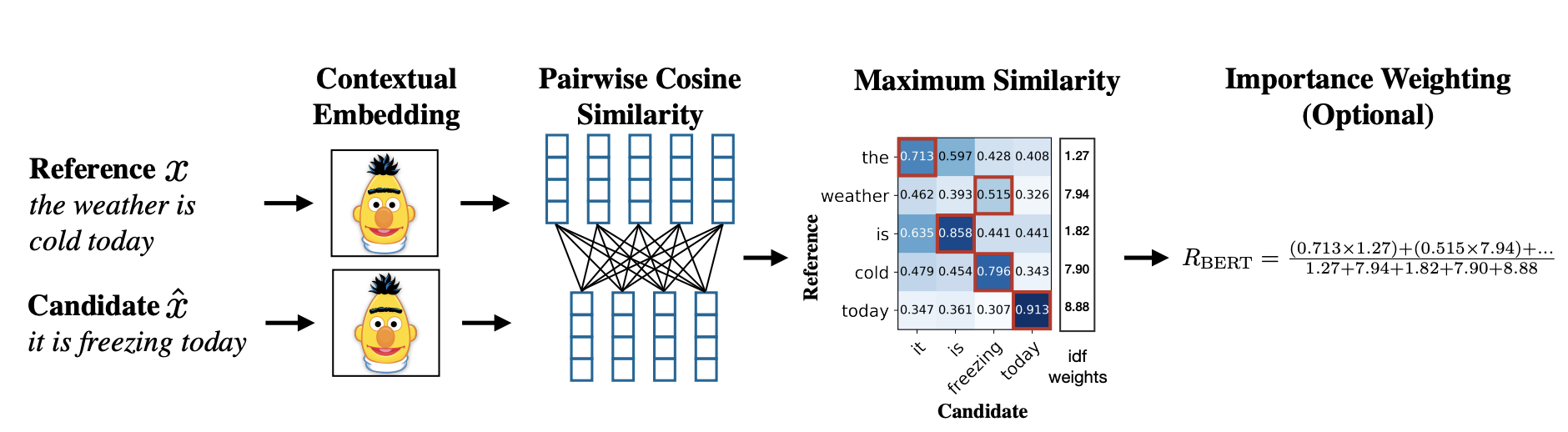}
\caption{Diagram of BERTScore. Retrieved from \cite{zhang2019}}
\label{fig:BERTScore}
\end{figure*}

Whilst being some of the most commonly used metrics in NLG evaluation, n-gram overlap metrics are often criticised for their lack of correlation with human judgement~\cite{celikyilmaz2020,van-der-lee2019,van-miltenburg2020}. Additionally, these metrics fail to account for more holistic qualities in generated texts including fluency and grammatical correctness~\cite{van-miltenburg2020}. Moreover, the assumption that a good text is equivalent to a text that closely mirrors the reference set is potentially weak, as unexpected texts could still be of sufficient quality~\cite{van-miltenburg2020}. This leads to additional difficulties in justifying the relevancy of the reference set and, in turn, the meaningfulness of the metric scores achieved~\cite{van-der-lee2019}. Due to the high degree of criticism levelled at untrained automated NLG metrics, more recent proposals have been made to leverage machine-trained evaluators to measure an NLG system's quality~\cite{celikyilmaz2020}. 

Whilst these metrics leverage more sophisticated ML models, they are typically still used to examine similarities between the generated samples and a series of reference texts~\cite{celikyilmaz2020}. The use of the machine-trained evaluator is thus a means of conducting more sophisticated comparisons that leverage more latent, and more relevant, aspects of text (e.g., semantics and syntax) as opposed to the more surface-level comparisons of the untrained metrics above. 

A common approach is to leverage machine-trained models for measuring the semantic similarity between generated and reference texts via the use of learned word and sentence embeddings~\cite{celikyilmaz2020}. Various attempts have been made towards this, including more traditional embedding approaches such as skip-thought~\cite{kiros2015}, fastsent~\cite{hill2016}, quick-thought~\cite{logeswaran2018}; and newer approaches that leverage PLMs and contextual embeddings like BERTScore (Fig.~\ref{fig:BERTScore})~\cite{zhang2019}. In essence, these approaches examine similarities in syntax and semantics between generated texts and gold-standard reference texts via examining the embedding distances between the two using some form of distance measure (e.g., cosine similarity)~\cite{celikyilmaz2020}. The assumption being that higher quality generated texts will more closely match the reference texts in terms of both semantics and syntax. 

Beyond this, further proposals have been made to leverage trained models to perform regression-style evaluation of NLG translation systems, such as the GRU-based RUSE~\cite{celikyilmaz2020}, which are used to predict a scalar value indicative of the quality of a generated (translated) text relative to a reference text. Semi-supervised methods, like ADEM~\cite{lowe2017} and HUSE~\cite{hashimoto2019}, have also been proposed as a means of leveraging human judgements in the decision making of the ML model~\cite{celikyilmaz2020}.

In general, the research conducted towards developing machine-learned metrics are promising, with high correlations often being found between their scores and human judgement. Despite this, however, their adoption is still relatively low in NLG evaluation~\cite{celikyilmaz2020}.

\subsection{Evaluation Methods}

As with the majority of the items above, the methodological process of evaluating NLG systems is typically categorised by the type of evaluator used: i.e., human-based methods and machine-based methods~\cite{celikyilmaz2020}. 

As discussed above, human judgement is typically considered the most effective form of NLG evaluation~\cite{celikyilmaz2020}. Due to the innate ability of humans towards language and their clearer appreciation of the role of context and semantics, human evaluator methods can be highly effective.

With human approaches to NLG evaluation, the most common methods utilise intrinsic, static forms of evaluation~\cite{celikyilmaz2020,finch2020}. Through this approach a set of human evaluators are generally presented with a series of generated texts and will be tasked with providing judgements on the quality of these texts based on their ability to maximise a set of quality criteria, via a prescribed scoring mode~\cite{belz2020}. The scores recorded by each human evaluator are then generally analysed in some manner, typically via basic statistical measures, to gain an overall appreciation of the capabilities of the NLG system in question~\cite{hamalainen2021}. 

Interactive intrinsic human evaluation methods are also possible, and are commonly advocated for when evaluating dialogue systems~\cite{finch2020}. These approaches again utilise the basic dimension-scoring measures, but allow the human evaluator to examine the artificial texts as they are generated in real-time~\cite{finch2020}. This approach also allows the human evaluator to directly prompt the NLG system, allowing for the evaluator to gain a better sense of the quality and relevance of the texts being generated. 

Additionally, some approaches have utilised human evaluation methods with a focus on interactive extrinsic evaluation~\cite{celikyilmaz2020}. These methods of human-based extrinsic evaluation thus measure how effective the NLG system is at allowing the user to succeed at a given task. One early example of this utilised an instruction generation system, where human evaluators were required to follow the generated instructions~\cite{celikyilmaz2020}. Evaluations were then made based on the success of the evaluators in achieving the desired tasks by following the instructions. 

Moreover, extrinsic interactive human evaluation is more commonly used in evaluating dialogue systems and chatbots~\cite{celikyilmaz2020}. These methods typically utilise some kind of feedback form or dimension-based scoring measure and evaluate the ability of the dialogue system to meet user needs over longer periods of time. This is distinct from intrinsic interactive evaluation, as the focus is on the satisfaction of user requirements  rather than the direct quality of the dynamically generated text~\cite{finch2020}.

There are, however, distinct limitations and inconsistencies in the current usage of human evaluation methods of NLG, the core issue being that there is little agreement or standardisation in the manner in which human evaluation should be conducted~\cite{belz2020,hamalainen2021,howcroft2020,schoch2020}.

In the literature at large, there is a wide range of evaluator numbers used, with some studies leveraging as few as two, whilst others use 500+ (typically through crowd-sourcing)~\cite{briakou2021,hamalainen2021,garbacea2019}. Additionally, some studies rely on expert evaluators, whilst others recruit non-experts -- typically with little justification for this decision~\cite{van-der-lee2019}. Moreover, many studies do not report the number of evaluators used~\cite{briakou2021,schoch2020,van-der-lee2019}.

There is also little consensus on how many generated samples should be evaluated, with as few as two up to more than 5,400~\cite{briakou2021,hamalainen2021,van-der-lee2019}. Moreover, some studies provide the same set of samples to all evaluators, whilst others provide different subsets to each evaluator~\cite{van-der-lee2019}. It is currently unclear as to how these variations between studies may effect the quality of the evaluations performed. 

There are also questions regarding the manner in which evaluators are selected~\cite{briakou2021,hamalainen2021}. Some research has published concerns as to the role that selection bias may play in evaluation, particularly with the use of crowd-sourced evaluators where demographic information is hard to produce. This, coupled with the typically low number of evaluators used, could cause further unconsidered effects on the evaluators given~\cite{hamalainen2021}.

\begin{figure}[!htb]
\centering
\includegraphics[width=0.8\linewidth]{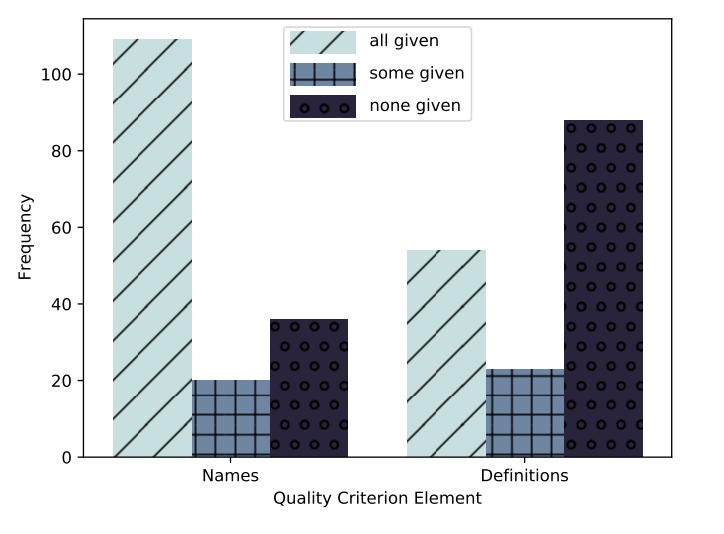}
\caption{The number of papers explicitly naming and defining the quality criterion used in their human evaluation. Retrieved from \cite{howcroft2020}.}
\label{fig:quality_criterion_reporting}
\end{figure}

Finally, there are significant inconsistencies in the reporting of human evaluation methods~\cite{belz2020,briakou2021,howcroft2020,schoch2020,van-der-lee2019}. In turn, it is not uncommon for NLG papers to not include details of the number of evaluators used, the questions posed to evaluators, or even the manner of scoring or quality criterion that the evaluators used~\cite{howcroft2020,schoch2020,van-der-lee2019}. An example of this is provided by \citet{howcroft2020} (as shown in Fig.~\ref{fig:quality_criterion_reporting}), in which they note that over half the papers studied did not define the quality criterion used in their evaluations. These inconsistencies in reporting within nearly every aspect of the human evaluation method further emphasise the problems above, compounding the difficulties of replicability and comparison between NLG studies~\cite{howcroft2020,schoch2020}.

Whilst machine-based evaluations can, in theory, be conducted using both static and interactive approaches and intrinsic or extrinsic approaches, in general automated evaluation has focused on the use of machine-based evaluation as applied to intrinsic, static evaluation~\cite{celikyilmaz2020,finch2020,garbacea2019}. 

Generally speaking, machine-based evaluation is centred around the leveraging of one or more of the untrained automated evaluation metrics discussed in Section~\ref{label:eval_measures}~\cite{celikyilmaz2020}. Often, multiple metrics will be reported as a means of attempting to balance the weaknesses of each individual metric~\cite{finch2020}. 

The popularity of this approach is likely due to the efficiency of implementation and the ease with which it can be scaled to evaluate large numbers of generated texts/systems. However, many criticisms have been levied towards the use of these automated metrics, arguing that they offer poor indications of the genuine quality of NLG systems with typically low correlations with human judgement~\cite{finch2020}. Additionally, issues of reporting are equally prevalent, with studies often neglecting to include relevant information such as the the number of generated samples evaluated and the sampling method used to select these generated samples.

Moreover, whilst the usage of machine-trained evaluation approaches are becoming more studied, they are still relatively uncommon~\cite{celikyilmaz2020}. This is likely to be an area of distinct progress in future, due to the unreliable nature of existing, untrained approaches to automated NLG evaluation.

To help compensate for these weaknesses, it is common for studies to utilise human evaluators to provide greater insights and confidence to the performance of a given NLG system, whilst also reporting automated metrics to better aid with replicability and comparison with the state-of-the-art~\cite{celikyilmaz2020}. 

However, limitations regarding the lack of reporting in general, replicability in regard to human evaluators, and lack of correlation with human judgement in regard to automated metrics, means these combined approaches are still vulnerable to some of the key weaknesses inhibiting current NLG evaluation methods~\cite{celikyilmaz2020,finch2020,hamalainen2021}.

\subsection{Standards}

The evaluation of NLG systems is currently hampered by the distinct lack of standardised approaches and generalisable methodologies~\cite{belz2020,howcroft2020,van-der-lee2019,van-miltenburg2020}. Instead, NLG papers -- even those conducting similar tasks -- often take highly contrasting approaches to their evaluation~\cite{celikyilmaz2020}. Even with more popular approaches, such as the use of human evaluators using a quality criterion-evaluation mode method, the exact specifications of these approaches can vary significantly~\cite{belz2020,van-der-lee2019}. This includes variations in definitions of quality criteria, scoring methods used, and the construction of the evaluation methodology as a whole~\cite{belz2020,hamalainen2021}. Moreover, a lack of adequate recording of the evaluation approaches taken is additionally problematic, often making it hard to fully understand the evaluative process of a given NLG study and making it highly difficult to replicate and/or compare studies~\cite{van-der-lee2019}.

\begin{figure}[!htb]
\centering
\includegraphics[width=0.5\linewidth]{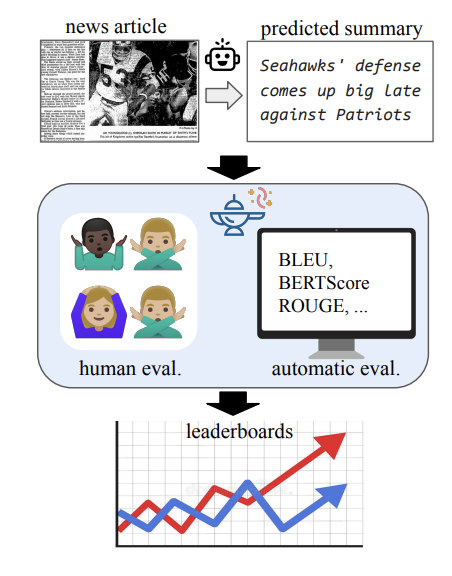}
\caption{The GENIE architecture applied to a text summarisation task. GENIE combines both human and automated evaluation techniques. Retrieved from \cite{GENIE2021}.}
\label{fig:GENIE_architecture}
\end{figure}

Given this, one could be forgiven for assuming no such standardised NLG evaluative approaches exist, but this is not the case. In \cite{celikyilmaz2020}, the authors detail several examples of existing platforms developed to standardise NLG evaluation. These include GENIE~\cite{GENIE2021} (an example of the GENIE architecture can be found in Fig.~\ref{fig:GENIE_architecture}) and GEM~\cite{GEM2021}; two evaluation platforms that have been proposed as a means of bench-marking NLG systems using both automated and human evaluation across multiple datasets and NLG tasks, including text summarisation, text simplification, and dialogue generation~\cite{celikyilmaz2020}. Moreover, other task-specific evaluation platforms also exist, including ChatEval which provides a web-interface for standardised evaluation and comparison of NLG chatbots and dialogue systems to the state-of-the-art results, leveraging both automated metrics and human judgement~\cite{chateval2019}. Despite the existence of these platforms, however, adoption has been very low across NLG research.

To help solve the problems posed by a lack of consistency and generalisability, many of the review papers included in this study propose best practices that should be followed when conducting NLG evaluation.

In \cite{schoch2020}, the authors discuss the need for clearer and more consistent reporting of NLG human evaluation methodologies. To this end, they list a set of important design parameters that should be detailed when reporting. These aspects focus on: including discussion of question design and presentation, including wording and scoring approaches; detailed discussion of the quality criteria used, including the clear naming of each criterion and a description of how these criteria are defined in the work; and, an in-depth description of any human evaluators used, including details of their expertise, demographics, recruitment, and compensation. The authors also note that justifications of these decisions should be included where relevant and be rooted in previous literature where possible to improve consistency and aid in cross-study comparison.

\citet{belz2020}, \citet{van-der-lee2019}, and \citet{howcroft2020} provide similar guidance as to the best practices in reporting human NLG evaluation. The authors, in turn, highlight similar needs for details regarding quality criteria, the evaluators used, and the questions posed. They also highlight the need for additional details not mentioned by \citet{schoch2020}, such as a clear discussion of the evaluation design, including the number of evaluators used, the number of samples provided to each evaluator, the information given to evaluators (e.g., training, instructions, interface), and the method for sampling the generated outputs provided to evaluators.

In summary, there exist a range of papers aimed at developing generalisable approaches and best practices to NLG evaluation. However, adoption of these appear low, with little indication of a trend towards more standardised approaches to NLG evaluation. This, in turn, remains an ongoing problem that needs addressing to ensure adequate NLG evaluation, and to allow for clear comparison between different approaches to various NLG tasks.

\section{Relevant AI Techniques for NLG\label{sec:techniques}}

Recent advancements in AI, particularly in the field of deep learning, have led to the development of more powerful LMs that can better `understand' natural languages. This, in turn, has led to increased capabilities towards the generation of more realistic and convincing natural language text. This section enumerates some of the commonly used AI techniques for NLG, including neural networks, Transformers, and combined techniques which make use of multiple methods.

\subsection{Neural Networks (NN)}

NNs have been extensively used in NLG. In spite of the paradigm shift toward Transformer-based language modelling in recent years~\cite{LTZW2021}, NNs are still used in many NLG applications. The most commonly used NNs are as follows:
   
\textbf{Recurrent Neural Network (RNN)}: RNNs are designed to model sequential information -- of which text is a common example -- and as such have been commonly used in NLG. Early encoder-decoder frameworks for NLG were generally based on RNNs~\cite{YZLHWJJ2020}. In addition, RNNs have been used in \emph{controllable NLG}, which aims to generate text with controllable attributes, such as sentiment, formality and politeness~\cite{PBS2020}. Apart from being utilised as a generation method, RNNs have also been used in the automatic evaluation of NLG models~\cite{celikyilmaz2020}. Whilst the RNN has proven effective at modelling language, it suffers from an inability to adequately `remember' relevant information over long sequences~\cite{PBS2020}. To solve this, two variants of RNN have been proposed: 
    
\begin{itemize}
\item \textbf{Long Short-Term Memory (LSTM)}: LSTM is a form of RNN that is equipped with an additional memory cell which allows it to better remember information over time (and thus handle longer text more effectively)~\cite{PBS2020}. To achieve this, LSTM utilises a series of gates in order to dictate when pieces of information are remembered and when they are forgotten~\cite{PBS2020}. Due to its ability to model longer texts, LSTM has been frequently used in NLG. Models leveraging LSTMs for NLG evaluation have also been proposed~\cite{celikyilmaz2020}.
    
\item \textbf{GRU}: Like LSTMs, GRUs are another refinement of the RNN. GRUs are similar to LSTMs, leveraging gating to help mitigate the problems posed by longer sequences~\cite{PBS2020}. However, GRUs are simpler in nature than LSTM, with fewer gates and no additional memory cell~\cite{PBS2020}. This typically allows GRUs to be trained faster and to achieve better performances than LSTMs on smaller amounts of training data. However, this also means that GRUs are typically less effective at handling longer sequences. Similar to LSTMs, GRUs have seen frequent use as a generation method, and have also been proposed as a means of evaluating NLG models~\cite{celikyilmaz2020}.
\end{itemize}
    
\textbf{Convolutional Neural Networks (CNN)}: Whilst more commonly used in computer vision tasks, CNNs have recently seen a wide area of utilisation in NLG, from text generation to topic modelling for knowledge-enhanced NLG~\cite{YZLHWJJ2020}. In NLG, CNN-based encoder-decoder frameworks have been increasingly preferred.
    
\textbf{Graph Neural Networks (GNN)}: GNNs are neural models that capture the dependence of graphs via message passing between the nodes of graphs. They have the potential to combine graph representation learning and text generation. This can enable the integration of knowledge graphs, dependency graphs, and other graph structures into NLG~\cite{YZLHWJJ2020}.

\begin{figure*}[!htb]
\centering
\includegraphics[width=\linewidth]{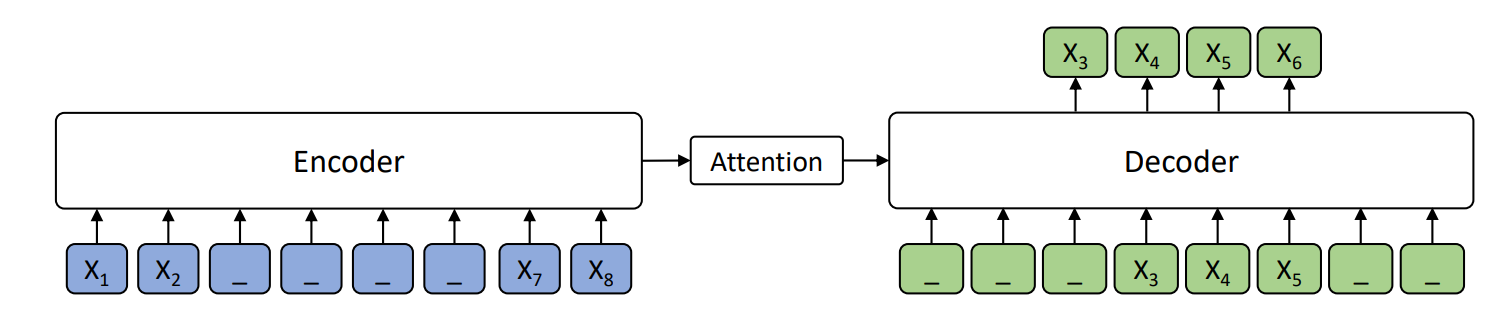}
\caption{Encoder-decoder architecture of MASS. Retrieved from \cite{STQLL2019}.}
\label{fig:mass}
\end{figure*}
   
\subsection{Transformers}

Transformers are deep learning models adopting an attention mechanism which can provide context for any position in the input~\cite{vaswani2017}. Therefore, unlike RNNs, Transformers do not need to process data in order. This paves the way for greater parallelisation, which reduces the amount of time required for training and allows models to be trained over larger corpora. As a consequence of this, transformer-based PLMs are commonly used for NLG~\cite{DCLT2018, radford2019}. In terms of their architectures, Transformers can be categorised into three categories:

\textbf{Encoder-Only Transformers: } These types of Transformer only leverage a single Transformer encoder block to build a LM. The most well-known examples are as follows:
    
\begin{itemize}
\item \textbf{Bidirectional Encoder Representations from Transformers (BERT)~\cite{DCLT2018}}: BERT is a PLM developed by Google, which is used in a wide range of NLP tasks. Moreover, it is contextual and bidirectional, meaning that BERT can contextualise each word in an input utilising both its left and right context. BERT has been widely adapted as an NLG method~\cite{LTZW2021} and has also seen use in the development of metrics for NLG evaluation, such as BERTScore, RoBERTa-STS and BLEURT~\cite{celikyilmaz2020}.

\item \textbf{Unified pre-trained Language Model (UniLM)~\cite{DYWWLWGZH2019}}: UniLM, developed by Microsoft, combines multiple LM pre-training objectives: unidirectional (both left-to-right and right-to-left), bidirectional and sequence-to-sequence prediction. 
\end{itemize}
    
\textbf{Decoder-Only Transformers}: This type of Transformer contains only a single Transformer decoder block used for language modelling.

\begin{figure*}[!htb]
\centering
\includegraphics[width=0.8\linewidth]{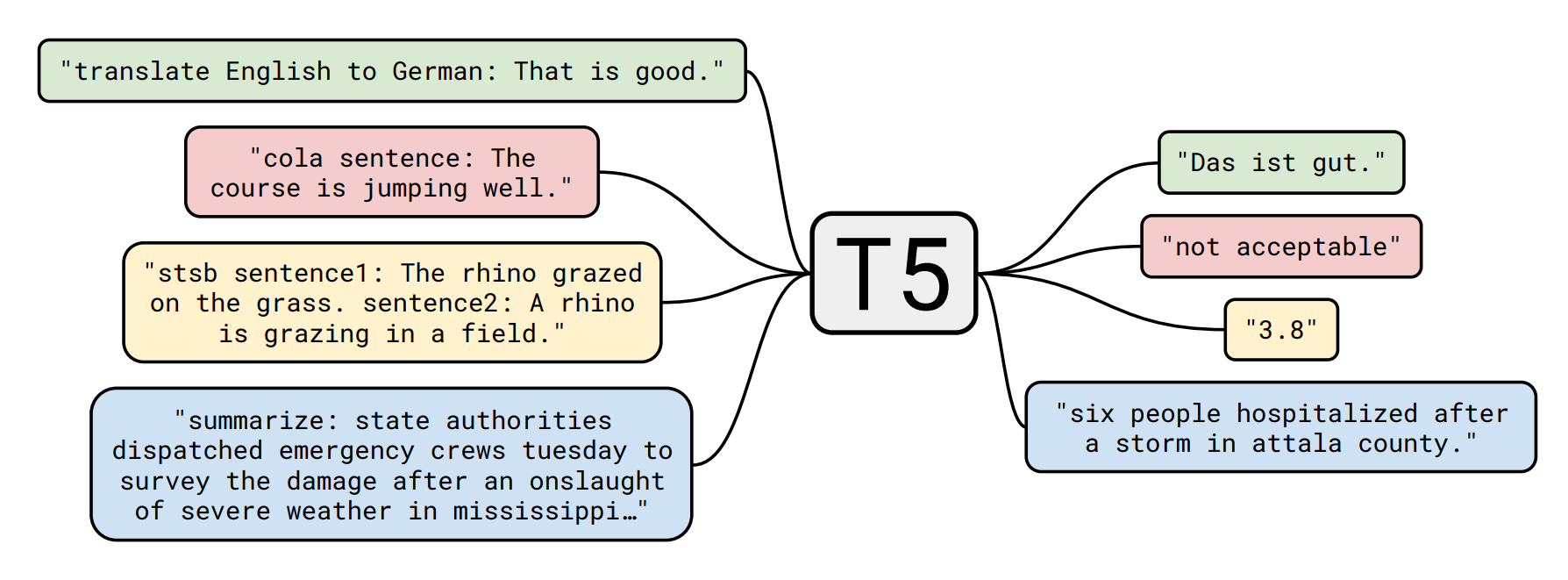}
\caption{Example input-output pairs generated with T5. Retrieved from \cite{RSRLNMZLL2020}.}
\label{fig:t5}
\end{figure*}
    
\begin{itemize}
\item \textbf{Generative Pre-trained Transformer (GPT)~\cite{BMRSK2020}}: GPT is a unidirectional auto-regressive PLM, developed by OpenAI. It has two successors, GPT-2 and GPT-3, which have been trained on larger datasets and have larger numbers of training parameters. While all GPT versions share a similar architecture, GPT-3 is the largest PLM with 175 billion parameters.

\item \textbf{Conditional Transformer Language Model (CTRL)~\cite{KMVXS2019}}: CTRL is a PLM developed by Salesforce that allows users to control generated content by providing \emph{control codes}. Control codes can be URLs, questions, or languages, and enable users to explicitly specify domains, subdomains, entities and dates. CTRL has been trained on 50 control codes.

\item \textbf{XLNet~\cite{YDYCSL2019}}: XLNet is a generalised autoregressive pre-trained method that utilises permutation language modelling to combine the advantages of autoregressive and bidirectional language modelling objectives. It employs Transformer-XL, an improved Transformer architecture, as the backbone model.
\end{itemize}
    
\textbf{Encoder-Decoder Transformers}: This corresponds to the standard encoder-decoder architecture in which there are two stacks of Transformer blocks. The encoder is thus fed with an input sequence,  and the decoder tries to generate the output sequence based on an encoder-decoder self-attention mechanism~\cite{LTZW2021}.
    
\begin{itemize}
\item \textbf{BART~\cite{LLGGMLSZ2019}}: BART is a denoising autoencoder for pre-training sequence-to-sequence models, developed by Facebook. It combines autoregressive and bidirectional language modelling objectives by using a bidirectional encoder (e.g., BERT) and an autoregressive decoder (e.g., GPT).

\item \textbf{Masked Sequence to Sequence Pre-training for Language Generation (MASS)~\cite{STQLL2019}}: MASS is a masked bidirectional sequence-to-sequence pre-training method for NLG, proposed by Microsoft. It jointly trains the encoder and decoder by feeding the encoder with a sentence containing a randomly masked fragment, and using the decoder to try to predict the masked fragment, as shown in Fig.~\ref{fig:mass}.

\item \textbf{Text-To-Text Transfer Transformer (T5)~\cite{RSRLNMZLL2020}}: T5 is a text-to-text framework proposed by Google to reframe all NLP tasks into a unified text-to-text format in which the input and the output are always text strings, rather than a class label or a span of the input. The idea behind this proposal is to be able to use the same model, loss function, and hyperparameters on any NLP task. Some example input-output pairs can be seen in Fig.~\ref{fig:t5}. While not its intended use, T5 can be adapted for controllable NLG~\cite{PBS2020}.
\end{itemize}

\subsection{Combined AI Techniques}

Considering the complexity of the NLG task, it is a reasonable approach to combine multiple AI techniques to be able to make use of the advantages of each technique. Some examples of popular combined AI techniques are mentioned below.

\textbf{Plug-and-Play Language Model (PPLM)~\cite{SAJJEPJR2020}}: PPLM, developed by Uber, is an LM aimed at controlled NLG which allows users to flexibly plug in small attribute models representing the desired control objective(s) into a large unconditional LM, e.g., GPT. The main difference between PPLM and CTRL is that PPLM does not require any additional training or fine-tuning.
    
\textbf{Generative Adversarial Network (GAN)}: GANs contain two neural networks, a generator and a discriminator, which compete with each other to provide more accuracy. CNN and RNN models, as well as their variants, are frequently used as generators and/or discriminators in GANs. Since GANs were originally designed for generating differentiable values, using it for discrete language generation is not easy. However, GANs still have several applications in the context of NLG, such as poetry and lyrics generation~\cite{shahriar2021}. There exist a number of GAN variants specialised for NLG:

\begin{itemize}
\item \textbf{seqGAN} is a sequence generation framework aimed at  solving the problems of GANs regarding discrete token sequence generation, e.g., texts. It considers the GAN generator as a Reinforcement Learning (RL) agent, and the RL reward signal is received from the GAN discriminator judged on a complete sequence~\cite{YZWY2017}.

\item \textbf{I2P-GAN} is a GAN-based model for automatic poem generation relevant to an input image. The model involves a deep coupled visual-poetic embedding model to learn poetic representations from images and a multi-adversarial training procedure optimised with policy gradient. In its architecture, there exists a CNN-RNN generator that acts as an agent, and two discriminators provide rewards to the policy gradient~\cite{LFKY2018}.

\item \textbf{RankGAN} focuses on one of the limitations of GAN discriminators, which is that they are  generally binary classifiers. In this manner, it aims to improve GAN for generating high-quality language descriptions by enabling the discriminator to analyse and rank a collection of human-written and machine-generated sentences. RankGAN uses LSTMs for the generator and a CNN for the discriminator~\cite{LLHZS2017}. 

\item \textbf{MaskGAN} is an actor-critic conditional GAN which can fill in missing text according to the surrounding context. It uses LSTMs for both the generator and the discriminator~\cite{FGD2018}.

\item \textbf{LeakGAN} addresses the limitation of GANs regarding long text generation, i.e., more than 20 words. Its main idea is that the discriminator leaks its extracted high-level features to the generator in order to provide richer information. LeakGAN architecture contains a CNN as the discriminator, and two LSTMs as the generator. While one of the LSTMs is in charge of obtaining leaked features from the discriminator as the \emph{Manager}, the other performs the generation as the \emph{Worker}, according to the guiding goal formed by the Manager~\cite{GLCZ2018}.
\end{itemize}

\section{NLG Tasks}
\label{sec:tasks}

In order to fully appreciate the dangers that can be posed by the misuse of NLG-based systems it is crucial to not only have a broad understanding of how NLG is conducted, but also to have a good appreciation of the many tasks that NLG can be applied to. By understanding this, we can then begin to build a more coherent picture as to how these various NLG tasks can be used maliciously.

In this section, we identify the key high-level tasks of NLG and their most common subtasks, examining how these tasks are conducted and evaluated, and how they are applied in the real world. We also consider ways in which each of these tasks and their common applications may allow for deception and misuse. In particular, we focus on four core NLG tasks identified from the academic literature that have the potential for deception and misuse: \textbf{stylised text generation}, \textbf{conversation}, \textbf{rewriting}, and \textbf{translation \& interpretation}. 

\subsection{NLG Tasks and Subtasks}
 
The common high-level NLG tasks considered in our survey are defined as follows.

\begin{itemize}
    \item \textbf{Stylised Text Generation}: Stylised text generation refers to the creation of NLG systems aimed at generating \emph{original} texts in a specific, user-designated style~\cite{mou2020}. Examples of the desired style of writing are typically used as training or fine-tuning data, from which a given NLG model then attempts to automatically generate new texts that mimic the style, but not the content, of the examples. Style, in turn, is a broad term encompassing specific genres (e.g., fiction, non-fiction~\cite{AA2021,radford2019}), text purposes (e.g., rhetoric, humour~\cite{amin2020,duerr2021}), and forms (e.g., academic papers, poems, novels~\cite{mou2020}). We provide a complete overview of this task in Section~\ref{sec:Stylised_Generation}.
    
    \item \textbf{Conversation}: Conversation refers to a series of subtasks in NLG in which the broad aim is the creation of a model that can dynamically generate responses to user inputs, thus facilitating conversation in some form~\cite{motger2021}. This encompasses a number of subtasks including \textbf{task-oriented conversation}, in which the NLG system attempts to conduct a desired task through conversation with a user~\cite{zaib2020}; and \textbf{Q\&A conversation}, in which the NLG system seeks to provide the desired answer to a given, user-specified question~\cite{patil2020,zaib2021}. This task is covered in Section~\ref{sec:Conversation}.
    
    \item \textbf{Rewriting}: Rather than generating entirely new texts, rewriting tasks instead aim to leverage NLG systems that are able to reinterpret a given input text such that its underlying content and/or semantics are retained, whilst some user-specified attribute of its writing is changed~\cite{JJHVM2020}. Common subtasks include style transfer~\cite{JJHVM2020}, in which a given model attempts to retain the topic of a given text whilst changing some specified stylistic attribute (e.g., sentiment, toxicity, formality). Moreover, style transfer can be adapted to preserve user privacy by removing or otherwise obfuscating aspects of authorial style~\cite{LPSBO2021,lockwood2017}, thereby protecting the source of the original text. Other rewriting subtasks include summarisation~\cite{gupta2021}, in which the summarising model attempts to generate a shortened version of a given input whilst retaining its overall content. We cover the Rewriting task in Section~\ref{sec:Rewriting}.
    
    \item \textbf{Translation \& Interpretation}: Translation has been broadly construed to cover all NLG tasks in which a given data input (e.g., text in another language) is translated so that it is represented in natural language text. This, in turn, encompasses more than just translation from one language to another, also including other mediums, such as the translation of an image to a text caption that describes the image (which can be thought of as translating the image from a visual medium to text). In some cases, interpretation is also required, such as when a machine attempts to process puns when translating natural language, summarise an image when generating image captions, or assess the context logic of source code when generating comments.
\end{itemize}

We examine each of these high level NLG tasks; identifying their key subtasks and discussing typical approaches to them, noting key relevant datasets and examining the evaluation measures that are typically conducted. We also discuss the common applications in which these NLG tasks are used, and the manner in which these applications of NLG could lead to problems of deception or other forms of misuse.

It is worth noting that whilst these tasks and subtasks are generally implemented and evaluated distinctly from one another, they are not inherently discreet and can be utilised together in the development of a given real-world system. For instance, a conversation agent could leverage a humour module to provide it with some joke telling capabilities. Whilst stacking or otherwise combining NLG tasks is possible, the academic literature typically takes a task-centric view to NLG, focusing on tackling each task individually. 

In order to best represent the current state-of-the-art research, we thus opt to leverage this task-based focus within this survey. In turn, the following sections are dedicated to covering each of the high level tasks above, with Section~\ref{sec:Stylised_Generation} focusing on stylised text generation, Section~\ref{sec:Conversation} covering conversation tasks, Section~\ref{sec:Rewriting} examining NLG rewriting, and Section~\ref{sec:Translation} discussing translation tasks.

\section{Stylised Text Generation} 
\label{sec:Stylised_Generation}

Stylised text generation, otherwise known as style-conditioned text generation, is an NLG task focused on the automatic creation of novel texts that contain a specific, desired writing style~\cite{mou2020}. The broad nature of style  encompasses a wide variety of subtasks including story generating~\cite{AA2021}, poem generation~\cite{S2021}, humour generation~\cite{amin2020}, and the various sub-forms of each of these styles (e.g., different genres of story, different forms of poetry).

As the scope for stylised generation is only limited by the vast scope of different writing styles available, in this section we opt to focus on some of the most commonly studied forms of stylised generation in the current literature. 

\begin{figure*}[!htb]
\centering
\includegraphics[width=0.8\linewidth]{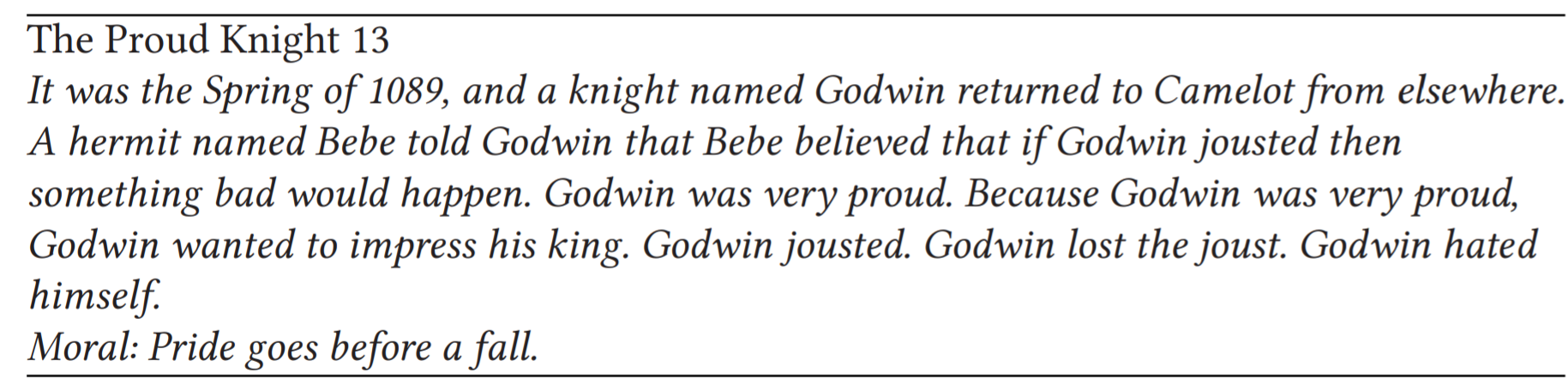}
\caption{An example story generated with a planning-based method. Retrieved from \cite{AA2021}.}
\label{fig:story}
\end{figure*}

In turn, we examine the Creative Writing subtasks, including story, poem, and lyric generation, the humour generation subtask, which focuses on joke generation, and rhetoric generation, which is a subtask aimed at creating persuasive texts. We also examine a more unconventional form of stylised generation: text augmentation, which is typically applied to boosting supervised ML performance by using stylised generation to create new training samples in the style of the existing training data~\cite{BKR2021}.

\subsection{Creative Writing}

Creative writing is a type of stylised text generation where AI and psychology intersect to teach computers how to mimic human creativity~\cite{AA2021}. This includes generating stories, poems, lyrics and prose, either from scratch or based on some existing data, such as an incomplete story, or a painting. In this subsection, we focus on one of the most commonly studied form of creative writing: story generation.

Story generation is a sub-field of creative writing where the aim is to generate stories~\cite{AA2021}. It is essentially the problem of mechanically selecting a sequence of events or actions that meet a set of criteria and integrating these together through prose writing to tell a story. The inputs of a typical story generation method include sets of events, an initial story, or a set of author goals. The model then leverages these inputs to automatically generate a coherent story.

Story generation approaches are classified into three groups: 

\textbf{Structural Models}: These models employ schemas to generate structured stories by dividing the stories into slots. Then, similar fragments of previously collected and annotated stories are placed into the new story's slots. Structural models include graph-based and grammar-based approaches based on how the annotations are used to attach the fragments. Although structural models are easy to implement, they have a couple of drawbacks. Firstly, they only consider syntax of the story despite stories being semantic models in nature. Moreover, they are limited to producing stories that satisfy the pre-defined story structure. Lastly, they can suffer from the over-generation problem, meaning that they can generate non-story texts~\cite{AA2021}. 

\textbf{Planning-based Models}: These models focus on the logical flow between the successive fragments rather than the overall structure of the story. The aim is to generate plots from the fragments, combining the fragments in a structured way to reach a story goal starting from an initial state. Planning-based approaches involve goal-directed, analogy-based and heuristic search approaches. An example output of a planning-based model is shown in Fig.~\ref{fig:story}.

\textbf{ML Models}: ML models, especially RNNs, are utilised by state-of-the-art story generation methods. They can learn the conditional probability distribution between story events from a story corpus to generate better stories. Other than generating new stories, ML models are also leveraged in other relevant tasks, including script learning and generation, and story completion.

Story generation evaluation is mostly based on assessment of quality rather than creativity. Human evaluation is the most commonly used approach although it is inflexible, time/effort consuming, and subjective. 

A typical human evaluation approach is to ask human evaluators to rate the generated stories based on common quality criteria such as consistency, coherence,
and interestingness. Another common approach is to ask evaluators to edit generated stories to make them more coherent, and calculate the story quality measure as the distance between the edit and the original story. The more edits that are needed to make the story coherent, the lower the quality of the generated story. Common edits include reordering, adding, deleting and changing events in the generated story.  

Regarding machine evaluation, \emph{Narrative Cloze Test (NCT)} is one of the most prominent approaches. It measures the system's ability to predict a single event removed from a sequence of story events by generating a ranked list of guesses based on seen events. The system is then evaluated using average rank, recall@N (the recall rate within the top $N$ guesses), and accuracy. \emph{Story Cloze Test (SCT)} is a NCT-based approach designed for supervised learning approaches. It measures the system's performance according to its ability to choose the correct ending for each story, labelled as ``right ending" and ``wrong ending". \emph{Multiple Choice Narrative Cloze (MCNC)} is another NCT-based approach where the system chooses the missing event from five randomly ordered events. Other than task-specific metrics, general NLG evaluation metrics, such as BLEU, METEOR, CIDEr and ROUGE are also commonly used for evaluating story generation methods. As highlighted in Section~\ref{sec:evaluation}, whilst these methods are useful for efficient evaluation (particularly of a large number of outputs), these also have a number of drawbacks. Firstly, they require a gold standard corpus to compare the generated text against, which can conflict with the creative nature of story generation. In addition, they typically do not correlate with human judgements, raising questions as to the relevance of their scores~\cite{AA2021}.

For story generation evaluation, the following datasets are commonly used:

\textbf{Andrew Lang fairy tale corpus}: This dataset contains more than 400 stories from Andrew Lang's Fairy Books (\url{http://www.mythfolklore.net/andrewlang}).

\textbf{ROCStories}: ROCStories is a story generation dataset containing over 98K everyday life stories, and it was constructed for use with SCT. (\url{https://cs.rochester.edu/nlp/rocstories/}).

\textbf{Children's Book Test}: This dataset was built by Facebook with freely available children's books taken from Project Gutenberg. The dataset contains over 600K stories. (\url{https://research.fb.com/downloads/babi/}).

\textbf{STORIUM}: STORIUM is a story generation dataset including 6K lengthy stories (125M tokens) with fine-grained natural language annotations, such as character goals and attributes.  (\url{https://storium.cs.umass.edu/}).

Story generation can be used for a variety of different applications, including entertainment, education, and gaming. For instance, stories can be customised for each learner's educational needs. Furthermore, interactive stories can be used to provide more interesting gaming experiences. From a deception perspective, it is possible that realistic and convincing stories can be generated to deceive and mislead people.
    
\subsection{Humour Generation}

Broadly speaking, humour generation is a stylised text generation subtask aimed at the creation of jokes~\cite{amin2020}. Humour generation finds its roots in the wider field of computational humour, which focuses on the use of computational methods in the analysis and evaluation of humour~\cite{binsted2006}. This has taken the form of a variety of subtasks including humour recognition~\cite{blinov2019,cattle2018}, the development of systems capable of learning humour preferences of users~\cite{weber2018}, the automatic evaluation of humour~\cite{barslavski2018}, as well as the development of relevant datasets and corpora~\cite{hasan2019}.

Humour generation leverages the work conducted towards computational humour and seeks to merge it with NLG to develop systems capable of automatic joke telling. Specifically, humour generation typically revolves around one of three formulations~\cite{amin2020}: (1) \textbf{Q\&A joke telling}, in which a question is posed to instigate the joke, and the NLG system attempts to generate a punchline response; (2) \textbf{narrative joke generation}, in which longer form story-based jokes are generated; (3) \textbf{lexical replacement joke telling}, in which the system attempts to replace the words in an existing text to turn it into some form of pun or joke.

In general, humour generation has received far less study when compared to other forms of stylised text generation (such as creative writing), and there is less of a clear methodology or set of approaches used in the generation of jokes~\cite{amin2020}. Moreover, most approaches fail to achieve quality humour generation that is sufficiently convincing or funny to human observers~\cite{amin2020}. Thus, humour generation as it now stands is still very much in its infancy, requiring more study before effective humour generation can be achieved. 

Currently, there stand two broad approaches to humour generation derived from the broader field of NLG: \textbf{Neural text generation}, which uses deep-learning approaches that have become commonplace in text generation more broadly~\cite{CB2020}; and \textbf{Template-based generation}, which leverages more traditional approaches to text generation in which the system attempts to choose words to fill in missing slots in an existing text template to create new text~\cite{deemter2005}.

Despite the omnipresence of neural methods in broader text generation~\cite{CB2020}, the application of neural text generation models to humour generation is less common~\cite{amin2020}. Generally, most approaches have found neural text generation to be broadly unsuited to generating humour, finding that whilst neural methods are capable of creating jokes with a high level of originality and creativity, they are typically unable to add humour or joke telling within these texts. One of the first examples of the use of neural methods for humour generation was conducted in an undergraduate project by \citet{yang2017}, who leverage an LSTM model to create jokes based on a user-specified topic. The authors trained their model on a large corpus of approximately 7,500 jokes alongside a corpus of news data to improve the model's knowledge of current affairs. In order to try and provoke a comedic response from the model, the authors attempted to promote incongruity in the generated text by having the model output words based on the probability they were assigned in the output layer, rather than the words with the highest overall probability~\cite{yan2017}. Despite these efforts, the model was generally incapable of producing humorous text~\cite{amin2020}.

One of the only other notable examples of neural-based humour generation comes from \citet{yu2018}, who focus specifically on pun generation~\cite{amin2020}. To do this, the authors aimed to maximise incongruity, training a neural network using a Seq2Seq model and Wikipedia data. The model is given as input a polysemic word (a word with multiple possible meanings), and two of its definitions~\cite{yu2018}. The model is then used to generate two sentences using this word -- one for each of the two meanings provided as input. An encoder-decoder model is then trained to generate a single sentence, based on these two sentences, which uses the input word ambiguously to allude to both meanings -- thereby creating a pun. Despite the novelty of the approach, however, this was still unable to consistently create humorous content~\cite{amin2020}. As an example, the input \textit{square: 1) a plane rectangle with four equal sides and four right angles, a four-sided regular polygon; 2) someone who does not understand what is going on.} yielded the resulting pun: \textit{Little is known when he goes back to the square of the football club}~\cite{amin2020}.

Given the current inadequacies of neural methods in humour generation, most approaches have instead focused on the use of template-based generation systems~\cite{amin2020}. Typically, a joke template is thus created, alongside a schema which encodes the relationships between the various template variables (the empty slots that the generation model attempts to fill)~\cite{amin2020}. In joke generation, this schema typically encodes relationships that provoke incongruity and resolution (key aspects of joke construction).

\begin{figure*}[!htb]
\centering
\includegraphics[width=0.78\linewidth]{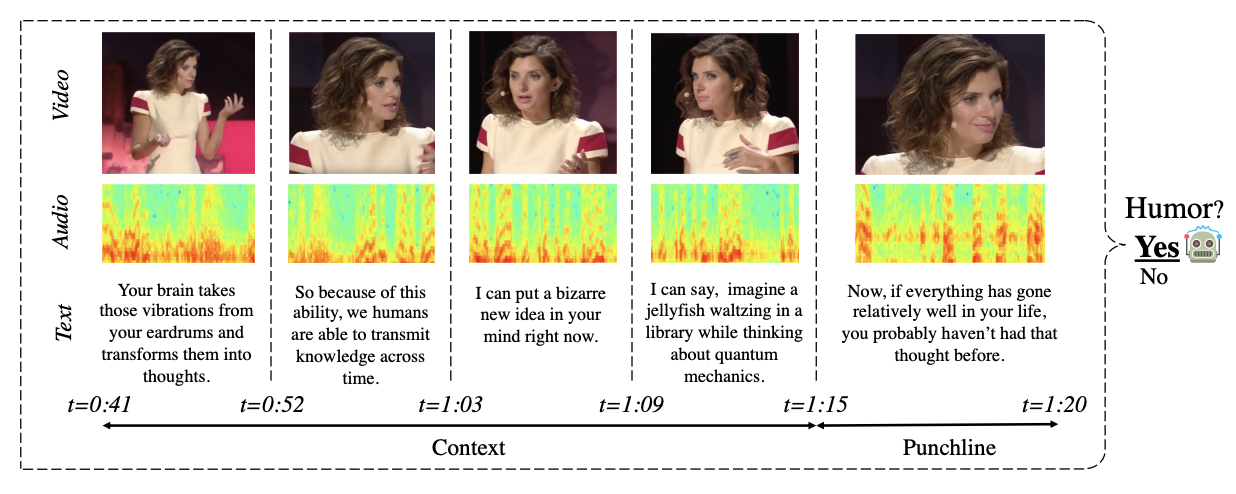}
\caption{Example from the UR-FUNNY dataset, annotated with context and punchline. Retrieved from \cite{hasan2019}.}
\label{fig:UR_Funny_example}
\end{figure*}

 Additionally, template systems rely on some form of knowledge base that provides information about the relationships between the various words that can be selected as candidates for each template variable~\cite{amin2020}. Common approaches to constructing these knowledge bases include ontology-based systems, either using manually constructed lexicons mapping each word's relationship to each other, or pre-existing ontologies and databases such as WordNet, ConceptNet, and UniSyn~\cite{amin2020}. Quantitative methods have also been proposed, which leverage probabilistic approaches such as the use of n-gram co-occurrence probabilities and vector similarity measures to select template variables that are best suited to humour~\cite{amin2020}. Whilst template-based systems are more commonly used for joke construction, and are typically more successful at generating humour, these approaches are far more limited than neural methods due to the constraints posed by the template and the lexicon from which template variable candidates are selected.

Beyond the generation process itself, efforts have also been made to develop adequate approaches to evaluating humour generating systems. Given the ambiguous and highly subjective nature of humour, however, this too has been hampered with difficulties~\cite{amin2020}. Whilst more typical NLG-based metrics for examining the coherence of a given text have some usage in assessing the overall coherence of the generated joke, these are not particularly effective in measuring the humorousness of the output. Typically, studies have relied on human evaluation as the gold-standard, using some form of Likert-scale scoring systems to assess the success of a given output at being humorous~\cite{amin2020}. Given the subjective nature of humour, however, this approach is by no means perfect -- the fact an evaluator does not find a joke funny does not necessitate that that joke is not funny, this may be a matter of their personal preference and sense of humour.

To mitigate this, other approaches have been proposed. One method is to examine humour frequency. Rather than focusing on measuring the humorousness of individual outputs, this approach asks evaluators to score a set of outputs as being \emph{funny} or \emph{not funny}~\cite{amin2020}. The percentage of funny jokes produced by the generator can then be measured. Whilst this method has benefits for measuring the overall performance of a system, it is still limited in its ability to assess the individual outputs. Other suggested approaches seek to leverage a modified version of the Turing test, in which evaluators are tasked with trying to differentiate between human and machine created jokes~\cite{amin2020}.

Although approaches to humour generation are still limited in their abilities, the broader field of computational humour has meant that a sizeable number of relevant humour datasets exist. These may become more relevant to humour generation in the near future, if state of the art neural methods that are commonplace in broader NLG are more successfully adapted to humour generation. Some examples of these humour datasets include:

\textbf{UR-FUNNY}: Created by \citet{hasan2019}, UR-FUNNY is a multi-modal dataset containing the video, audio, and transcripts from 1,866 TED talks from 1,741 speakers across more than 400 topics. The transcriptions include markers for audience behaviour, which were used to identify snippets from the talks in which a joke's punchline was told using the audience laughter marker~\cite{hasan2019}. These 8,257 punchlines, and their preceding context were then extracted and annotated (including audio and video time points) as such. 8,257 negative samples were also extracted, where the last sentence/utterance of the snippet did not end in laughter. An example from the UR-FUNNY dataset can be found in Fig.~\ref{fig:UR_Funny_example}. (\url{https://github.com/ROC-HCI/UR-FUNNY}).

\textbf{One-Liner Dataset}: The One-Liner dataset contains approximately 16,000 one-liner jokes collected using a web-based bootstrapping approach to automatically extract one-liners from a set of webpages~\cite{mihalcea2005}. (\url{https://www.kaggle.com/moradnejad/oneliners-datasets/version/1}).

\textbf{Pun of the Day Dataset}: The Pun of the Day dataset was collected by \citet{yang2015}, and contains 2,423 puns and 2,403 not-punny sentences. All puns were extracted from the Pun of the Day website, and negative samples were extracted from a variety of news sources, including AP (associated press) News, the New York Times, and Yahoo! Answer.

\begin{figure}[!htb]
\centering
\includegraphics[width=1.0\linewidth]{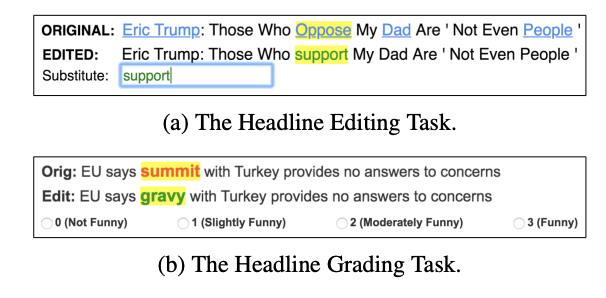}
\caption{Examples of the editing and evaluation tasks for the Humicroedit dataset. Retrieved from \cite{hossainb2019}.}
\label{fig:humicroedit_example}
\end{figure}

\textbf{Humicroedit}: A more recent dataset, the Humicroedit dataset is constructed of a series of 15,095 English news headlines posted on Reddit, paired with versions of each headline that have been edited, typically using single word replacement, to make them humorous~\cite{hossainb2019}. The editing task, along with the humour evaluation of each edited headline were conducted using Mechanical Turk (MTurk) crowdworkers. Examples of the editing and evaluation tasks are presented in Fig.~\ref{fig:humicroedit_example}. (\url{https://www.cs.rochester.edu/u/nhossain/humicroedit.html}).

Whilst humour generation is still in its infancy, there are clear applications that warrant its continued study. In turn, studies have also been dedicated to examining applications of computational humour, which have highlighted its potential value in a variety of aspects, including education -- particularly in terms of aiding students with complex communication needs~\cite{ritchie2007}, and as part of conversational systems (covered in Section~\ref{sec:Conversation}) as a means of improving user experience~\cite{iwakura2018}. However, the static nature of most humour generation approaches -- and their reliance on template-based systems -- means that current approaches are less suited to integration with downstream tasks such as education systems and chatbots where they may be of value, but where dynamic generation is needed.

The relatively poor performances of humour generators does mean, however, that their usage in deception or otherwise malicious purposes is currently quite limited. With this being said, the proposed integration of these systems with chatbots and other conversation forms of AI warrants further analysis and consideration, as this could lead to conversation systems far more capable of mimicking human dialogue~\cite{luo2019}. This mimicry, in turn, could open the door for more dangerous forms of deception as people become less capable of distinguishing whether they are conversing with a human or a machine~\cite{gros2021}.

\subsection{Rhetoric Generation}

\begin{figure*}[!htb]
\centering
\includegraphics[width=0.8\linewidth]{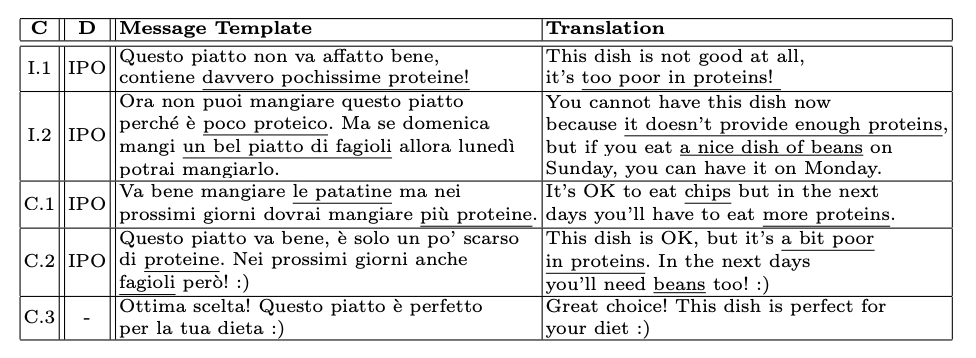}
\caption{The five templates available in \citet{anselma2015}'s diet management system. Column \textbf{C} indicates the classification of the food based on the STP reasoner (e.g., I.1 indicates the food is incompatible with the user's DRV). Column \textbf{D} indicates whether the dish is too rich or too poor in the given macro-nutrient value. Retrieved from \cite{anselma2015}.}
\label{fig:health_persuasion_template}
\end{figure*}

Rhetoric generation specifically refers to the use of NLG methods to create persuasive communications.~\cite{duerr2021}. Specifically, this involves the creation of a text output aimed at persuading an individual (or individuals) -- the persuadee -- to accept a given argument through the use of persuasive messaging embedded in the generated text~\cite{duerr2021}.

Many studies in the field of rhetoric and persuasive generation have focused on the role of various psychosocial aspects that a given NLG system will need to leverage in order to create persuasive texts~\cite{duerr2021,hunter2019}. In turn, \citet{duerr2021} identify four categories that underlie persuasive language generation. These categories are \textbf{Benevolence}, \textbf{Linguistic Appropriacy}, \textbf{Logical Argumentation}, and \textbf{Trustworthiness}.

\textbf{Benevolence}: This category refers to aspects of language aimed at creating value for the persuadee. Identifying the absence or impact of factors that may alter the perusadee's benevolence towards the argument at hand is thus of value to successful persuasion~\cite{hunter2019}. Examples in this category include example giving, appeal to morality, and the use of social proof/expectations~\cite{duerr2021}.

\textbf{Linguistic Appropriacy}: This category involves the profiling of the persuadee's lexical style in order that the persuading NLG system can leverage a style in its generation that achieves the highest degree of congruence between the generated persuading message, and the persuadee. Approaches in other NLG tasks, such as authorial style transfer~\cite{JJHVM2020}, have shown some degree of success in learning the latent styles of a target author in order to automatically rewrite texts in that author's style (see Section~\ref{subsec:style-transfer}). Examples of linguistic appropriacy include the use emphatics (pronouns like `myself', `yourself', etc.), specific word frequencies, and word familiarity~\cite{duerr2021}.

\textbf{Logical Argumentation}: This category encompasses the ability of the persuading NLG model to present a text that contains arguments with consistent logic. Some attempts have been made to create systems capable of conducting or recognising logical reasoning and argumentation through the use of first order logic and semantic argumentation graphs~\cite{block2019,moens2018}. Logical argumentation includes the use of analogies, logical operators (e.g., if, then), and logical consistency~\cite{duerr2021}.

\textbf{Trustworthiness}: The final category focuses on the capacity of the persuading system to establish trust with the persuadee. Attempts at psychological profiling as a means of identifying how this trust may be established have thus been suggested, using ML models to infer characteristics about individuals from their writing~\cite{wall2019,zarouali2020}. This is particularly important to persuasive NLG, with previous studies having identified that users often display a lack of trust when dealing with chatbots and other dialogue systems~\cite{luo2019}. Example of this include the use of agreeableness, empathy, and emotionally~\cite{duerr2021}.

Whilst some of the categories above have received a reasonable degree of focus in terms of technical implementations, particularly in regard to the automatic profiling of individuals from their writing, there are fewer works that have focused on creating NLG systems capable of persuasive generation. Additionally, these works are often spread across a wide range of domains, likely owing to the breadth of applicability of persuasive generation. Currently, there is a lack in unified approaches to creating persuasive NLG systems.

Some examples of attempts towards persuasive NLG include that of \citet{anselma2015}, who developed an NLG system to encourage or discourage a user from eating a certain food based on their chosen diet, the foods they had previously eaten that day, and the nutritional value of the dish in question. The authors thus used a simple template-based generation system, which outputs one of five pre-determined responses based on the system's evaluation of the input food. The reasoning module used to inform the template selection is built on a simple temporal problem (STP) framework, which is used to calculate whether a food is admissible based on a series of constraints derived from the user's macro-nutrient dietary recommended values (DRV). The degree to which a food is permitted based on these constraints and the degree to which it meets the user's DRV can then be used by the NLG module to select the appropriate template and template variables (see Fig.~\ref{fig:health_persuasion_template} for an example of the generated outputs).

Other attempts include the work by \citet{munigala2018}, who created a system capable of generating persuasive sentences based on a fashion product specification. This approach used a series of modules to extract keywords from the input product description via Word2Vec embeddings, before leveraging these keywords and a domain specific knowledge base to identify the most relevant domain noun-phrases using these keywords. A neural LM was then used, leveraging as input the keywords and top phrases, alongside other domain-relevant verbs and adjectives and selected persuasive verbs, to generate persuasive summaries about the product~\cite{munigala2018}. 

Given the wide range of approaches used, there also exist little in the way of established metrics for evaluating persuasiveness. This lack of standardisation in this area is also likely in part due to the difficulty in measuring persuasion, which can be highly subjective~\cite{munigala2018}. In this space, common NLG-based metrics are still typically leveraged including BLEU, METEOR, and ROUGE~\cite{celikyilmaz2020}, but these only provide a sense of the overall quality of the generated text, not of its persuasive abilities~\cite{munigala2018}. In addition, other qualitative measures have been proposed, such as catchiness (i.e., is the text catchy or not), and relatedness (i.e., is the text related to the target/argument) domain, but even these fail to offer true measurements of persuasiveness~\cite{ozbal2013}. Other studies have relied on the use of human evaluators participating as persuadees, using questionnaires to measure the degree to which the persuasive systems changed each evaluator's mind on the target subject~\cite{hunter2019}.

Given the breadth of the subject and its myriad applications, there are also a plethora of datasets that are potentially of use in the creation of persuasive NLG systems. Example of datasets focused on collecting and annotating arguments, debates, and persuasive text include:

\textbf{16k Persuasiveness Dataset}: Presented in \cite{habernal2016}, this dataset contains 16,000 argument pairs for 32 topics, where one argument is \emph{for} the topic, and one \emph{against}. All arguments were sampled from the debate portals \emph{createdebate.com} and \emph{procon.org}. MTurk annotators were then used to choose which argument was more convincing.

\textbf{Argument Annotated Essays}: This dataset is construed of 402 persuasive essays written by students on \emph{essaysforum.com}~\cite{stab2017}. The essays are written on a range of controversial topics, including ``competition or cooperation—which is better?''. These essays were then annotated with their argument components, including any major claims and premises (\url{https://tudatalib.ulb.tu-darmstadt.de/handle/tudatalib/2422}). Extending this, \citet{eger2018} translated the dataset into a variety of languages (beyond the original English), including German, French, Spanish, and Chinese. All translations were conducted using Google translate, though the authors also used native German speakers to translate a subset of 402 essays. (\url{https://github.com/UKPLab/coling2018-xling_argument_mining}).

\textbf{Debate.org Corpus}: This dataset, curated by \citet{durmus2019}, consists of 67,315 debates from \emph{debate.org} across 23 different topic categories including politics, religion, health, science, and music. The dataset includes the debate texts themselves, alongside votes provided by users indicating their preference along a variety of dimensions, including the quality of the arguments and the debater conduct. Debates are staged across a series of rounds, with one debater being \emph{for} the claim and the other \emph{against}. For each round, each debater is able to put forth a single argument.

\textbf{Persuasion For Good Corpus}: Aimed at facilitating the development of persuasive systems that can be used for social good, this dataset contains 1,017 dialogues between MTurk workers \cite{wang2019}. Pairs of participants were assigned, with one participant the persuader and the other the persuadee. The persuader was then tasked with persuading the persuadee to donate to the charity \emph{Save the Children}. All dialogues were multi-turn, with at least 10 conversational turns required. Alongside the dialogues, a number of annotations are included. These annotations record the various persuasion strategies used, e.g., logical appeal, emotional appeal, and personal stories. (\url{https://convokit.cornell.edu/documentation/persuasionforgood.html}).

Given the capacity for persuasive language generation to be integrated with a wide variety of other NLG tasks, this allows it to have a considerable range of potential applications. One of the more obvious roles is in advertising, where persuasive NLG systems could be used to generate advertising campaigns at scale, and potentially even to adapt them to individual user profiles~\cite{duerr2021}. This would offer the capacity for further scaling of existing personalised advertisement campaigns. Indeed, current work such as that of \citet{munigala2018} and their development of persuasive fashion product statements already shows indications of the potential in this area. Beyond business interests, this could also be leveraged for more positive campaigns aimed at achieving social good, such as in charity donation appeals as suggested in \cite{wang2019}. Additional suggestions have also been made for the use of persuasive generation for the advocacy of vaccinations and other medical and personal health areas~\cite{anselma2020,duerr2021}.

Beyond these broad range of applications, however, come the potential for considerable social risk. Whilst most NLG tasks have the capacity for deception and misuse, persuasive generation is particularly problematic in this regard. This is especially troubling as NLG-produced text reaches the point of being indistinguishable from human text, which could lead to difficult ethical problems in which users are unwittingly duped by persuasive text generation systems~\cite{wang2019}. The capacity of these systems to leverage personal profiling as part of their persuasion is particularly dangerous, as this could allow dishonest parties to identify those most vulnerable for persuasion, and use these powers to mislead them or convince them into doing things against their best interests. 

Moreover, issues of a machine's inabilities to tell ``right'' from ``wrong'' could lead to persuasive NLG systems inadvertently leveraging persuasive acts most would consider socially unacceptable in order to achieve its intended goal~\cite{stock2016}. Given the automated nature of many of these systems, it may be difficult, even for well-intentioned designers, to create persuasive systems capable of ethical persuasion within the confines of social acceptability~\cite{stock2016}. Additionally, the fact that individuals typically appear to have inherently negative views of systems capable of persuasion raise further ethical and moral questions in regards to their use -- especially in cases where the exact behaviours of the NLG systems are hidden from the user~\cite{stock2016}.

\subsection{Generative Text Augmentation}

Generative text augmentation is a subset of the wider field of study, data augmentation (DA). DA is the process of artificially creating new data samples through the modification of existing data samples, or the generation of new samples using existing data samples as training data~\cite{BKR2021}. This is typically used as a means of creating additional training data samples to help diversify a training set, thereby helping to reduce overfitting when training ML models -- especially in cases where the amount of `genuine' training data available is limited~\cite{FGWVCMH2021}.

\begin{figure}[!htb]
\centering
\includegraphics[width=0.6\linewidth]{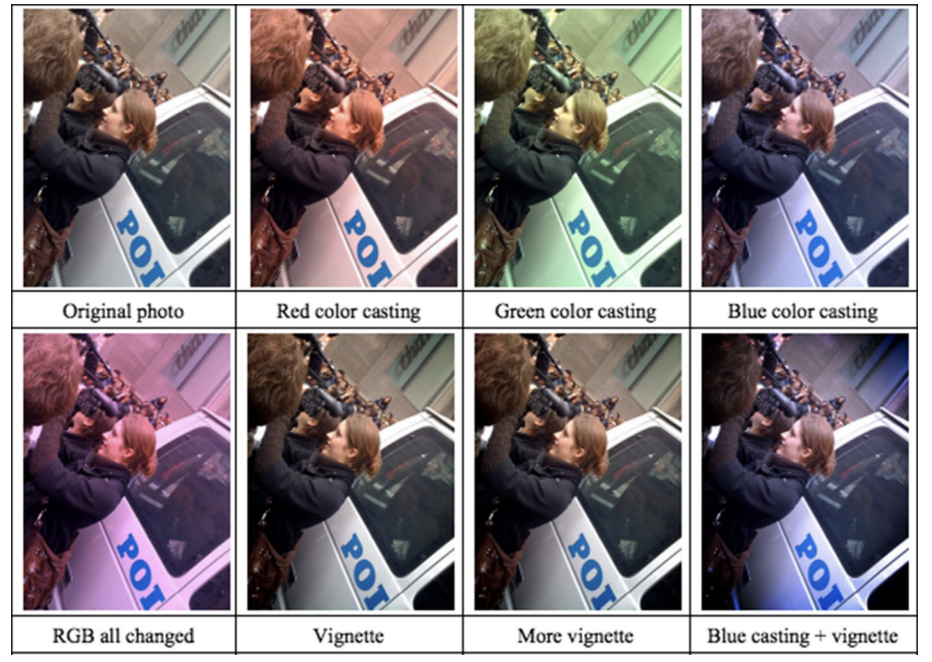}
\caption{Examples of colour augmentation. Retrieved from \cite{wu2015}.}
\label{fig:colour_augmentation}
\end{figure}

Initially, DA was almost exclusively applied to the field of computer vision (CV), using a variety of image transformation methods like cropping, flipping, and colour jittering existing images in the training dataset to create `new' images to better train CV models~\cite{shorten2019}. An example of CV DA using a colour augmentation approach can be found in Fig.~\ref{fig:colour_augmentation}. 

With the value of applying DA techniques being well established in the CV community~\cite{wu2015}, a natural extension to this is the application of DA to NLP problems, using augmentation methods to enlarge text-based datasets. The use of text augmentation poses a more difficult problem, however, as text data (characters, words, phrases, etc.) have far less granularity than image data (pixels, colours, etc.). This makes the text augmentation problem far more challenging than CV-based augmentation.

\begin{figure*}[!ht]
\centering
\includegraphics[width=0.8\linewidth]{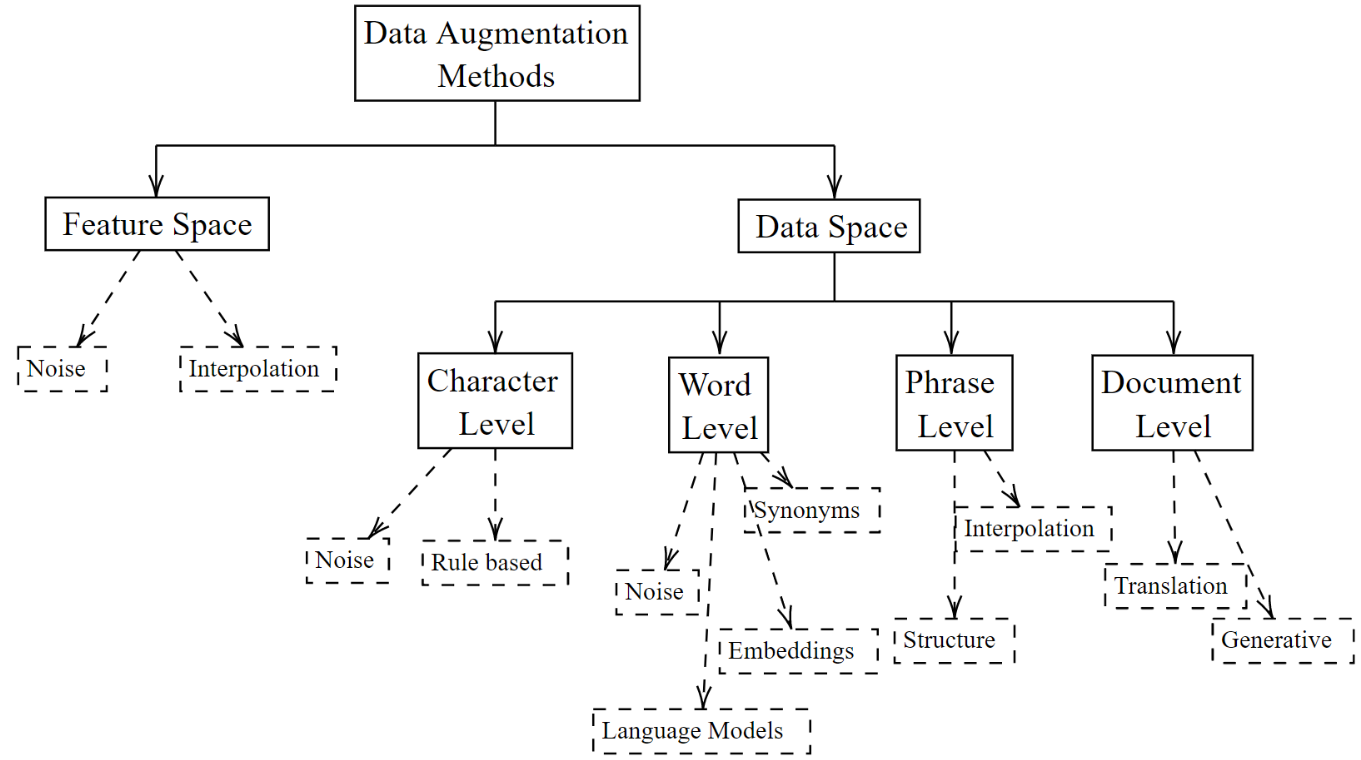}
\caption{Taxonomy and grouping of data augmentation methods. Retrieved from \cite{BKR2021}.}.
\label{fig:DA_Taxonomy}
\end{figure*}

A further challenge of text augmentation is the problem of \textit{label preservation}. When making modifications or generating new data samples for use in training ML models, it is crucial that any modified or generated samples maintain the desired training labels. In the case of text augmentation, it can often be easy through the manipulation of text data to accidentally alter the original sample label. For instance, in the case of sentiment analysis (in which a text is classified as either \emph{positive} or \emph{negative} in sentiment), a DA method that randomly inserts words could accidentally alter the intended sentiment (i.e., by randomly inserting new negative words into a positive text)~\cite{BKR2021}. Despite these challenges, a wide range of methods have been proposed for text augmentation. A taxonomy of these approaches can be found in Fig.~\ref{fig:DA_Taxonomy}.

Typical approaches used for text augmentation are widely varied in nature. More straightforward methods, such as those at the character level, often focus on introducing noise artificially, through approaches such as random character swapping within text data samples~\cite{BKR2021}. Rule-based methods instead leverage rules of grammar to make modifications to the text samples in the dataset, such as through the expanding or adding of common contractions, or through the addition of common spelling mistakes~\cite{BKR2021}.

More sophisticated approaches have also been suggested, including the use of PLMs to identify appropriate substitute replacement words in order to modify existing text samples~\cite{BKR2021}. Other approaches utilise translation tools at the document level, in which the text samples are translated to a given language, and then translated back into the original language. This has been shown to allow for reasonable degrees of paraphrasing and label preservation~\cite{BKR2021,FGWVCMH2021}.

With the demonstrated power of new neural techniques for NLG, which have shown increased capacities towards capturing specific styles~\cite{mou2020}, methods have also been proposed to leverage NLG as a form of generative augmentation~\cite{FGWVCMH2021}. A number of approaches have thus been suggested, leveraging a variety of methods including RNNs, seqGAN, and PLMs like GPT-2~\cite{BKR2021,radford2019}. In turn, these models are typically trained or fine-tuned on existing text data samples, and then tasked with generating new text samples in their `style'~\cite{anaby2020,BKR2021,wang2019b}. Through this, the generative models are able to artificially construct `new' training texts that are still representative of the class label of the original data samples.  

PLMs have shown particular promise in this area, with additional experiments being conducted to try to ensure the preservation of class labels during generation. Given the probabilistic nature of the generation process, this is essential to ensuring that the models retain the stylistics attributes of the original data sufficiently to preserve the desired class label. Examples of these approaches include \citet{wang2019b}, who opt to only use rarer instances to fine-tune their augmentation model; and \citet{anaby2020}, who use an intermediary classifier to identify text samples generated by their GPT model that retain the desired class label, before using the complete training set (the original data combined with the artificial, generated data) to train a final classifier.

Despite the promising role of generative text augmentation, and text augmentation more broadly, in developing better performing ML models -- especially in situations where data is limited, or annotation is expensive -- issues currently exist in regard to standardised approaches for DA evaluation. Currently, most approaches rely solely on the final ML model's performance as the marker for successful augmentation. If the augmented data boosts model performance, then it is considered successful. Whilst this is certainly the key marker to evaluating text augmentation methods, critics have suggested that other metrics such as resource usage and language variety warrant consideration too~\cite{BKR2021}. This is especially relevant to NLG-based approaches, which can often require large amounts of computational resources in order to run effectively. Moreover, the lack of language variety in most PLMs, which are predominately trained on English-only datasets, may inhibit their utility in text augmentation of other languages~\cite{BKR2021,FGWVCMH2021}.

Given that the core role of generative text augmentation is the boosting of ML model performance, it thus has a wide range of applications. These include a range of other NLG tasks, such as text summarisation and question answering~\cite{FGWVCMH2021}. As NLG tasks, and NLP tasks more broadly, generally require large amounts of training data, DA, and particularly generative text augmentation owing to its heightened abilities to introduce linguistic variety, have the potential to be of great value in improving model performance in a wide range of tasks~\cite{BKR2021}. Text DA may be of particular value in scenarios where sensitive or private data is needed to train a given model~\cite{BKR2021}. By using augmentation, ML engineers can reduce the amount of private data they need to gather, which could be of particular benefit to privacy preservation~\cite{FGWVCMH2021}.

\section{Conversation} 
\label{sec:Conversation}

Arguably one of the most commonly studied tasks in NLG, and NLP more broadly, conversation tasks aim to build some form of model capable of automatically generating dynamic natural language messages in response to conversational inputs by a user (or set of users)~\cite{motger2021}.  

Although often referred to by a variety of different terms, including \emph{conversational agents}, \emph{chatbots}, and \emph{dialogue systems}, these terms all encapsulate the task of developing a model capable of simulating conversation~\cite{motger2021}. 

\begin{figure}[!htb]
\centering
\includegraphics[width=1.0\linewidth]{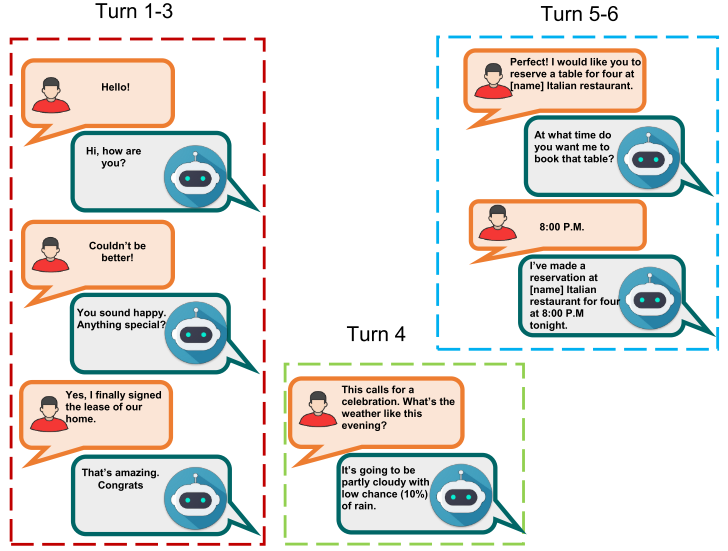}
\caption{Examples of subtasks in conversational AI~\cite{zaib2021}. Turns 1--3 depicts a chat-oriented system, turn 4 a Q\&A system, and turns 5--6 a task-oriented system. Retrieved from \cite{zaib2021}.}
\label{fig:conversation_subtasks}
\end{figure}

Given its scope, the conversation task is generally split into three, broadly distinct (though some overlaps exist) subtasks: \textbf{task-oriented conversation}, \textbf{chat-oriented conversation}, and \textbf{Q\&A conversation}~\cite{zaib2021}. Examples of these subtasks can be found in Fig.~\ref{fig:conversation_subtasks}.

Task-oriented conversation refers to the subtask of developing conversation systems capable of helping users complete a specified task, or set of tasks~\cite{zaib2020}. Chat-oriented conversation instead focuses on more broadly simulating natural conversation, typically across a wider set of domains~\cite{zaib2020}. Finally, Q\&A conversation focuses specifically on the creation of systems capable of answering user questions~\cite{alqifari2019}.

In this section, we begin by examining general approaches to the conversation task, looking at the broadly applicable design choices, modules, evaluation procedures, and datasets that are typically used in this space. We also highlight the key general applications of conversation systems, whilst also considering the role that deception could play in these use-cases. From there, we then examine the specific formulations and common approaches taken for task-oriented and Q\&A-based dialogue systems, alongside any subtask specific evaluation methods, datasets, and applications. We do not include a specific focus on chat-oriented conversations as this has been less well studied, with current approaches generally relying on expanding task-oriented methods of developing conversational systems.

\subsection{General Approaches}
\label{subsec:ConversationGeneral}

\begin{figure*}[!htb]
\centering
\includegraphics[width=0.9\linewidth]{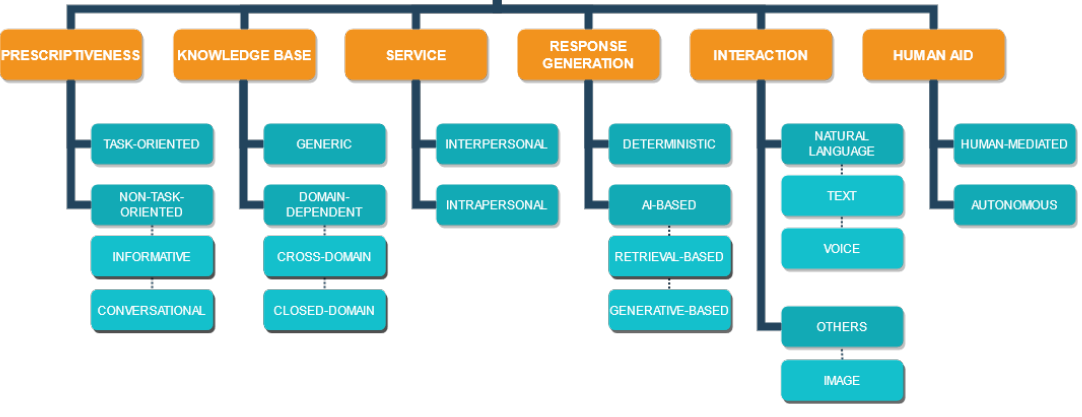}
\caption{Breakdown of the various design dimensions used when developing conversational agents. Retrieved from \cite{motger2021}.}
\label{fig:conversation_system_design}
\end{figure*}

Broadly speaking, all conversational models can be centred around a series of general, common goals. The model's ability to achieve these goals (or a subset of them) will, in turn, allow it to more or less successfully simulate conversation. These are:

\textbf{User Support}: Probably the most common goal of conversation agents, this goal focuses on the ability of the conversation system to support the user in whatever application it is being used in~\cite{motger2021}. This support could come through assisting the user in achieving a set task, or by retrieving information that the user requests~\cite{bell2019}.
    
\textbf{Information Request}: Moving beyond user support, information request refers specifically to the goal of retrieving specific information that a user desires. Whilst clearly relevant to the Q\&A conversation subtasks, some degree of information request ability is still typically needed in the construction of other forms of conversational system. Given the central nature of knowledge sharing and question asking in more natural forms of conversation, most conversational systems require some ability towards identifying and returning specific information. This is typically leveraged through the curation of specific knowledge databases, either more broadly defined or specific to a set of relevant domains~\cite{motger2021}.
    
\textbf{User Engagement}: Also central to most conversation systems, particularly those that facilitate conversation across multiple conversational turns, is the ability of the conversational agent to adequately interest and engage the user it is conversing with~\cite{motger2021}. This overlaps with the common human-computer interaction (HCI) aspects of developing chatbots, and is crucial in developing systems that are capable of `natural' conversation~\cite{chaves2019}.
    
\textbf{Information Collection}: A final key goal of most conversational systems is information collection. Whilst this is often specifically formulated towards the collection of key pieces of information about a user (e.g., in a task-oriented systems aimed at providing medical diagnoses~\cite{gentner2020}), some degree of information collection is essential in most conversational systems~\cite{balaraman2021}. Through information collection, the conversational system is better able to build a profile of the user it is chatting with, allowing for the creation of more personal and relevant responses and utterances~\cite{motger2021}.

Regardless of the specific subtasks, the development of quality chatbots involves the implementation of many modules~\cite{zaib2020,zaib2021}, all of which need integrating in order to create an effective conversational agent. This moves beyond basic NLG, to consider natural language understanding (NLU) to comprehend user inputs~\cite{louvan2020}, information retrieval (IR) to extract and generate relevant responses~\cite{zaib2020}, intent classification to understand user intent~\cite{louvan2020}, and myriad other tasks besides~\cite{motger2021}.

In order to design a given conversational system, considerations have to be given to a series of design dimensions; namely, the prescriptiveness of the system, the knowledge base used, its intended service, the form of response generation used, the mode of interaction, and the degree of human-aid to be leveraged. Each of these design dimensions will require implementation and integration into the overall architecture of the chatbot itself. A breakdown of the most common design dimensions can be found in Fig.~\ref{fig:conversation_system_design}.

\textbf{Prescriptiveness}: This design element refers to the intended subtasks(s) that the chatbot aims to achieve. As described earlier, these can be divided into task-oriented systems, chat-oriented conversational systems, and informative Q\&A systems~\cite{zaib2021,motger2021}.
    
\textbf{Knowledge Base}: This dimension describes the manner and form by which the knowledge base used by the conversational agent is implemented. The knowledge base is a key element, as it will impact the model's ability to respond to inputs regarding different topics, and the choice of knowledge base typically reflects the intended use of the conversational system~\cite{zaib2021}. Generic knowledge bases contain data that is non-specific to a certain topic or set of topics, and aim to encapsulate a wide range of (typically general) knowledge~\cite{motger2021}. Domain-specific knowledge bases, instead, contain data relevant to a specific set of subject areas only~\cite{motger2021}. Cross-domain knowledge bases contain information from a small set of topics, and closed domain a single subject area~\cite{motger2021}.
    
\textbf{Service}: This design element defines how the conversational agent interacts with the user. Interpersonal systems are more generic, and do not build any specific relationship with each user it interacts with~\cite{motger2021}. Intrapersonal systems, instead, aim to build context and user dependent profiles that inform its response generation~\cite{motger2021}. 
    
\textbf{Response Generation}: This design choice informs how the system will go about building the desired response to a given conversational input. Deterministic systems use prescribed structures to link the input data to a relevant (often pre-written) response~\cite{alqifari2019,zaib2021}. This includes template approaches, in which the system selects the relevant key terms to fill in blank spaces in a pre-written response. Retrieval-based systems instead leverage ML methods to predict relevant responses from a set of pre-defined responses -- this is often formulated as a typical ML classification problem, where each response is the equivalent of a given class~\cite{motger2021}. Generative approaches, instead, leverage NLG -- typically through the use of deep learning models -- to dynamically generate responses based on the user input~\cite{zaib2020}.
    
\textbf{Interaction}: This element refers to the manner in which the chatbot system interacts with users. Typically, this will be via natural language through text, but voice responses are also possible -- as seen in the recent rise in virtual assistants such as Amazon  Alexa~\cite{motger2021,tiwari2020}.
    
\textbf{Human Aid}: Whilst generally less considered, this refers to the level of autonomy given to the chatbot. Whilst most research has focused on the development of autonomous conversational agents, it is also possible to develop human-in-the-loop systems, in which human assistance is leveraged to aid the conversational system~\cite{motger2021}. 

In turn, these design considerations will determine the intended task that the conversational agent is developed toward, the manner in which it will be implemented to achieve the task (including the means by which it generates response), and the way in which its interactions will be developed.

Beyond the design of the system itself, it is also essential that consideration is given to the creation of the evaluation approaches that will be used to assess the performance of the conversational system in question. Whilst many of the evaluation procedures used are specific to each subtask (measuring its ability to achieve that subtask specifically)~\cite{zaib2020,zaib2021}, there are also more general evaluative measures that are often implemented to measure chatbot quality. In turn, these evaluation strategies often mirror the broader approaches taken to evaluate NLG systems (as discussed in Section~\ref{sec:evaluation}).

A common means of evaluating the abilities of a conversational system is through the quality criterion approach~\cite{howcroft2020}. Through this, the conversational system is scored based on how well it meets a defined quality criterion or set of quality criteria~\cite{motger2021}. These are typically focused on the specific text output, measuring how well it meets human standards of natural language. Common criteria include \emph{fluency}: how well the intended language is mimicked~\cite{celikyilmaz2020}; \emph{factuality}: how logically coherent and `true' the response is~\cite{celikyilmaz2020}; and \emph{typicality}: how likely it is you'd expect to see a response of this nature from a human author~\cite{howcroft2020}. These criteria are typically assessed through human evaluators, leveraging some form of scale-based scoring system (e.g., Likert Scales)~\cite{celikyilmaz2020}.

In the case of conversational systems, these quality criteria are typically characterised broadly into \textbf{functional suitability}, \textbf{efficiency}, \textbf{usability} and \textbf{security}~\cite{motger2021}.

\textbf{Functional Suitability}: This set of criteria encapsulates how correct and appropriate the system's outputs are. Correctness overlaps most strongly with broad NLG quality criteria, measuring the overall ability of the system to create convincing responses of a reasonable quality. Appropriateness, then measures the degree to which the content of these responses is appropriate, given the user's input and the desired tasks to which the conversational agent is being used towards.~\cite{motger2021}
    
\textbf{Efficiency}: Less relevant to the model's response generation, this examines how effectively the conversation system manages its resources, and how quickly it can generate responses to user input~\cite{motger2021}.
    
\textbf{Usability}: This set of quality criteria refers to the ease with which users can interact with the systems. This and efficiency (above) are key HCI considerations when developing conversational systems, though are less related to the model's performance in terms of generating `correct' responses~\cite{chaves2019}.
    
\textbf{Security}: These quality criteria measure the degree to which the conversational agent is capable of protecting user privacy, and is resistant to malicious interaction. This includes data management considerations such as how personal data is stored, but also encompasses how `trustworthy' the conversational agent is. As these chatbots typically act autonomously, it is crucial that users can trust that the responses they generate will be appropriate and reliable, and not mislead them in some manner.

As some of these quality criteria are more widely conceived than the more general NLG-based criteria, often requiring broader considerations of the overall behaviour and/or performance of the chatbot systems in question, more dynamic evaluation approaches are often favoured. These almost always necessitate a reliance on human evaluation (as opposed to automated metrics), with interviews, broader questionnaires and focus groups all being typical in evaluating a given conversation agent~\cite{motger2021}.

\begin{figure*}[!htb]
\centering
\includegraphics[width=0.95\linewidth]{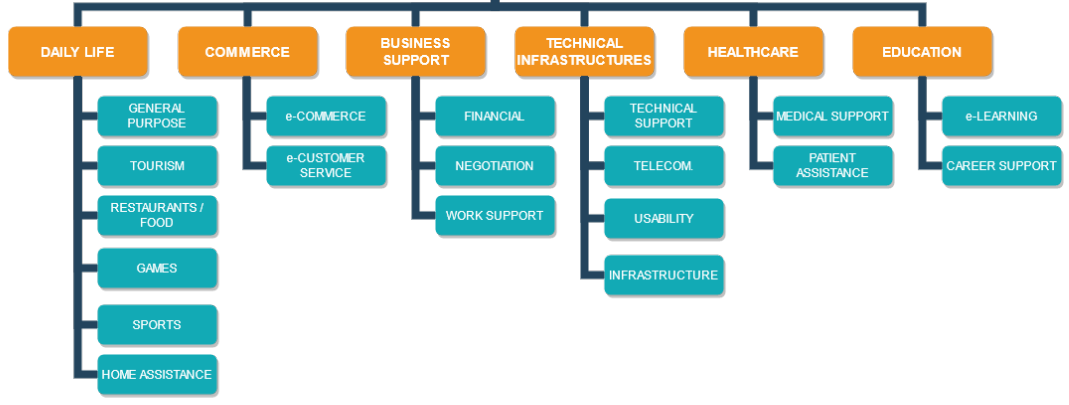}
\caption{Breakdown of the common application domains relevant to the development of conversational systems. Retrieved from \cite{motger2021}}
\label{fig:conversation_applications}
\end{figure*}

It is worth noting that quantitative and more automated metrics are still commonly used, but these are more often leveraged in measuring a system's performance in a given subtask (e.g., accuracy in a Q\&A setting~\cite{zaib2021}), or to measure the performance of individual modules within the conversational system's hierarchy~\cite{balaraman2021}. For instance, accuracy might be leveraged to measure the quality of the dialogue state tracking module, a common element in many chatbot systems that monitors the conversation history to help inform the chatbot's response~\cite{balaraman2021}.

Given the scope of conversational AI, including its broad potential for real-world application, and the variety of design dimensions needed to create fully-fledged conversational agents, there exists a wide array of datasets developed for both training and evaluating dialog systems. We present a few of the most popularly used below:

\textbf{The Multi-Domain Wizard-of-Oz (MultiWoz) Dataset}: The MultiWoz dataset includes a set of human conversations across a wide range of domains and topics~\cite{multiwoz2018}. These domains include hotel, restaurant, police, and hospital. The dataset includes multiple conversational turns, and multi-domain conversations. The dataset also includes annotated dialogue acts indicating the intents and slot-value pairs of a given piece of dialogue. For instance, the dialogue act \textit{INFORM(domain=restaurant,price=cheap)} indicates an intent to inform, with slots for the `domain' and the `price', and values of `restaurant', and `cheap'~\cite{multiwoz2018}. The annotations can then be used to evaluate the performance of the intent classification and slot-value prediction of a given conversational agent. This ability is often essential to the agent's ability to generate a relevant response or accomplish a given task. (\url{https://github.com/budzianowski/multiwoz}).

\textbf{Ubuntu Dialogue Corpus}: The Ubuntu Dialogue Corpus is an unlabelled dataset containing almost 1 million two person conversations in English extracted from Ubuntu's chat-logs, with more than 7 million utterances being recorded~\cite{ubuntuDialogueCorpus2015}. This corpus is generally leveraged by LM-based conversational systems capable of utilising vast amounts of unlabelled text data. (\url{https://github.com/rkadlec/ubuntu-ranking-dataset-creator}).

\textbf{The Naval Postgraduate School (NPS) Chat Corpus}: The NPS Chat Corpus contains more than 10,000 English posts extracted from a variety of online chat services~\cite{NUSCorpus}. The dataset has also been annotated with part-of-speech (PoS) tags and dialogue speech acts~\cite{NUSCorpus}. (\url{http://faculty.nps.edu/cmartell/npschat.htm}). 

\textbf{OpenSubtitles}: The OpenSubtitles dataset is a multilingual dataset containing the subtitles for a wide range of movies and TV programs in more than 60 different languages~\cite{opensubtitles2016}. The dataset contains more than 300 million pieces of dialogue, and also captures multi-person dialogue -- with an average of 2 to 6 speakers per script~\cite{mahajan2021}. (\url{https://opus.nlpl.eu/OpenSubtitles-v2018.php}).

\textbf{British National Corpus (BNC)}: The BNC is an multi-party English corpus created by Oxford University Press in the 1980s, making it one of the oldest corpora of its kind~\cite{BNC1992}. The dataset contains more than 800 pieces of dialogue across a wide range of genres and domains, including transcribed speech, fiction, magazines, and newspapers. The dataset also includes basic PoS tagging~\cite{mahajan2021}. (\url{http://www.natcorp.ox.ac.uk/}).

Due to their ability to be leveraged in both open-domain contexts and subject-specific areas, and the value of automating dynamic user interactions in a wide range of fields, chatbots have been proposed for a vast array of general applications. An overview of some of the common application domains for conversational agents can be found in Fig.~\ref{fig:conversation_applications}.

This has led to the proposal of chatbots of various kinds in specific industries, including tourism (e.g., flight booking, holiday planning)~\cite{motger2021,zaib2021}, and the restaurant industry (e.g., restaurant finding, restaurant booking)~\cite{motger2021,zaib2020}. Chatbots have also been more generally proposed as facilitator of customer service and technical support in commerce more broadly~\cite{cui2017,xu2017}, with different subtask-based conversational systems having the potential to be applied here~\cite{zaib2020,zaib2021}.

Extending this, chatbots have also been proposed as facilitators of workplace support, offering technical support and/or assistance with financial tasks, negotiation and teamwork~\cite{motger2021}. By automating these support tasks through the use of a chatbot-based system, businesses can aim to make these interactions more natural for employees, increasing the ease with which they can leverage these forms of assistance~\cite{motger2021}.

Healthcare is another area of application in which chatbot-based systems have been proposed~\cite{kowatsch2017,jovanovic2020}. Beyond previously mentioned applications of general customer support, chatbots have also been suggested as support tools for healthcare professionals, assisting with prescriptions and diagnoses. Chatbots have also been suggested as being of value in aiding patients, with therapy-based conversation systems and conversational symptom-checker systems being proposed~\cite{jovanovic2020,motger2021}.

A final, common application domain is that of education~\cite{hobert2019,perez2020}. Chatbots have thus been proposed as a means of answering student's FAQs, providing e-tutoring to students, as well as being suggested as a form of automated careers advice counselling and as assistants for accessing and searching institution libraries~\cite{motger2021,perez2020}. 

Due to its broad capability for application to the real-world, its direct interaction with users, and the typical need for the chatbot to profile each user in some manner, there are many fears in regard to the potential capabilities toward deception and misuse that may come with developing real-world chatbot systems. Moreover, given that many modules and sub-systems need to be integrated together to develop the conversational agent, this opens up the potential for weaknesses to be exploited throughout a given chatbot's architecture. 

\begin{figure*}[!ht]
\centering
\includegraphics[width=0.92\linewidth]{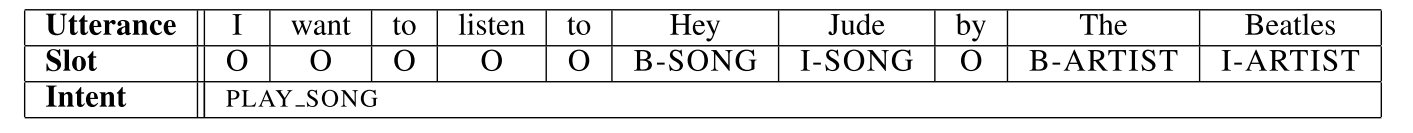}
\caption{Output of the slot filling and intent classification tasks conducted by an NLU module, for a given input dialogue. Slot labels are in BIO format, where B indicates the start of a slot span, I the inside of a span, and O a word that does not belong to a slot. The NLU attempts to predict the correct intent label and slot labels for the given utterance. Retrieved from \cite{louvan2020}.}
\label{fig:NLU_diagram}
\end{figure*}

Additionally, the dynamic nature of the system, in which it needs to respond to a range of (sometimes hard to predict) user inputs, means that it can be hard to gauge how chatbot systems might react in certain scenarios. This can make it difficult for users to truly place their trust in these systems, whilst also raising difficult questions in regard to the protection of user privacy when interacting with these systems. This is particularly problematic, given the proposed use of chatbots in a variety of sensitive applications, including education and healthcare.

Examples of these potential issues include out of domain problems, in which the chatbot is asked to engage with a user about a subject that is outside its knowledge base~\cite{ye2020}. In this case, the chatbot may respond with information that is misleading or otherwise false, which could inadvertently deceive or trick the user in some manner~\cite{ye2020}. If we consider the case of a healthcare symptom checker system being asked about an illness it lacks requisite knowledge of, we can see how this could be potentially dangerous. These out of domain issues could also be leveraged by malicious users, who could utilise the lack of knowledge of a chatbot in a given area to fool users into unknowingly divulging private information~\cite{ye2020}.

Other dangers include the potential for chatbots systems to produce undesired responses that are derogatory or offensive, when presented with certain inputs~\cite{ye2020}. This has been noted as particularly problematic for generative chatbot systems that leverage powerful PLMs. Since these PLMs leverage vast amounts of (typically web-based) data, distinct biases and tendencies towards extreme language have been noted~\cite{bender2021}. It is thus possible that in certain situations, chatbots may provide responses to users that reflect the biases in its training data and/or knowledge base, which could disrupt user trust or potentially even lead to the user being misled or biased in some manner~\cite{ye2020}.

Additionally, issues exist in regard to user privacy concerns, particularly in regard to divulging personal data~\cite{saglam2021}. The use of chatbots has led to the potential removal of user agency over their data, with users typically having limited understanding over how personal information divulged to a chatbot might be leveraged, with users typically having limited recourse to delete this data in future~\cite{saglam2021}. Again, given the proposed use of chatbots in sensitive applications where personal data disclosure is likely necessary, such as in healthcare, this is a pertinent problem~\cite{jovanovic2020}. Moreover, given the often generative nature of most chatbots, it can be hard to ensure that the conversational agents do not inadvertently request information from the user that they do not wish to provide -- an issue that could be particularly problematic in out of domain scenarios~\cite{ye2020}. It is also difficult, given the lack of control over user inputs, to ensure that users do not inadvertently surrender more private data than is intended~\cite{sauglam2020}. Given that some chatbots may integrate user inputs into their training data or knowledge base to improve their performance, this could lead to issues of chatbots unintentionally leaking user data in future conversations~\cite{sauglam2020,ye2020}.

\subsection{Task-Oriented Conversation Systems}

Task-oriented conversation systems are a conversation subtask aimed at creating conversational agents designed to perform a set of desired tasks on behalf of a user~\cite{zaib2021}. These systems therefore rely on conversation with the user in order to gain a sense of their desires and preferences towards a given task, before using this knowledge to perform the task in the way the user wishes~\cite{zaib2021}.

At their core, most task-oriented conversation systems are constructed around three separate modules: the Natural Language Understanding (NLU) module, the Policy Learning module, and the Response Generation module~\cite{balaraman2021,louvan2020,zaib2020}.

The NLU module is responsible for the initial processing of a given utterance or input text from the user~\cite{louvan2020}. In turn, the core function of the NLU is to provide intent classification and slot-value prediction for the given piece of dialogue (See Fig.~\ref{fig:NLU_diagram} for an example). This, in turn, can be integrated with a dialogue state tracking (DST) module (typically as a joint model), which allows the NLU to leverage previous utterances from the user in its prediction of the overall intent of the current dialogue presented to the conversational systems. The range of possible intents and slot-values are generally encoded in a pre-defined ontology which is typically closed-domain -- using domains relevant to the desired task the conversational system is used to conduct. 

From this, the Policy Learning module is used to decide what action is to be taken by the conversational system, in order to guide the user to a specific task, leveraging the predicted dialogue states achieved by the NLU module~\cite{louvan2020,ilievski2018}. 

Finally, the generation module is used to create a response based on the predicted dialogue state and the chosen policy. This is generally implemented either through a template-based approach, in which the system identifies an appropriate response template and template values, or (less frequently) using a probabilistic generative method (such as through the use of powerful generative PLMs like GPT-2)~\cite{louvan2020}. 

In order to train and evaluate task-oriented conversational systems, many of the datasets mentioned in Section~\ref{subsec:ConversationGeneral} have been popularly used. This is especially true of the MultiWoz dataset, as it comes annotated with dialogue acts that are particularly useful for evaluating the NLU components of the conversational system~\cite{multiwoz2018}. As a large proportion of the work on task-oriented systems has been dedicated to its NLU aspects, there also exist a range of datasets aimed at specifically evaluating task-oriented intent classification and slot-value prediction:

\textbf{Airline Travel Information System (ATIS) dataset}: The ATIS is a highly popular, single-turn dataset used for benchmarking NLU modules~\cite{ATIS1990,louvan2020}. This dataset is composed of approximately 5,000 utterances focused on airline travel, e.g., queries focused on flight searching. Each utterance is annotated with the appropriate slot and intent labels needed to evaluate the accuracy of a given NLU system in this domain. (\url{https://www.kaggle.com/hassanamin/atis-airlinetravelinformationsystem}).

\textbf{MEDIA}: The MEDIA dataset is focused around hotel booking scenarios, containing a series of simulated conversations in French between a tourist and a hotel concierge~\cite{MEDIA2005}. The dataset contains approximately 18,000 utterances, and is labelled with slots (intent labels are not provided) including the number of people, the date, and the hotel facility~\cite{louvan2020}. (\url{https://catalogue.elra.info/en-us/repository/browse/ELRA-S0272/}).

\textbf{Snips Dataset}: This dataset was curated by crowdsourcing spoken conversation using the Snips voice platform~\cite{SNIPS2018}. The conversational data was generated by using Amazon Mechanical Turks (MTurk) and other crowdsourcing platforms to create artificial utterances based on a provided set of intents and slots. A variety of domains were used for the intent-slot sets provided, including restaurant bookings, movie schedule requests, and song playing requests. (\url{https://github.com/sonos/nlu-benchmark}).

\textbf{Facebook Multilingual Dataset}: This dataset attempts to address the lack of non-English data by curating a multilingual dataset of task-specific dialogues~\cite{schuster2018}. This dataset thus contains 57,000 dialogues, including 8,600 Spanish utterances and 5,000 Thai utterances across a variety of domains including weather, alarm, and reminder. All dialogues are annotated with both intents and slots. (\url{https://ai.facebook.com/blog/democratizing-conversational-ai-systems-through-new-data-sets-and-research/}).

\textbf{Dialog State Tracking Challenges (DSTC) 2 \& 3}: DSTC 2 \& 3 are two English datasets aimed at specifically evaluating DSTs~\cite{DSTC2,DSTC3}. These datasets are composed of human-machine conversation related to both restaurants and tourism. DSTC2 contains over 3,000 dialogues (1,612 training dialogues, 506 development dialogues, and 1,117 testing dialogues), and is labelled  with the turn-level semantics of each dialogue, which the given DST attempts to predict via the current dialogue and dialogue history to that point. DSTC3, on the other hand, is aimed at assessing the ability of DSTs to predict slot-values in unseen, out of domain situations, and thus only contains 2,265 dialogues for testing. (\url{https://github.com/matthen/dstc}). 

As is the case with most datasets in NLP, English is the predominant language in dataset curation for task-oriented conversation systems. However, there have been attempts in recent years to develop datasets for additional languages. Beyond the MEDIA and Facebook datasets mentioned above, the popular ATIS dataset has been translated into a variety of different languages, including Hindi, Turkish, and Indonesian~\cite{susanto2017,upadhyay2018}, as well as, through the MultiAtis++ dataset~\cite{xu2013}, into Spanish, Portuguese, German, French, Chinese, and Japanese. The SNIPS dataset has also been adapted to Italian by \citet{bellomaria2019}.

\begin{figure*}[!htb]
\centering
\includegraphics[width=0.8\linewidth]{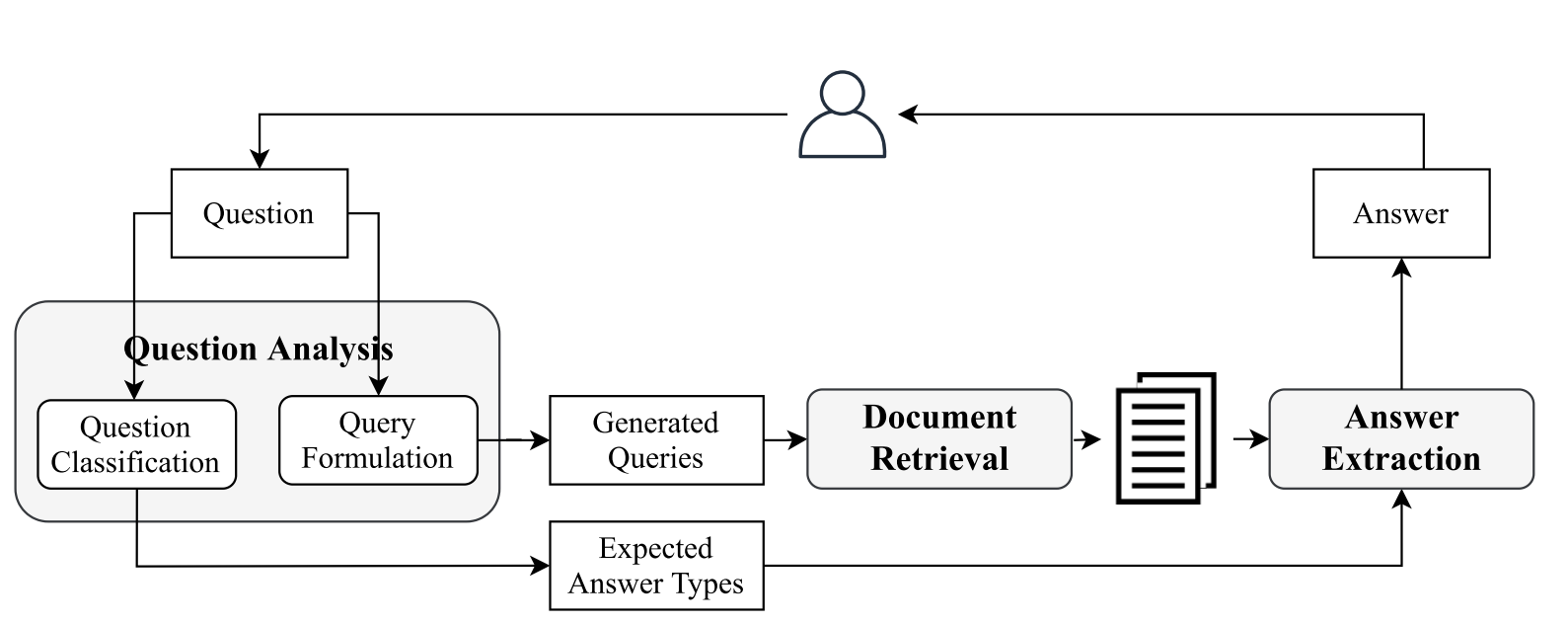}
\caption{Typical architecture of a single-turn Q\&A system. Retrieved from \cite{zhu2021}.}
\label{fig:single-turn_QA}
\end{figure*}

Due to its task-oriented nature, these type of conversational system have been popularly leveraged in a range of applications, especially as a means of improving booking systems for a range of businesses. In turn, task-oriented conversational systems have been proposed as a means of aiding in hotel booking, the reservation of restaurants, and for holiday bookings. Given the very specific domain of these tasks, it has been more straightforward to implement systems capable of extracting the relevant intents and slot-values from these forms of conversation in order to extract enough information from a customer to facilitate the booking. Other applications have also been proposed, including the usage of task-oriented systems in online shopping, in which the system can be leveraged as a means of finding relevant products, as well as an interface in which purchases can be made~\cite{yan2017}. 

Task-oriented nature is also inherently related to the development of virtual assistants (such as Alexa, Siri, etc.), as these systems are inherently tasks oriented by design. Through this, a variety of task-oriented applications are encompassed, including (beyond the above) tasks such as alarm and reminder setting, and weather checking.

Given these applications, task-oriented systems are thus vulnerable to the wide range of deceptive or otherwise malicious actions discussed in Section~\ref{subsec:ConversationGeneral}. Out of domain issues pose a particular problem here, as task-oriented systems are typically narrowly focused on one, or a small set of domain-specific tasks. This could lead to the task-oriented system attempting to perform actions it is not capable of doing, or of performing tasks in an undesired manner. Privacy issues are also problematic in this case, especially when the applications of the systems involve the parsing of sensitive user data. Applications such as holiday booking or product purchases bring risks as this may involve the task-oriented conversation system having to handle user payment data. Thus, issues of adequate storage of this data by the conversational systems also exist, as poor security here could result in the leaking of this private data.

\subsection{Q\&A Systems}

Q\&A systems are centred around the development of dialogue systems capable of answering user questions. These systems can be formulated either as closed-domain, in which the questions are related to one, or a small subset, of topics; or open-domain, in which questions can be drawn from a wide range of topics~\cite{zaib2020,zaib2021}. To some extent, Q\&A systems can be viewed as a highly specific version of task-oriented systems (indeed, question answering is often part of a task-oriented system's overall task), but the approaches taken to developing Q\&A systems typically differ from those of task-oriented system, often relying more heavily on comprehension and information retrieval rather than response generation~\cite{alqifari2019}.

Beyond the classes of open-domain and closed-domain, Q\&A systems can also be either single-turn, or multi-turn~\cite{zaib2021}. Single-turn systems are required to provide an answer based on a single dialogue input from a user, whereas multi-turn (also called conversational Q\&A) systems allow for multiple dialogue turns and questions -- where the Q\&A system will typically have to rely, in some part, on the dialogue history to answer any questions posed.

For single-turn Q\&A, the most common approach is to try and identify the answer to the question within some form of knowledge base. For closed-domain tasks, this will typically be a dataset, or datasets, specifically relevant to these specific topics, whereas in an open-domain setting, these will typically be large knowledge bases such as Wikipedia. The common modules for a single-turn Q\&A system are the \textbf{Question Analysis} module, the \textbf{Document Retrieval} module, and the \textbf{Answer Extraction Module}~\cite{zhu2021}. Alongside the question, the system will often also be provided with some form of context to aid in answering the question~\cite{zhu2021}. Figure~\ref{fig:single-turn_QA} provides an overview of a common single-turn Q\&A architecture. 

The Question Analysis module is generally tasked with predicting the correct question class in order to inform the structure of the answer to be generated, and formulating the optimum query by which the knowledge base's documents can be searched for the answer. For question classification, common question classes include factoid questions, e.g., how, why, where; confirmation questions, e.g., yes, no; and listing questions, i.e., listing items in a given order~\cite{zaib2021}.

The Document Retrieval module is then tasked with using the query created by the Question Analysis module to identify the document or subset of documents within the knowledge base that are most likely to contain the answer to the question~\cite{zhu2021}. This is typically treated as a form of information retrieval task (IR). Common approaches include Boolean models, in which Boolean expressions are used to match questions to documents and vector models, in which vector representations and similarity metrics are leveraged to find the most relevant documents~\cite{zhu2021}. LM-based approaches have also been suggested, which rank documents based on the probability of the model generating the question given the document -- which is used to fine-tune the model~\cite{zhu2021}.

After the most relevant document has been identified, paragraph ranking is then conducted to find the paragraph within the candidate document that is most likely to contain the answer to the question. One of the most popular approaches to achieve this is through learning to rank (L2R)~\cite{huang2020}, which utilises supervised learning approaches to identify the optimum ranking of candidate paragraphs relative to the question. 

The Answer Extraction module is then leveraged to identify the answer, within the candidate paragraph, to the user's question. This is often approached via span prediction, in which the text span in the candidate paragraph that contains the answer is identified~\cite{huang2020}. Many approaches exist to do this, though the use of PLMs such as BERT have become especially popular in recent years~\cite{huang2020}. NLG-based approaches for selecting the answer using text generation rather than span selection have also been proposed, those these are less common and are typically unable to achieve the performances recorded by span-selection-based models~\cite{zhu2021}.

For multi-turn systems, approaches still typically conceive of the Q\&A problem as one of answer span selection from a knowledge base of documents~\cite{zhu2021}. The key challenge presented then is one of dialogue history modelling in order to best leverage any historical information in past utterances that can be used to identify the most relevant answer-span~\cite{zaib2021}. To do this, a history selection module is typically incorporated to select the past utterances most relevant to answering the question. This is generally either done through a $k$-turn-based approach, in which the past $k$ number of utterances are used, or dynamic selection where a trained model is used to dynamically assess the contributions of each utterance to answering the question~\cite{zaib2021}. Some form of history encoding and modelling is then used to integrate the historical dialogue with the current question. Commonly used approaches include conventional word embeddings, and contextualised word embeddings leveraging PLMs~\cite{zaib2021}.

Evaluation of Q\&A-based system is generally conducted via the use of some form of common accuracy metric such as F1-score to measure the degree to which the Q\&A systems correctly identifies the answer to the question~\cite{zhu2021}. For generative answering approaches, common automated metrics such as BLEU have also been leveraged to measure the quality of the answer generated~\cite{zhu2021}. A further metric that has been proposed is Human equivalence score (HEQ), which measures a system's performance relative to that of the average human~\cite{quac2018}. 

The metrics used are typically tied into the specific dataset used to train and evaluate the Q\&A system. Some of the most commonly used datasets in this subtask are:

\textbf{SQuAD}: The Stanford Question Answering Dataset (SQuAD) is a dataset containing more than 100,000 questions posed by crowd-workers on a specific set of Wikipedia articles~\cite{squad2016}. This takes a span-finding approach to Q\&A, where the annotated answer to each question is a specific segment of text in one of the Wikipedia articles. Alongside each question, the relevant passage is also provided to the system as context. Extensions to this dataset include SQuAD 2.0, which expands the dataset to include unanswerable questions (which the system must identify as such)~\cite{squad22018}, and SQuAD$_{open}$~\cite{squadopen2017}, which expands SQuAD to leverage the entirety of Wikipedia~\cite{zhu2021}. (\url{https://rajpurkar.github.io/SQuAD-explorer/}).

\textbf{QuAC}: The Question Answering in Context dataset (QuAC) also leveraged Wikipedia articles for span-prediction-based Q\&A systems~\cite{quac2018}. Unlike SQuAD, however, QuAC examines multi-turn Q\&A using a student teacher scenario, in which the student keeps asking for further clarifications on a specific topic. This dataset contains more than 100,000 conversation turns across 14,000 conversations, where each turn consists of a question and an answer, with each question reliant on knowledge of past dialogue to answer correctly. The dataset also leans more heavily on open-ended style questions (why, how). Evaluation is done through macro-averaged F1-score for word overlap of the system's answer and the correct span is used for evaluation, alongside HEQ (as described above). (\url{https://quac.ai/}).

\textbf{CoQA}: Similar to QuAC, the Conversational Question Answering (CoQA) dataset examines conversational question answering using span prediction of Wikipedia articles ~\cite{coqa2019}. CoQA contains more than 127,000 conversation turns across 8,000 conversations, where each turn consists of a factoid question and an answer, and each question is reliant on knowledge of past dialogue to be answered correctly. Unlike QuAC, CoQA includes unanswerable questions. As seen in QuAC, macro-average F1 score is again used for evaluation on CoQA. (\url{https://stanfordnlp.github.io/coqa/}).

\textbf{SearchQA}: SearchQA utilises a span-prediction approach to Q\&A using question and answer pairs selected from the television programme \textit{Jeopardy!}~\cite{searchqa2017}. Alongside each question answer pair is a series of relevant Google snippets within which the answer span can be identified. The system is thus challenged with identifying the correct answer span within the Google snippets. SearchQA contains more than 140,000 questions, with each question coupled with an average of 50 snippets. As with CoQA and QuAC, F1-score is most typically used to evaluate Q\&A systems on SearchQA. (\url{https://github.com/nyu-dl/dl4ir-searchQA}). 

These example datasets present just a few of the vast number of Q\&A training and benchmark datasets that have been released in recent years~\cite{zhu2021}. This is due to the problem that Q\&A datasets are often quickly ``solved''~\cite{rogers2020}. This solving, however, is typically not a sign of the quality of the Q\&A system so much as its ability to leverage annotation artefacts and lexical cues to artificially achieve high performances. Moreover, the current approach to training and evaluating Q\&A means that systems are unable to generalise away from the dataset in which they are trained. Even in cases where the questions being asked are of the same domain as the training set, most systems are unable to generalise to different datasets. This severely limits the current applicability of most Q\&A approaches.

These weaknesses also mean that many Q\&A systems are vulnerable to adversarial attack. In \cite{wallace2019}, for example, the authors identify the presence of universal adversarial triggers that can be used to prompt specific offensive outputs from a Q\&A-based system. Examining Q\&A systems trained on the SQuAD dataset, the authors proposed a method of identifying specific triggers that could be used to prompt the answer ``to kill American people'' for 72\% of all ``why'' questions posed. Other studies have noticed similar weaknesses, in which adversarial question framing can provide incorrect responses and a loss in performance from the Q\&A model~\cite{jia2017,rosenthal2021}. These vulnerabilities emphasise the current inability of Q\&A systems to genuinely perform language comprehension, finding that they all too often rely on superficial cues to generate answers~\cite{jia2017}. This, in turn, raises questions in regard to their current suitability to real-world applications, such as in education systems, FAQs and other customer-service roles~\cite{motger2021,rogers2020}.

\section{Rewriting} 
\label{sec:Rewriting}

We define rewriting tasks as tasks in which a given text is rewritten such that its original meaning is preserved, but some additional attributes are changed. This section covers two main subtasks involving text rewriting: \textbf{text style transfer} and \textbf{summarisation}. 

\subsection{Text Style Transfer}
\label{subsec:style-transfer}
  
Text Style Transfer (TST) is an NLG task which aims to rewrite a text according to a specific style ``property'' or ``attribute''. Therefore, the TST task ``aims to change the stylistic properties of any given text while preserving its style-independent content''~\cite{HLAZ2020}. This is a data-driven approach~\cite{JJHVM2020} focused on changing the syntax aspect of a text with respect to a given style attribute, whilst keeping the semantics intact~\cite{HLAZ2020}.  

Traditionally, TST has been achieved using \emph{parallel data} where there are matching text pairs for different styles e.g., informal and formal, positive and negative, modern English and Shakespearean English. In this case, sequence-to-sequence models and variations of those are often applied~\citep{JJHVM2020}. Despite the existence of a number of such datasets for specific use cases, for many TST cases there is a lack of parallel data available. Therefore, non-parallel datasets and the methods based on explicit or implicit \emph{disentanglement} of style and content~\cite{HLAZ2020} (i.e., methods that ``disentangle text into its content and [style] attribute in the latent space''~\citep{JJHVM2020}) have started to emerge and remain an active research area. 

\begin{figure*}[!htb]
\centering
\includegraphics[width=0.82\linewidth]{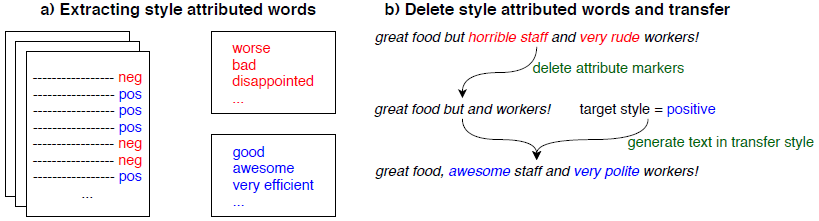}
\caption{An example of explicit disentanglement for text style transfer (Delete-Retrieve-Generate framework). The source style attribute is negative sentiment and the target style attribute is positive sentiment. Retrieved from \cite{HLAZ2020}}
\label{fig:Delete-Retrieve-Generate}
\end{figure*}

Explicit disentanglement follows three types of  action~\citep{JJHVM2020}: (1) encode the given text $t$ with the source style attribute $a$ in a latent representation; (2) manipulate the latent representation to remove $a$; and (3) decode into text $t\prime$ with target style attribute $a\prime$. An example technique in this category is illustrated in Fig.~\ref{fig:Delete-Retrieve-Generate} where a) corresponds to action (1) and b) includes actions (2) and (3).

Implicit disentanglement involves~\cite{HLAZ2020}: (1) learning the latent representations of content $c$ and of style attribute $a$ for the given text $t$; and (2) combining content $c$ with the latent representation of the target style attribute $a\prime$ to generate text $t\prime$. An example technique in this category is adversarial learning where two models are typically used: an adversarial network model for (1) and a style-embedding model for (2). 

Another stream of solutions for TST is to not rely on  disentanglement but rather build a pseudo-parallel dataset and then apply more traditional methods such as sequence-to-sequence models. There are two main approaches to achieve this~\citep{JJHVM2020}: \textbf{retrieval-based}, using existing datasets to identify pairs of sentences semantically similar using a metric; and \textbf{generation-based}, using an iterative process to generate the dataset.

Moving on to automated evaluation, the quality of TST solutions are typically determined by the following three main criteria~\citep{HLAZ2020}:

\textbf{Transferred style strength}: The \emph{Style Transfer Accuracy} metric measures ``whether each sample generated by the model conforms to the
target [style] attribute''~\citep{JJHVM2020} and is calculated as

\begin{center}
$\frac{\#\ test\ samples\ correctly\ classified}{\#\ all\ test\ samples}$.
\end{center}

An alternative metric is the \emph{Earth Mover's Distance} (EMD), which measures the minimum cost to turn the style distribution of the given text $t$ into the generated text $t\prime$. It can also be regarded as a measure of intensity of the style transfer~\citep{MFOR2019}.

\textbf{Semantic preservation}: The goal here is to measure the similarity of content between the given text $t$ and the generated text $t\prime$. \emph{BLEU} is the most widely used metric for TST solutions using parallel datasets, although others such as ROUGE and METEOR are also used~\citep{JJHVM2020}. 
On the other hand, \emph{sBLEU} and \emph{Cosine Similarity} are the mostly used for non-parallel settings~\citep{HLAZ2020}.

\textbf{Text fluency (or naturalness)}: The Perplexity score, calculated using a PLM for all style attributes on the training data, is the most commonly used metric to evaluate fluency of TST output, although its correlation with human evaluation remains a subject of debate~\citep{JJHVM2020}. The lower the perplexity score of a generated sentence (from text $t\prime$), the more aligned it is with the training dataset (from text $t$)~\citep{HLAZ2020}.

Human evaluation, although hampered by issues such as  subjectivity~\citep{HLAZ2020} and irreproducibility~\citep{JJHVM2020}, is also often used to provide insights about transferred style strength, semantic preservation, and text fluency given pairs of sentences from text $t$ and from the generated text $t\prime$. According to \citet{HLAZ2020}, best practices indicate the need to ``use 100 outputs for
each style transfer direction (e.g., 100 outputs for formal $\rightarrow$ informal, and 100 outputs for
informal $\rightarrow$ formal), and two human annotators for each task''.

In turn, TST has applications in range of key domains. Three of the most commonly considered are:   

\textbf{Writing assistance}: TST functionalities can be incorporated into tools to help users tailor or improve a written text according to a specified attribute. For example, it may be applied to the text of business emails or reports to make them look more professional, therefore improving formality (involving attributes informal $\rightarrow$ formal) or politeness (involving attributes impolite $\rightarrow$ polite). 

\textbf{Persuasive communication}: TST is a powerful mechanism for:
\begin{itemize}
\item Better engaging with an intended audience such as consumers (e.g., the style of a generic marketing text can be personalised according to a user profile), readers (e.g., image captions or headlines, which can be adapted to become more attractive according to attributes like humour, romance and clickbait~\citep{JJZOS2020}) and laymen or experts (e.g., a layman style can be used to make an expert text more readable while an expert style can be used to make a layman text appear more accurate and professional~\citep{HLAZ2020}).

\item Reaching a target community effectively using gender (e.g., male $\longleftrightarrow$ female), political ideology (e.g., Democrats $\longleftrightarrow$ Republicans), or through shared views and interests (e.g., emotions, topics).

\item Improving accessibility such as by using text simplification (complex $\rightarrow$ simple).

\item Adapting to user preferences and circumstances such as changing the sentiment conveyed in a text (negative $\longleftrightarrow$ positive). Another example of this are chatbots which can adapt their script style according to the type of interaction (e.g., a casual style for suggesting products to customers and a formal style for handling customers complaints~\citep{HLAZ2020}). 
\end{itemize}

\textbf{Re-styling for Social Good}: TST can also be used to improve text, such as social media posts and tweets, in terms of biasness (biased $\rightarrow$ neutral) and toxicity (offensive $\rightarrow$ non-offensive).

Building on the above, TST can be leveraged towards a wide range of subtasks. In turn, we focus on some of the most commonly studied subtasks in TST research, discussing their typical formulations and the common datasets that are generally leveraged in existing research.

\subsubsection{Sentiment Subtask}: This is a very popular TST subtask; it  involves the styles ``positive'' and ``negative''. There are 3 main datasets related to it mentioned below. These datasets were  all pre-processed to exclude neutral reviews.

\textbf{Yelp}: This is a non-parallel dataset  containing positive and negative real-world restaurant reviews~\citep{SLBJ2017}. All reviews have up to 10 sentences, with 250K negative sentences and 350K positive sentences in total. (\url{https://www.yelp.com/dataset}).

\textbf{Amazon}: This is a non-parallel dataset  containing positive (approximately 278K) and negative (approximately 279K) real-world Amazon users' reviews of products~\citep{HM2016}. Please note that these numbers come from \citet{HLAZ2020}. (\url{https://s3.amazonaws.com/amazon-reviews-pds/readme.html}).

\textbf{IMDb}: This is a non-parallel dataset containing real-world movie reviews collected from the Internet Movie Database~\citep{MDPHNP2011}. It contains 50K reviews with no more than 30 reviews for the same movie, and an equal split between positive and negative reviews. (\url{https://ai.stanford.edu/~amaas/data/sentiment/}).\\

\subsubsection{Formality Subtask}
\begin{figure}[!htb]
\centering
\includegraphics[width=1.0\linewidth]{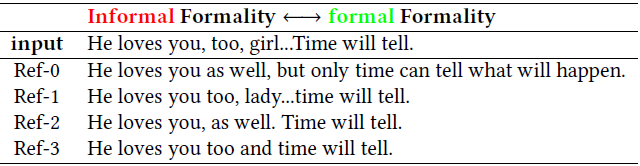}
\caption{An illustration of formality text style transfer with an informal sentence and 4 corresponding formal versions of it (Ref-0 to Ref-3). Retrieved from \citep{HLAZ2020}.}
\label{fig:formality}
\end{figure}

: This TST subtask involves the contrasting styles ``formal'' and ``informal''.   Figure~\ref{fig:formality} shows an example of an informal sentence (input) and four corresponding formal sentences (Ref-0 to Ref-3) with variations of length and punctuation. \citet{HLAZ2020} pointed out that formality is more complex than sentiment style transfer because it is more subjective -- different individuals may have very  different perceptions of what is formal. The Grammarly Yahoo Answers Formality Corpus (GYAFC) is the most popular dataset used for this subtask.

\textbf{GYAFC}: The GYAFC dataset contains parallel data, i.e., 110K pairs of formal-informal sentences~\citep{RT2018}. It was built with sentences collected from a Yahoo Answers corpus for Entertainment \& Music and Family \& Relationship, and processed to remove sentences that were too long or too short. The sentences deemed informal were then translated into formal ones via crowd-sourcing. (\url{https://github.com/raosudha89/GYAFC-corpus}).\\

\subsubsection{Politeness Subtask}: This TST subtask relates to the text styles ``polite'' and ``impolite''. It is interesting to notice that written expressions of politeness are culture-dependent even for the same language, and are also affected by social structures~\citep{MSPPNYSBP2020}.

\textbf{Politeness}: This is a non-parallel dataset containing 1.39 million sentences that were automatically labelled~\citep{MSPPNYSBP2020}. The sentences were collected from the Enron corpus, and therefore reflect politeness in the context of email exchanges in an American corporation. (\url{https://github.com/tag-and-generate/politeness-dataset}).\\

\subsubsection{Authorship Subtask}: This TST subtask aims to target a particular writing style from a linguistic point-of-view, therefore, it is more artistic compared to all other subtasks~\citep{JJHVM2020}. It is a form of  ``paraphrasing''~\citep{XRDGC2012}. There are two datasets that are commonly used for research in this space.

\textbf{Shakespeare}: This parallel dataset contains sentences in Shakespearean English style and corresponding sentences in modern English style~\citep{XRDGC2012}, totalling around 21K pairs. (\url{https://github.com/cocoxu/Shakespeare}).  

\textbf{Bible}: This parallel dataset contains over 1.5 million pairs of sentences aligned by verse numbers from ``the eight publicly available versions'' of the Bible~\citep{CRR2018}. (\url{https://github.com/keithecarlson/StyleTransferBibleData}).\\

\subsubsection{Simplicity Subtask}: This TST subtask relates to text styles ``complex'' and ``simple''. It aims to simplify text to make it more accessible for laymen, e.g., removing lexical or syntactic complexity. The target audience of text simplification also involves people with low literacy levels, such as children and non-native speakers, and people suffering from different kinds of reading comprehension, e.g., autism, aphasia, dyslexia~\cite{AA2021}. Some techniques useful to achieve simplification are~\citep{ZBG2010}: splitting (e.g., transforming long sentences into shorter ones), dropping (e.g., making  sentences more concise and sharper), reordering (e.g., rearranging sentences to make them easier to understanding), and substitution (e.g., replacing jargon and difficult terms with simpler synonyms).

For evaluating text simplification, common automated metrics such as BLEU are used. However, there also exist various task-specific metrics proposed for simplification evaluation. For instance, FKBLEU combines a paraphrase generation metric, iBLEU, with a readability metric, Flesch-Kincaid Index, to measure how adequate and readable a simplified text is~\cite{xu-etal-2016-optimizing}. Another metric is SARI, which measures the goodness of words that are added, deleted and kept by the simplification system~\cite{xu-etal-2016-optimizing}. Furthermore, readability indices are also commonly used to estimate how difficult a simplified text is to read~\cite{AA2021, SM2020}.

Several datasets are available for TST simplification where they aim to foster research and development for more effective communication between healthcare professionals and healthcare consumers, i.e., for medical text simplification.

\textbf{PWKP}: The Parallel
Wikipedia (PWKP) dataset contains over 108K pairs of complex sentences from \url{http://wikipedia.org/}{English Wikipedia} and  simple sentences from the \url{http://simple.wikipedia.org/}{Simple English Wikipedia}, which targets children and adults learning English~\citep{ZBG2010}. (\url{https://huggingface.co/datasets/turk}). 

\textbf{van den Bercken et al.'s EXPERT datasets}: These are three separate datasets proposed by \citet{BSL2019}: EXPERT-FULLY, EXPERT-PARTIAL and AUTOMATED-FULLY. The datasets contain pairs of complex medical sentences and corresponding simple sentences, also drawing from Wikipedia and Simple Wikipedia. The EXPERT-FULLY dataset has 2,267 fully aligned medical sentences, the EXPERT-PARTIAL dataset has 3,148 partially aligned sentences, and the AUTOMATED-FULLY dataset has 3,797 fully aligned medical sentences. (\url{https://github.com/myTomorrows-research/public/tree/main/WWW2019}). 

\textbf{MIMIC-III}: This is a non-parallel dataset containing  real-world clinical sentences written in professional (medical) style and in consumer (layman patient) style. It has 443K sentences in professional language, and 73K sentences in consumer language~\citep{WCS2019}. (\url{https://physionet.org/content/mimiciii/1.4/}). 

\textbf{BenchLS}: BenchLS is a combination of two non-parallel lexical simplification datasets, LexMTurk and LSeval, containing 929 instances in total. Each instance consists of a sentence, a target complex word, and several (7.37 on average) candidate substitutions ranked according to their simplicity~\cite{paetzold2016benchmarking}. (\url{https://zenodo.org/record/2552393#.YYkZLdnP0_8}).

\textbf{NNSeval}: NNSeval is a non-parallel dataset that covers complex words for non-native speakers. All sentences in the dataset were taken from Wikipedia, LexMTurk and LSeval, and 400 non-native speakers identified the complex target words. The resulting dataset contains 239 instances~\cite{Paetzold_Specia_2016}. (\url{https://zenodo.org/record/2552381#.YYkZRNnP0_8}). 

\textbf{SS Corpus}: This is a parallel corpus containing 492,993 aligned sentences extracted by pairing \href{http://simple.wikipedia.org/}{Simple English Wikipedia} with \href{http://wikipedia.org/}{English Wikipedia}~\cite{kajiwara2016building}. (\url{https://github.com/tmu-nlp/sscorpus}). 

\begin{figure}[!htb]
\centering
\includegraphics[width=1.0\linewidth]{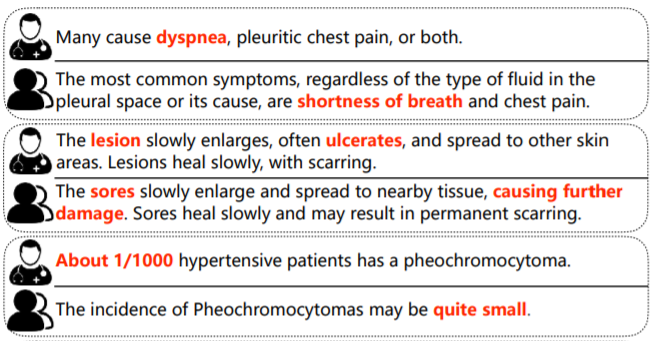}
\caption{An illustration of simplification text style transfer with 3 pairs of sentences in expert style (upper sentences) and in layman style (lower sentences). Adapted from figure by \citet{CSPKLC2020}.}
\label{fig:simplification}
\end{figure}

\textbf{MSD}: 
MSD is a parallel dataset motivated by the example shown in Fig.~\ref{fig:simplification} where expert sentences (upper sentences) are simplified into layman sentences (lower sentences)~\citep{CSPKLC2020}. It contains 130K expert-style  sentences and 114K layman-style sentences. (\url{https://srhthu.github.io/expertise-style-transfer/#disclaimer}).

\textbf{Newsela}: This is a parallel corpus containing 1,130 news articles with four simplified versions each. The simplified versions were written by professional editors at \href{https://newsela.com/}{Newsela}, a company that produces reading materials for children~\cite{xu2015}. (\url{https://newsela.com/data/}).\\

\textit{Gender Subtask}: This TST subtask draws from socio-linguistics research showing that gender is associated with language choices~\citep{NDRJ2016}. Assuming gender as a  biological binary attribute of an individual, it typically involves text styles ``female'' and ``male''. 

\textbf{Yelp Gender}: This non-parallel dataset has been made available by \citet{PTSB2018}. It builds on a previous private dataset by \citet{RK2016} compiled from the Yelp Dataset Challenge 2016 and annotated with ``male'' and ``female'' labels for reviews in gender-neutral domains. The reviews were divided into more than 2.5 million sentences, where ``only sentences that are strongly
indicative of a gender'' were kept~\citep{PTSB2018}. (\url{http://tts.speech.cs.cmu.edu/style_models/gender_data.tar})

\textbf{RtGender}: This is a non-parallel dataset containing over 2.5 million sentences compiled from responses to online posts or videos where the gender of the (source) author and the gender of the responder were clear. The source-responder sentences have been extracted from comments on  Facebook (US politicians and public figures), TED talks, Fitocracy (fitness), and Reddit~\citep{VJPJT2018}. (\url{https://nlp.stanford.edu/robvoigt/rtgender/}).

\subsubsection{Toxicity Subtask}: This TST subtask relates to text styles ``offensive'' and ``non-offensive''. Changing style from the former to the latter contrasts with the approach of simply filtering and removing such content online, especially in relation to social media posts. 

\textbf{Nogueira dos Santos et al.'s Twitter and Reddit datasets}: These are non-parallel datasets by \citet{SMP2018}. The authors used ``sentences/tweets with size between 2 and 15 words and removed repeated entries''. Their Twitter dataset contains just under 2 million entries, while their Reddit dataset contains over 7 million entries. 

\subsubsection{Anonymisation Subtask}

According to the GDPR legislation, ``anonymisation is the complete and irreversible process of removing personal identifiers, both direct and indirect, that may lead to an individual being identified''~\citep{LPSBO2021}. It has been approached from a privacy preserving perspective and from a NLP perspective. The former focuses on risks of disclosure of personal identifiers by an adversary, while the latter focuses on linguistic patterns helpful to infer personal identifiers, such as gender, age, race, geographical location and affiliations (called quasi-identifiers~\citep{LPSBO2021}), that can be used, e.g., for user profiling~\citep{NDRJ2016} or discrimination~\citep{,RK2016}. 

In the domain of NLP, anonymisation is a TST subtask (non-anonymised $\rightarrow$ anonymised), that has been predominately approached in two ways: \emph{de-identification} and \emph{obfuscation}. De-identification aims to detect and remove personal identifiers, and is often applied to the medical domain, i.e, to Protected Health Information (PHI). Whereas obfuscation aims to detect and rewrite text to ``reduce the leakage of sensitive information''~\citep{XQXC2019} while retaining text semantics and fluency. To illustrate obfuscation, let's consider the following examples from~\citep{XQXC2019}:

\begin{center}
\noindent\fcolorbox{blue}{white}{\begin{minipage}{8cm}
Original: \textit{I am a software engineer with 18 years of working experience.}\\
Rewritten: \textit{I am a software engineer with more than 10 years of working experience.}\\
Original: \textit{I went with my girlfriend and another  couple.}\\
Rewritten: \textit{I went with my friend and another couple.}
\end{minipage}}
\end{center}

The obfuscation subtask can use the following  tailored evaluation metrics~\citep{XQXC2019}: average entropy (applied to all predictions of the classifier; higher entropy means less sensitive data leakage); predicted accuracy and modified accuracy (rate of accepted sentence modification). 

Datasets used for anonymisation have all gone through a ``surrogate process'' to replace real-world personal identifiers with fictitious but realistic ones. 

\textbf{2010 i2b2 NLP challenge corpus}: This dataset contains over 800 diabetic patient medical records~\citep{USSD2011}, manually ``annotated for an extended set of PHI categories''~\citep{LPSBO2021}. 

\textbf{VHA}: The Veterans Health Administration dataset contains 800 clinical notes also manually annotated with PHI categories such as ``Social Security Numbers, Patient Names, and Dates''~\citep{FSSFSM2013}.
(\href{https://oup.silverchair-cdn.com/oup/backfile/Content_public/Journal/jamia/20/1/10.1136_amiajnl-2012-001020/1/20-1-77_Supplementary_Data.zip}{Download.)}

\textbf{2016 CEGS N-GRID}: This dataset contains 1K  psychiatric intake records and ``more than 34,000 PHI phrases, with an average of 34 PHI phrases per record''~\citep{SFU2017}. It is annotated with PHI such as doctor name, patient name and hospital location. 

\textbf{ITAC}: The Informal Text Anonymisation Corpus contains 2.5K personal emails with 31,926 personal identifiers including ~\citep{M2006} with direct identifiers (e.g., name of individuals) and quasi identifiers (e.g., name of organisations)~\citep{LPSBO2021}.

Efforts similar to the ITAC resulted in a number of datasets in languages other than English. One such dataset is:

\textbf{CodE Alltag 2.0}: This dataset contains over 240K emails in German~\citep{EKH2020}. (\url{https://github.com/codealltag})

\subsection{Summarisation}

Text summarisation is a rewriting subtask which aims to generate a short and coherent version of a text that contains the main ideas, topics, and/or concepts of the original text~\cite{HZF2019}. Considering the vast amounts of digital textual data available, text summarisation can decrease the time needed for text processing in multiple contexts. 

In the literature, two major approaches for text summarisation exist: \textbf{extractive summarisation} and \textbf{abstractive summarisation}. The former tries to identify the most relevant utterances or sentences from input text which describe the main theme. The latter aims to generate a fluent and concise summary, paraphrasing the intent of the input text in a shortened form~\cite{HZF2019}.

Human evaluation is commonly used for evaluating text summarisation methods. Human evaluation of text summarisation mostly focuses on the ability of the generated text to capture the key contents of the input text. The focus of human evaluators is therefore generally directed towards measuring the informativeness, coverage, focus, and relevance of the summary text. Furthermore, more general measures of text fluency, readability, coherence and repetition are also often considered to evaluate linguistic quality. Human evaluation methods typically leverage common scoring methods, including Likert-type scales, rank-based annotations, and pairwise comparisons. Other proposed methods are Best-worst scaling (BWS), which is a specific type of ranking-oriented evaluation that requires annotators to specify only the first and last rank, and question-answering (QA)~\cite{SM2021}. 

Some of the popular datasets used for evaluating text summarisation are as follows:

\textbf{Document Understanding Conferences (DUC)}: DUC is a series of conferences run by the National Institute of Standards and Technology (NIST) focusing on the area of text summarisation. From 2001 to 2007, text summarisation datasets have been provided in the scope of the conferences. The datasets contain news articles from AQUAINT, TIPSTER and TREC corpora, and are available upon request. (\url{https://duc.nist.gov/data.html}).

\textbf{Text Analysis Conference (TAC)}: TAC is another conference series run by NIST between 2008-2011. Each conference had a summarisation track where a relevant dataset has been distributed. The datasets contain news articles, and are available upon request. (\url{https://tac.nist.gov/data}).

\textbf{New York Times Annotated Corpus (NYTAC)}: The NYTAC dataset contains over 1.8M news articles published by New York Times between 1987-2007, as well as 650K article summaries. The articles were manually summarised by library scientists. (\url{https://catalog.ldc.upenn.edu/LDC2008T19})

\textbf{Large Scale Chinese Short Text Summarization Dataset (LCSTS)}: The LCSTS dataset is a Chinese text summarisation dataset constructed from a Chinese micro-blogging website, Sina Weibo~\cite{hu2016lcsts}. It contains over 2M real Chinese short texts with short summaries provided by the author of each text. (\url{http://icrc.hitsz.edu.cn/Article/show/139.html}).

\textbf{CCF Conference on Natural Language Processing \& Chinese Computing (NLPCC)}: NLPCC is a series of conferences on NLP organised annually in China. The conferences had a track for text summarisation in 2015, 2017 and 2018, where a text summarisation dataset was distributed for each year. (\url{http://tcci.ccf.org.cn/}).

Text summarisation has various applications. For example,  medical conversation summarisation aims to summarise conversations between doctors, nurses, and patients about the proposed diagnoses and treatments so that patients can review them later without having to deal with a full record or transcript~\cite{L2019}. Text summarisation can also be used to increase the efficiency in processing long documents such as scientific papers~\cite{gupta2021} and long speeches~\cite{RBQKWLC2020}.

Beyond more conventional text summarisation, which aims to summarise text inputs, speech summarisation has also been the focus of a considerable degree of study.

Speech summarisation aims to identify the most important content within human speech and then generate a condensed form of text suitable for the needs of a given task~\cite{RBQKWLC2020}. Unlike standard text summarisation approaches which mostly generate text from another text, speech summarisation methods take audio data as the input, and utilise speech recognition to process it. Nonetheless, speech summarisation approaches are similar to the standard text summarisation approaches, using the major approaches of \textbf{extractive summarisation} and \textbf{abstractive summarisation} discussed earlier.

For evaluating speech summarisation, qualitative and quantitative metrics have been used. Human evaluation typically leverages similar quality criteria to that of standard text summarisation, including readability, coherence, usefulness and completeness. Beyond human evaluation, automated evaluation using common NLG metrics, such as ROUGE (and its variants, ROUGE-1, ROUGE-2, ROUGE-3, ROUGE-L, ROUGE-SU4 and ROUGE-W), are common. Other performance measures, including precision, recall, F-measure, word accuracy and Pyramid, are also often used to measure how well a generated summary's content matches the content of the reference summary.

For evaluation of speech summarisation, the following datasets have been used:

\textbf{AMI}: The AMI meeting corpus is a multi-modal data set consisting of 100 hours of meeting recordings in English. Furthermore, it contains manually produced orthographic transcriptions for each individual speaker, as well as a wide range of other annotations, including extractive and abstractive summaries. (\url{https://groups.inf.ed.ac.uk/ami/corpus/}).

\textbf{International Computer Science Institute (ICSI) Meeting Corpus}: The ICSI Meeting Corpus is an English audio dataset consisting of approximately 70 hours of meeting recordings. It also contains orthographic transcriptions, and manual annotations of dialog acts and speech quality. (\url{https://groups.inf.ed.ac.uk/ami/icsi/})

\textbf{MultimodAl (Task-oriented) gRoup dIsCuSsion (MATRICS)}: The MATRICS corpus contains discussions in English among four native Japanese speakers on three different topics. It involves 9 hours of group meeting recordings consisting of 29 dialogues by 10 conversation groups. (\url{https://github.com/IUI-Lab/MATRICS-Corpus}).

\textbf{Corpus of Spontaneous Japanese (CSJ)}: CSJ is a dataset containing Japanese spoken language data and information for use in linguistic research. The dataset consists of 958 hours of lectures and task-oriented dialogues in Japanese. (\url{https://ccd.ninjal.ac.jp/csj/en/}).

\textbf{Topic Detection and Tracking (TDT2)}: TDT2 contains news data collected daily from nine news sources in two languages (American English and Mandarin Chinese) over a period of six months. (\url{https://catalog.ldc.upenn.edu/LDC2001T57}).

\textbf{RT-03 MDE}: This dataset contains English Conversational Telephone Speech (CTS) and Broadcast News (BN) transcripts and annotations covering 40 hours of CTS and 20 hours of BN data. Annotations include fillers (e.g., ``um'', ``err''), discourse markers (e.g., ``you know''), and semantic units (e.g., statements, questions). (\url{https://catalog.ldc.upenn.edu/LDC2004T12}).

\textbf{Mandarin Speech Data Across
Taiwan Broadcast News (MATBN)}: The MATBN Mandarin Chinese broadcast news corpus contains a total of 198 hours of broadcast news from the Public Television Service Foundation (Taiwan) with corresponding transcripts and annotations. (\url{http://slam.iis.sinica.edu.tw/corpus/MATBN-corpus.htm}).

\textbf{Switchboard-1}: This dataset contains a collection of approximately 2,400 two-sided telephone conversations among 543 speakers (302 male, 241 female) from the US, containing 260 hours of speech. (\url{https://catalog.ldc.upenn.edu/LDC97S62}).

\textbf{Fisher}: Fisher is another telephone speech dataset, containing 2,000 hours of conversational speech data in English. It has been built for the DARPA Effective, Affordable Reusable Speech-to-text (EARS) program.

\textbf{Spoken Language Data base (BEA)}: BEA is a Hungarian spontaneous speech dataset that consists of 250 hours of speech data.

Speech summarisation has several bespoke applications, including improved efficiency and cost reduction in telephone contact centres (e.g., by identifying call topics, automatic user satisfaction evaluation, and efficiency monitoring of agents), more efficient progress tracking in project meetings, the facilitation of learning using online courses, digital scribes, and conversational agents. Speech summarisation can be used for deception, with the main idea of a speech being taken out of context in its textual summary to mislead readers.

\begin{figure*}[!ht]
\centering
\includegraphics[width=0.8\linewidth]{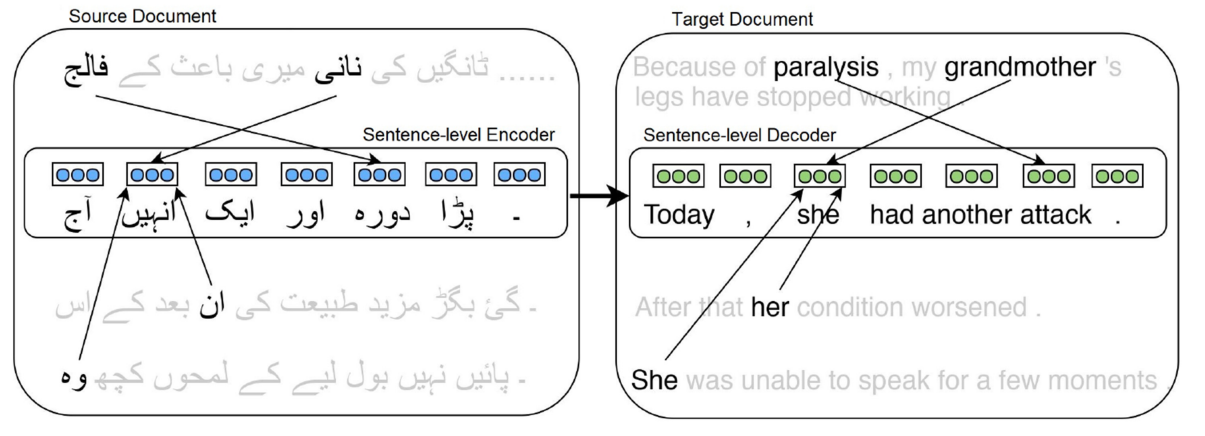}
\caption{Example of a document-level MT model. Retrieved from \cite{maruf2021}.}.
\label{fig:document_level_MT_example}
\end{figure*}

\section{Translation \& Interpretation}
\label{sec:Translation}

Translation has been broadly construed to cover all NLG tasks in which a given data input (e.g., text in another language) is translated so that it is represented in natural language text.

This, in turn, encompasses more than just translation from one language to another, also including other mediums, such as the translation of an image to a text caption that describes the image (which can be thought of as translating the image from a visual medium to text). In some cases, interpretation is also required, such as when a machine attempts to process puns when translating natural language, summarise an image when generating image captions, or assess the context logic of source code when generating comments.

In this section, we split the translation and interpretation task into five subtasks: \textbf{language translation}, \textbf{image captioning}, \textbf{speech recognition}, \textbf{explainability}, and \textbf{code comment generation}. Language translation refers to the use of a machine to translate a text written in a source language into a text written in a target language. Image captioning means translating image information into text describing the image's content. Speech recognition translates audio information into meaningful text. As a translation subtask, explainability aims at using NLG to generate explanations of the behaviours of black box models. Code comment generation focuses on automatically generating code comments based on a given source code input, which can be conceptualised as translating source code to text. 

For each subtask, we define some of its relevant potential end-user applications. We also discuss typical approaches, typical evaluation methods, and any relevant datasets.

\subsection{Language Translation}

Language translation, typically referred to as machine translation (MT), is the process of automating the translation of text from one natural language to another~\cite{maruf2021}. In this subsection, we cover some of the key aspects of the MT subtask, including document-level machine translation, humorous wordplay translation, and translation quality assessment.

Document-level machine translation (otherwise known as discourse-level machine translation) refers to a translation process that utilises inter-sentential context information, which includes the discourse of text and the surrounding sentences in the input document~\cite{maruf2021,Z2020}. Figure~\ref{fig:document_level_MT_example} provides an example of the basic architecture of a document-level MT model, which also indicates how inter-sentential context information is useful. 

The document-level machine translation context approaches are classified based on two dimensions:

\begin{itemize}
    \item Whether the approaches use local or global models; where local models use only the neighbourhood of the current sentence while the global models consider both past and future translation decisions.
    
    \item Where the context comes from i.e., from the source-side, or from both the source and target-side contexts. For example, in Fig.~\ref{fig:document_level_MT_example}, the left side is the source document, the right side is the target document. Approaches can utilise both sides or source-side only. There is no model that works on the target side only as it is natural to leverage the existing source-side context when performing translation~\cite{maruf2021}.
\end{itemize}

Approaches are then classified into four groups: \textbf{(1)} local source-only context, \textbf{(2)} local source and target context, \textbf{(3)} global source-only context, \textbf{(4)} global source and target context. Furthermore, the learning approaches are classified into two groups: \textbf{(1)} modifying the training strategy and \textbf{(2)} utilising contextualised word embeddings. The former involves the introduction of regularisation and reward functions in the training objective, or even modifying the learning process. The latter provides a warm-start for the training process, 
using the MT model to predict both the target-side sentence and the source-side context. The idea is that the source-side sentence embeddings can be integrated into the MT model in order to utilise the source-side document-level dependencies~\cite{maruf2021}.

There has been much research focused on encodings for document-level MT. Cache memory methods have been introduced, which can carry over word preferences from one sentence to the next~\cite{Z2020}. NN-based discourse-level approaches, which predict the next possible sentence by retrieving the historical conversation, have also been suggested~\cite{maruf2021}.

Humorous wordplay translation is a popular research problem in language translation -- referring to the challenge of translating puns from one language to another~\cite{naiyu2021,miller2019}. However, most studies have aimed to resolve the translation problems as the single ``correct”  interpretation~\cite{miller2019}, rather than addressing the problem of translating the pun itself, which typically contains multiple intended meanings. \citet{delabastita1996wordplay} covers eight common strategies for translating wordplay:

\begin{enumerate}
    \item Replace the source-language pun with a similar target-language pun.
    
    \item Substitute the pun with non-punning language that preserves one or both of the original meanings.
    
    \item Replace the pun with some non-punning wordplay or rhetorical device (e.g., irony, alliteration, vagueness).
    
    \item Omit the language containing the pun.
    
    \item Leave the pun in the source (original) language.
    
    \item As a compensatory measure, introduce a new pun at a discourse position where the original had none.
    
    \item As a compensatory measure, introduce entirely new material containing a pun. 
    
    \item Editorialise: insert a footnote, endnote, etc. to explain the pun.
\end{enumerate}

In general, existing MT approaches are unable to deal with humour translation problems. There are, in turn, three major research directions that have been identified for further study: \textbf{(1)} to study how human translators process puns, \textbf{(2)} to generate and rank lists of pun translation candidates, \textbf{(3)} to develop interactive NLP-based methodologies for supporting human translators, which help assess whether a given pun can be replaced with a target-language pun~\cite{miller2019}.

For evaluation of MT, human assessment methods are most commonly used~\cite{han2021translation}. As is typical amongst NLG more broadly, the quality criteria of \emph{intelligibility}, \emph{fluency}, \emph{adequacy}, and \emph{comprehensibility} have been primarily considered. For document-level MT, evaluation of pronoun translation, lexical cohesion and discourse connectives have also been utilised. Beyond this, more advanced evaluation approaches also exist. These include using extended quality criteria including: \emph{suitability}, whether the results are suitable in the desired context; \emph{interoperability}, whether the MT system works with other platforms; \emph{reliability}, whether the MT system will fail (and its fault rate); and \emph{usability}, whether the MT system is easy to learn and operate.

Automated assessment methods have also been leveraged in evaluating MT~\cite{han2021translation}. This includes common metrics such as edit distance, word error rate, translation edit rate, position-independent word error rate, BLEU, METEOR, ROUGE, precision, and recall. Neural networks, especially Deep Learning, have also been suggested for translation quality assessment, such as by using a NN to find the best translation from a series of candidate translations, via comparison with a reference translation~\cite{GJMN2015,GJMN2017}.

Machine translation has many key applications, and has been widely applied in real-world systems. Common applications of MT thus include business, education and government (e.g., real-time translation during meetings, online translation software, websites with multiple languages)~\cite{maruf2021}. Moreover, electronic dictionaries, translation memories (a pre-defined database for aiding human translators), computer-assisted translation (CAT) tools and component-based CAT workbenches (for professional human translators) have all benefited from the development of MT~\cite{miller2019}. 

However, despite these benefits, machine translation runs the risk of being targeted by attacks from malicious users. \href{https://www.theguardian.com/technology/2017/oct/24/facebook-palestine-israel-translates-good-morning-attack-them-arrest}{In 2017, the Israeli police arrested a Palestinian man after he posted ``good morning" in Arabic}. This is due to the fact that the MT system incorrectly translated this Arabic as ``attack them" in Hebrew. Similarly, \href{https://www.theguardian.com/technology/2017/oct/13/chinese-messaging-app-error-sees-n-word-used-in-translation}{an MT system incorrectly translated a neutral Chinese phrase into a racially discriminatory phrase on Chinese social media}. Attacks against MT systems may occur in the model training phase or through corpus poisoning, and could be used maliciously to cause similar incidents of mistranslation.

In language translation,  a wide range of datasets have been leveraged in the training and evaluation of proposed MT systems. Some of the most commonly used datasets include:

\textbf{WordNet}: A widely used synonym database in the NLP literature, which groups English words into sets of synonyms. (\url{https://wordnet.princeton.edu/}).

\textbf{OntoNotes}: A corpus consisting of 2.9 million words in English, Arabic and Chinese. OntoNotes contains text from news, conversational telephone speech, weblogs, usenet newsgroups, broadcast, and talk shows alongside structural information (syntax and predicate-argument structure) and shallow semantics. (\url{https://catalog.ldc.upenn.edu/LDC2013T19}).

\textbf{FrameNet}: A database of approximately 1,200 scripts (semantic frames) covering over 13,000 English word senses. (\url{https://framenet.icsi.berkeley.edu/fndrupal/}).

\textbf{ACL 2019 Fourth Conference on Machine Translation (WMT19) corpus}: This corpus contains news text in ten language pairs. The language pairs are Chinese-English, Czech-English, Finnish-English, German-English, Gujarati-English, Kazakh-English, Lithuanian-English, Russian-English, German-Czech, and French-German. \textbf{Rapid corpus}, \textbf{Newscrawl Corpus}, and \textbf{Europarl v7/v9} are datasets that combine the WMT19 corpus with additional languages and sources. (\url{http://www.statmt.org/wmt19/translation-task.html}).

\textbf{International Conference on Spoken Language Translation (IWSLT)}: IWSLT is an annual scientific conference with an open evaluation campaign. In the IWSLT, the evaluation campaign is for translating TED talks from Chinese to English (tst2010-2013), from French to English (tst2010), and from Spanish to English (tst2010-2012). Each of these language tasks contains around 200,000 sentence pairs. (\url{https://iwslt.org/}).

\textbf{RotoWire}: This dataset contains 4,853 NBA basketball game summaries in English between 2014 and 2017. The 3rd Workshop on Neural Generation and Translation (WNGT 2019) manually translated a portion of the RotoWire dataset to German. The dataset provider recommends using the following SportSett:Basketball dataset, which corrects some dataset contamination issues from the standard Rotowire dataset. (\url{https://github.com/nlgcat/sport_sett_basketball})
    
\subsection{Image Captioning}

Image captioning, otherwise known as image description generation, is aimed at automatically generating a description of an image, using the image as input~\cite{hossain2019}. This is typically construed as a translation problem -- translating the image information to text. This subsection discusses the general approaches for image captioning, as well as the typical evaluation approaches, popular datasets, and related applications relevant to this subtask.

There are three main approaches to image captioning~\cite{hossain2019}: 

\textbf{Template-Based Methods}: These methods use a fixed text template for caption creation. The template contains blank slots (typically called variables), and object, attribute, and action words relevant to the image are selected by the captioning system to fill these slots. Whilst useful in some circumstances, template-based methods are limited by the static nature of the template itself, which limits the flexibility of any captions generated.

\textbf{Retrieval-Based Methods}: This approach focuses on retrieving the correct caption from a set of captions. Whilst this approach ensures the fluency of the output caption, retrieval-based methods are limited to producing more general captions instead of image-specific ones.

\textbf{Automatic Generation-Based Methods}: These methods typically leverage some form of deep-learning model (e.g., encoder-decoder, bi-LSTM) to generate an original description of the input image. This approach offers greater capability in creating original image-specific descriptions but is often hampered by weak fluency in the generated caption.

Deep learning has been widely used for image captioning~\cite{hossain2019}. Here we discuss some of the common deep-learning-based approaches that have been used in the literature~\cite{hossain2019}.

\textbf{Feature Mapping}: Visual space approaches are most common here;  these independently pass the image features and the captions to the decoder. Multi-modal space embeddings have also been proposed for feature representation, which combine image and text embeddings. AlexNet and VGGNet are commonly used as image encoders.

\textbf{Learning Types}: Common supervised-learning-based approaches include encoder-decoders, attention-based models, and dense image captioning. These approaches use labelled data in the training phase. Other deep-learning-based approaches include reinforcement and unsupervised techniques, which extract the image features using image encoders and then pass these features to the language decoders. GAN-based methods have also been successfully used in image captioning.

\textbf{Captioning Types}: This refers to the regions of the scene that are being captioned, and thus the form of image captioning that is being conducted~\cite{hossain2019}. These are generally classified as:

\begin{itemize}
    \item Whole-Scene captioning, in which the captions aim to summarise the entire image. Common approaches to this include:
    \begin{itemize}
        \item Attention-based methods: Image features are obtained using a CNN, and from this an attention-based LM generates some words or phrases. Parts of captions are then constructed from generated terms, and the captions are dynamically updated to account for the various regions of the scene within the caption. This method allows for the inclusion of images during learning steps to emphasise key regions during captioning.
        \item Novel-object-based methods: Existing methods rely on paired image caption datasets. This approach uses a separate lexical classifier and LM trained on separate image and text data. Then, using paired image caption data, a deep-caption model is trained. Finally, both models are trained together. This approach allows for the generation of captions that describe novel objects not present in that paired image caption training data.
    \end{itemize}
    
    \item Dense captioning, in which the system focuses on captioning specific regions of the image. To do this, typical approaches first divide the target image into different regions. Then, features are extracted from each of the different regions. These features are then passed to an LM, which generates captions for each region.
\end{itemize}

\textbf{Model Architecture}: There are a wide range of deep-learning architectures that have been proposed for use in image captioning. These include:
\begin{itemize}
    \item Encoder-decoder architecture: This approach is generally used in conjunction with a CNN trained to classify scene type, objects and relations within the image. After this classification, the encoder-decoder model converts the outputs of the CNN into words and generates the caption. Each word in the caption is selected based on visual information and previous context to ensure coherence and accuracy.
    
    \item Compositional architecture: Image features are obtained by using a CNN. Then, the visual concepts (e.g., regions, objects, and attributes) obtained from these features are used as additional image features. Multiple captions are generated by an LM using the output of the above two processes. A deep multi-modal model is then used to rank and select the final captions.
\end{itemize}

The evaluation of image captioning is mainly automatic and quantitative based~\cite{hossain2019}. Several metrics are commonly used when evaluating the quality of image captioning, with BLEU and METEOR being particularly popular, though ROUGE, CIDEr and SPICE have also seen some use. WMD is another metric that has been suggested, though its failure to account for word order and its inability to measure readability limit its utility~\cite{pavlopoulos2019}.

A common application of image captioning systems is content-based image retrieval (CBIR), which relies on image indexing (i.e. the annotation of images in a database) as a crucial component. Image captioning, in turn, is highly suited to automatically generating captions for these data samples~\cite{hossain2019}.
Another major application of image captioning is in the biomedical field, where it can be used to help physicians identify lesions in PET/CT scans or radiology images~\cite{pavlopoulos2019}. Various social media applications are also possible, such as identifying places, events, and clothes in images.

\begin{figure*}[!htb]
\centering
\includegraphics[width=0.7\linewidth]{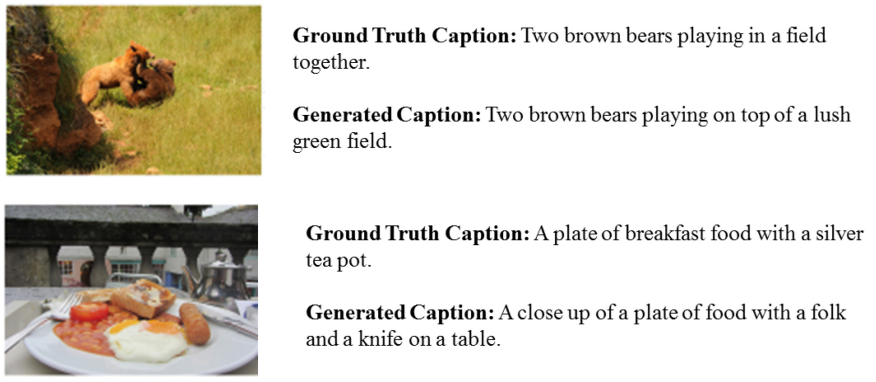}
\caption{Examples of image captioning on MS COCO datasets. Retrieved from \cite{wu2015image}.}.
\label{fig:sample_from_MS_COCO}
\end{figure*}

For image captioning, the following datasets are commonly used:

\textbf{MS COCO}: The Microsoft COCO dataset is used for object detection, segmentation, and captioning. It contains over 330K images with five captions per image. It also has 80 object categories and 1.5 million object instances. Figure~\ref{fig:sample_from_MS_COCO} gives two examples of image captioning on MS COCO datasets by ~\citet{wu2015image}. (\url{https://cocodataset.org/}).

\textbf{FLICKR 30K}: FLICKR 30k has been a standard benchmark for automatic image description. It contains 30K images collected from Flickr with 158K captions provided by human annotators. (\url{https://github.com/BryanPlummer/flickr30k_entities}).

\textbf{Visual Genome}: This dataset has multiple region captions. It contains over 108K images with an average of 35 objects, 26 attributes, and 21 pairwise relationships between objects per image.  (\url{https://visualgenome.org/}).

Beyond the above datasets, there also exist datasets specifically aimed at biomedical image captioning~\cite{pavlopoulos2019}. These include:

\textbf{IU-X-ray}: This dataset contains 7,470 x-rays images and is publicly accessible from the Open Access Biomedical Image Search Engine (OpenI). (\url{https://openi.nlm.nih.gov/}).

\textbf{ICLEF-CAPTION} (Image Concept Detection and Caption Prediction): This dataset has 232,305 biomedical images with captions. (\url{https://www.imageclef.org/2018}). 
    
\subsection{Speech Recognition}
Speech recognition, also referred to as automatic speech recognition (ASR), is a task that aims to automatically translate audio information into meaningful text. Speech translation (ST), a task that builds upon work in ASR and MT, focuses on translating the audio into text in a different language~\cite{sperber2020speech}.

\citet{sperber2020speech} surveyed the historical development of speech translation and divided it into 4 stages:

\textbf{Loosely Coupled Cascades}: In simple terms, researchers would separately build ASR and MT systems, then use the results of the former as the input to the latter. Since early MTs were unable to handle input with irregular formats, errors may be transmitted from the ASR stage.

\textbf{Toward Tight Integration}: Further studies then tried to address the early decisions problem -- where errors in the output of the ASR are passed on to the MT system. To solve this, the N-best translation approach was introduced, which summarises all possible transcriptions of the ASR outputs. Another suggested approach processed the intermediate results and optimised the input structure of the MT system based on its domain. Prosody (e.g. pitch and loudness) transfer was also suggested, which is used for applying the source-side prosody to target-side words during the transformation. 
    
\textbf{Speech Translation Corpora}: Before this, ASR and MT used separate corpora for training. This often led to a mismatch between ASR and MT trained on data from different domains. Researchers thus started to use the same corpus to train ASR and MT. However, the cost of manual annotation is high and suffers from language coverage limitations.
    
\textbf{End-to-End Models}: End-to-end ST corpora and models for MT and ASR are now commonly used. Other approaches like end-to-end trainable cascades and triangle models, multi-task training and pre-training (incorporating additional ASR and MT data), and speech-to-speech translation have also been proposed.

The authors also defined three types of end-to-end training data (i.e. pairs of speech inputs and output translations)~\cite{sperber2020speech}. \textbf{(1)} Manual: The speech corpora used for training is translated by humans. \textbf{(2)} Augmented: Data is obtained by extending an ASR corpus with automatic translation or an MT corpus with synthetic speech. \textbf{(3)} Zero-Shot: Using no end-to-end data.

Four of the commonly used corpora in ASR are:

\textbf{MuST-C}: This dataset contains more than 385 hours of audio recordings from the English TED Talks into eight languages (German, Spanish, French, Italian, Dutch, Portuguese, Romanian and Russian). (\url{https://ict.fbk.eu/must-c/})

\textbf{MaSS}: This dataset contains more than 172 hours of audio recordings from the Bible across eight languages (Basque, English, Finnish, French, Hungarian, Romanian, Russian and Spanish) (\url{https://github.com/getalp/mass-dataset})

\textbf{LibriVoxDeEn}: This dataset contains 110 hours of audio recordings from German audiobooks with German text and English translation. (\url{https://www.cl.uni-heidelberg.de/statnlpgroup/librivoxdeen/})

\textbf{Europarl-ST}: This dataset contains paired audio-text samples in nine languages (Romanian, Polish, Dutch, German, English, Spanish, French, Italian and Portuguese). (\url{https://www.mllp.upv.es/europarl-st/})
    
\subsection{Explainability}
A clear definition of ``explanation'' has not been agreed upon by the scientific community. Generally, explanations are classified by different types, such as explanation by example, counterfactual explanation (i.e. explaining how a model's behaviour would change if its input was $x$ instead of $y$), local explanation (explaining the effect of a single input), feature importance (i.e. which feature(s) have the greatest influence), or a combination thereof. In principle, eXplainable Artificial Intelligence (XAI) aims to add a ``linguistic explanation layer'' to decision tools to help end users and improve adoption in the market~\cite{MAG2020}.

Broadly, there are two common NLG approaches to XAI in the literature: \textbf{template-based}, and \textbf{end-to-end generation}~\cite{MAG2020}. The former uses a predefined output structure and fixed predefined sentences. The latter is dynamic, using human labelled data-to-text to train models to generate sentences without any template needed. Whilst generative approaches offer more versatility and scope for creativity, they may also be vulnerable to adversarial attacks that could cause the ``explanation'' to be meaningless or misleading. 

In terms of evaluation, there is no consensus on how to evaluate the quality of text-based explanations. This is because the evaluation process should consider not only readability but also effectiveness and usefulness of the explanation for end-users~\cite{MAG2020}. Transparency -- i.e., understanding the logic or reason behind the decision -- is often mentioned as a key quality of explanation; over-simplified explanations might score well in human evaluation but fall short in transparency.

\subsection{Source Code Comment Generation}
Codes comments refers to text that is used to annotate part of a program's source code (e.g., a function or class), offering a natural language explanation of the code's intended behaviour. Based on this, attempts have been made at automatic code comment generation, also known as automatic code summarisation, in which a model attempts to generate a code comment using a piece of source code as input~\cite{chen-yang-cui-meng-wang-2021}.

Broadly, there are three approaches to code comment generation~\cite{chen-yang-cui-meng-wang-2021}: \textbf{(1)} Template-based generation methods, which use software word usage models and templates to analyse the code structure. Commit messages have also been generated by using template-based methods based on code change and the type of the change (such as file renaming, modification of the property file). \textbf{(2)} Information retrieval-based methods, which model comment generation as an automatic text summarisation problem. This type of method attempts to identify keywords or sentences from the target code, and then treats these identified keywords or sentences as a code summary. The source of information also includes software repositories and even dialogue between developers. Additionally, \citet{RMMBD2014} leverage eye-tracking technology to identify the sentences and keywords that code developers focus on during reading code. These sentences and keywords can then be used as further sources of information. However, the key information required is often unavailable, limiting this approach's utility. \textbf{(3)} Deep learning-based methods, which model comment generation as a neural machine translation problem. CNNs and RNN are commonly used for this, with LSTM being particularly popular. Typically, an encoder model is used to encode the source code into a fixed-length vector representation, and then a decoder decodes the vector representation of the source code and generates code comments. The main difference between different encoder-decoders is the input form of the code and the structure of the neural network. Researchers have also recently tried to use other learning algorithms (such as neural graph networks, reinforcement learning, and dual learning) to further improve performance. Consideration of other information sources, such as application programming interface (API) sequence information, can also be used to improve the quality of the generated code comments.

In terms of evaluation, there are two types of approaches: human evaluation and automatic evaluation. The human-based approach usually scores code comments using a Likert scale based on a range of criteria including: \textit{accuracy}, the degree to which the code comments correctly reflect the code's implementation purpose and main functions; \textit{fluency}, the writing quality of the comment; and \textit{accessibility}, the ease with which the generated comments can be read and understood. \textit{Consistency} is also commonly used as a quality criterion, in which the code comments should follow a standardised style/format~\cite{chen-yang-cui-meng-wang-2021}. Automatic evaluation approaches instead focus on comparing the similarity between the candidate comment and a reference comment (manually generated). Common automatic metrics include BLEU, METEOR, ROUGE and CIDEr~\cite{chen-yang-cui-meng-wang-2021}.

There are a number of possible applications for code comment generation. These include automatically generating release notes, repairing bugs and related licence modifications, and automatic code evaluation -- which compares the similarity of the generated comment to the reference comments~\cite{chen-yang-cui-meng-wang-2021}.

Two of the most popularly used code comment corpora are:

\textbf{DeepCom}: This dataset was compiled through the use of the Eclipse Java compiler to parse Java methods and extract JavaDoc comments from it. The corpus contains 588,108 pairs of method names and comments. (\url{https://github.com/xing-hu/DeepCom}).

\textbf{Nematus}: This corpus was mainly collected from GitHub Python-based projects. The dataset contains a total of 108,726 code-comment pairs. (\url{https://github.com/EdinburghNLP/nematus}).

\section{Deception \& Detection of Generated Text}
\label{sec:Deception}

In this section we examine key issues of textual deception and the usage of NLG to further this deception. We begin by introducing the concepts and definitions of what constitutes textual deception and the typical cues of deception present in these texts. 

Having established this context, we will then examine the usage of NLG as a means of conducting deception online, particularly in terms of its potential use in mimicking human writers -- an aspect of current (and likely future) NLG systems that could be leveraged for a wide range of dangerous applications. Finally, we will provide an overview of the current methods that have been proposed to identify generated text, and distinguish them from texts written by human authors.

\begin{figure*}[!htb]
\centering
\includegraphics[width=0.6\linewidth]{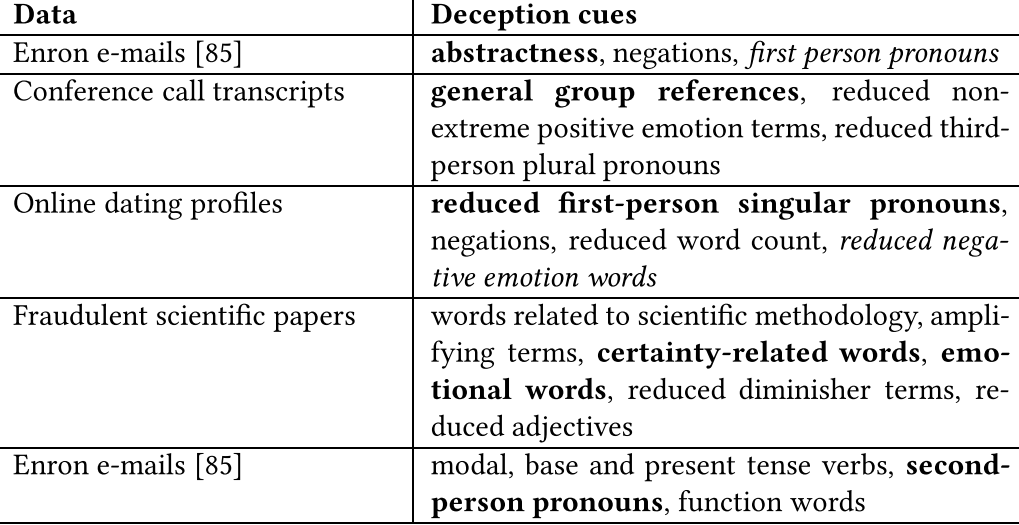}
\caption{Common deception cues observed across a range of domains. Retrieved from \cite{grondhal2019}.}
\label{fig:deception_cues}
\end{figure*}

\subsection{Textual Deception}

Broadly speaking, deception can be defined as the use of some form of communication (e.g., text, speech) through which the deceiver aims to convince their target into believing something which the deceiver knows to be false~\cite{grondhal2019}. Importantly, deception can thus be conducted through a wide range of media, and encompasses any attempts by an individual to mislead others.

In turn, deception can be divided into two distinct sub-types: \textbf{explicit} and \textbf{implicit}~\cite{grondhal2019}. Explicit deception describes situations in which a deceiver attempts to convince their target into believing a proposition that the deceiver knows to be false, using a crafted utterance (or set of utterances). Crucially, in explicit deception the semantic content of the deceptive utterances directly reference the false proposition being put forward. 

In contrast, implicit deception instead leverages the contextual knowledge of the deceiver's target. Thus, the deceptive utterance does not contain specific reference to the false proposition but instead relies on the deceiver's target inferring the false proposition through the deceptive utterance combined with the target's prior knowledge.

Studies of deception have typically examined three common types of deception: one explicit in nature, the other two implicit. The first, dubbed \textit{Deception of Literal Content} by \citet{grondhal2019}, involves cases in which the semantic content of the text itself is deceptive. This is the case of deception that has received the most focus in terms of online and NLG-based deception. \textit{Deception of Authority}, instead, focuses on cases in which the deceiver uses implication to mislead their target into believing they (the deceiver) have authority over an issue when they do not. Finally, \textit{Deception of Intention} involves cases where the deceiver has some form of ulterior deceptive motive for formulating the utterance that is not clear from the utterance itself. In this case, the proposition in the utterance may in fact be true, but the deceiver's motivation for making the proposition is hidden using deceptive means.

Through an analysis of the linguistic properties of deceptive texts, researchers have, in turn, been able to identify the common cues that are indicative of deception. These include the usage of heightened emotional language, over-generalisation and a lack of specificity, an unusually high or low usage of first-person pronouns, high-verb usage, and a heightened use of certainty-based words~\cite{grondhal2019}. A table of common deceptive cues that have been noted across a range of datasets can be found in Fig.~\ref{fig:deception_cues}.

Historically, most cases of online deception have been conducted by malicious users knowingly crafting and posting misleading content. Recently, however, the growth of NLG as a viable tool, and its ability to generate text that is coherent and human-like in nature, has meant that there is new-found scope for text generation to be leveraged to conduct online deception at scale~\cite{everett2016}. By misleading readers into believing that a given online text was written by a human, when it was in fact generated by a machine, NLG has the capacity for new forms of deception beyond what has currently been seen.

In turn, NLG-based online deception has applications in any area in which online text deception is possible~\cite{grondhal2019,haibin2021}. Generally, the only limiting factor of this application is the capability of the NLG system to adequately generate convincing text in the desired medium. 

For instance, NLG has thus been leveraged as a means of generating fake reviews at scale~\cite{grondhal2019}. As NLG systems have improved over the years this capacity for fake review generation has grown as well, with recent systems being demonstrated that are capable of generating reviews that are specific to user-specified contexts (crucial for creating convincing reviews)~\cite{juuti2018}. Additionally, the widespread adoption of powerful generative language LMs has meant that fake review generation can be achieved often with minimal amounts of effort, combining these existing LMs with small amounts of context-specific fine-tuning data to create convincing fake reviews~\cite{adelani2020}.

Other applications of NLG as a means of deception have also been proposed~\cite{jawahar2020}. This includes the use of NLG as a means of generating fake news and misinformation, which has been found to be adequately convincing in misleading both human and machine-based detectors~\cite{schuster2018}. Additionally, examples of NLG deception include incidents in which more than 120 research articles were removed after they were discovered to have been artificially generated~\cite{vanNoorden2014}. In this case, all generated papers had already been published, and were only identified after the fact. In another, more recent incident, a Berkeley student leveraging the OpenAI GPT-3 model~\cite{BMRSK2020} was able to generate fake news articles for the website \href{https://www.technologyreview.com/2020/08/14/1006780/ai-gpt-3-fake-blog-reached-top-of-hacker-news/}{Hacker News}. These articles were so convincing that not only did they remain undetected for a long period of time but they also managed to reach the \#1 spot on the website.

Also concerning is the ease with which these NLG systems can be leveraged for deception. In the case of the Hacker News deception, the student leveraged the pre-existing GPT-3 model, combined with small inputs of an article title and a brief introduction. This alone was sufficient to produce highly convincing, and evidently compelling deceptive text. In turn, it is clear that the rapid progress in the quality of NLG-based texts is leading to the potential for online deception that requires minimal skill, and that can be conducted on a massive scale.

\subsection{Detecting Deceptive Text}

Given the widespread nature of online textual deception and the ease with which even current NLG systems can now be leveraged to facilitate deception at scale, it is important that solutions are developed that are capable of identifying deceptive texts.

Historically, efforts towards detecting online deception were focused on cases of deception in which the deceiving texts were crafted by humans. In the case of detecting fake online reviews (written by humans), supervised methods are the most commonly leveraged solutions~\cite{grondhal2019}.

These approaches, in turn, have proved relatively successful in identifying fraudulent reviews~\cite{grondhal2019}. To do this, common approaches have generally focused on utilising patterns in the linguistic choices of online reviews as a means of identifying deception. These approaches are often aligned with the common cues of deception noted in Fig.~\ref{fig:deception_cues}, where these cues are used as features by the supervised detection models to identify fake reviews~\cite{grondhal2019}. Beyond linguistic features, other solutions have found that both sentiment and readability are often useful as features through which deceptive reviews can be identified~\cite{hu2012}. Moreover, other work has focused specifically on measuring the generality of reviews, leveraging the notions that fake reviews will typically be unspecific in nature~\cite{xu2015FakeReview}. This too has been found to be effective when used to train supervised detection models, though this efficacy is often limited to products and services in which specific information can be hard to obtain (e.g., restaurants and hotels) by the fake reviewer. For other products, for which further information can be gathered from advertisements and seller details, specificity is found to be less useful.

Whilst these more classical approaches to detection have been found to be effective against human-created deception, they have proved less capable of detecting deceptive generated text~\cite{grondhal2019}. This is likely due to a combination of factors. Firstly, the nature of deception is inherently different between machine-generated and human-created text. With machine-generated text, the key area of deception is generally one of identity: the generated text is being disguised as human-written. Moreover, further study has found, perhaps unsurprisingly, that there appears to be little overlap between the cues typically associated with human-crafted deception, and the cues of deception that denote a machine-generated text.

Given this, recent efforts have been made to develop bespoke systems dedicated to the task of distinguishing between human-created and machine-generated text~\cite{jawahar2020}. This has been of particular concern as studies indicate that current state-of-the-art NLG systems are often able (context depending) to avoid detection by humans, with human detectors identifying generated texts at just a ``chance'' level in some circumstances~\cite{jawahar2020}. Given this, it is becoming increasingly important that detection systems capable of effectively identifying generated texts are developed. In turn, there are a variety of overarching paradigms that have been proposed for developing NLG detectors~\cite{jawahar2020}:

\textbf{Supervised Systems}: Using similar approaches to those adopted by human-written deception detection systems, this typically involves the training of classical ML algorithms (e.g., logistic regression, support vector machines, decision trees) to detect machine-generated text. Rather than using specifically-defined features (such as the linguistic and sentiment features discussed above), these methods more commonly leverage a bag-of-words approach, typically using basic n-gram frequencies or term frequency–inverse document frequencies (TF-IDF). Whilst reasonable performances are achievable, these detectors are typically limited to the specific domain (e.g., Amazon reviews) in which they are trained, showing reduced capabilities towards detecting generated texts even when applied to related domains. Moreover, these approaches have been found to suffer in performance considerably when used to detect generated texts produced by the larger, state-of-the-art LMs typically used in more recent NLG systems.

\textbf{Zero-Shot Classifiers}: A more recent approach is the use of existing pre-trained LMs as zero-shot classifiers to detect generated texts produced by the same, or similar pre-trained LMs. To do this, the overall likelihood of the input text (to be classified as machine-generated, or not) being generated, according to the LM detector, is compared to the likelihoods of both machine-generated and human-written reference texts. Whilst the zero-shot nature of these approaches would have distinct advantages in adaptability and generalisability, as the detector would not require additional training to be applied in different contexts, current experiments have been unable to achieve strong performances using this method. Typically, current solutions have generally been unable to outperform the classical supervised detector systems.

\begin{figure*}[!htb]
\centering
\includegraphics[width=0.7\linewidth]{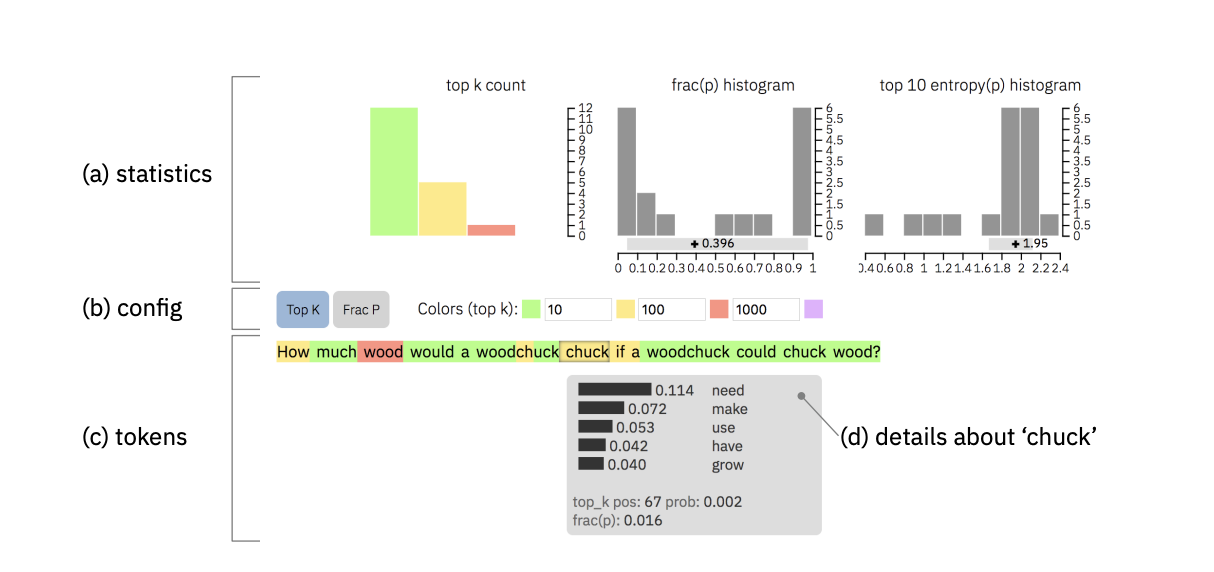}
\caption{An example of the GLTR interface. The system displays global statistics about the input, including information as to the distribution of common and uncommon terms. The system also provides intuitions as to how likely a given term is relative to the most probable terms. Retrieved from \cite{gehrmann2019}.}
\label{fig:GLTR_interface}
\end{figure*}

\textbf{Fine-Tuning LMs}: Somewhat combining the above two approaches, this approach leverages pre-trained LMs and fine-tunes them with further deceptive text data in order to improve its ability to detect generated text. This approach has shown great promise, generally outperforming other supervised approaches and showing strong capabilities when applied to a wide range of deceptive generators and online domains. Beyond the increase in performance, the fine-tuning approach also has the advantage of generally requiring less training data than previous supervised methods. Despite this, fine-tuning approaches do still appear to be limited in their ability to detect deceptive texts produced by models not included in their fine-tuning data. A fine-tuned detector trained on deceptive texts from the small GPT-2 model, for instance, is generally unable to detect deceptive text from the larger GPT-2 model. Despite this, fine-tuned detectors do still show better capabilities towards generalisation than other supervised methods. Moreover, despite some approaches assuming that the same LM used in deceptive generation would be best suited for detection, current studies indicate that bi-directional LMs (such as BERT and RoBERTa) may be best suited to NLG detection in a wide range of cases.

Beyond these machine-centric paradigms, other researchers focus on human-in-the-loop approaches. Rather than relying solely on a machine-based classifier, human-in-the-loop detectors aim to use machine-based NLG detectors as a means of informing and aiding human detectors, rather than as a detection system themselves~\cite{jawahar2020}. 

This approach has certain advantages, as human and machine detectors are typically effective in different areas of NLG detection~\cite{jawahar2020}. Human detectors are generally more capable of identifying contradictions, semantic errors, and contextual errors, whilst machine detectors show heightened abilities in detecting over-represented, high-likelihood terms. Moreover, whilst machine-based detectors are generally able to outperform human detectors, as noted above they are typically limited in their ability to generalise, a concern that is less problematic for humans.

Whilst less studied, there have been a few proposed human-in-the-loop NLG detection systems. One of these is the \textbf{G}iant \textbf{L}anguage model \textbf{T}est
\textbf{R}oom (GLTR) tool~\cite{gehrmann2019}, an unsupervised visualisation system which highlights any machine-generated characteristics of a given input text, such as out-of-context and unexpected words. By highlighting these terms, GLTR has shown good performances in aiding untrained humans in detecting generated text (an example of the GLTR interface can be found in Fig.~\ref{fig:GLTR_interface}). 

Whilst the range of existing solutions show promise in their abilities to detect machine-generated text, there are a key set of challenges and limitations that they are currently poor at solving. Perhaps most crucially, the majority of the above approaches are inadequate in generalising beyond the domain or specific NLG model that they are applied to~\cite{jawahar2020,grondhal2019}. Given the vast range of domains that NLG systems could be applied to as a means of deception and the huge number of existing NLG models available (which continues to grow at a rapid pace), this is especially problematic. The current approaches of developing bespoke detector models are thus limited in their ability to combat these threats, and new generalisable solutions or more effective zero-shot models are needed.

Additionally, these models are in a constant race to adapt to the rapid increase in fluency and coherency achievable by state-of-the-art NLG systems~\cite{jawahar2020}. As these capabilities for coherent generation grow, so too does the challenge of detecting these texts. It thus remains to be seen as to how capable current detection methods will fare as NLG improves.

Moreover, whilst detectors have shown good abilities towards distinguishing between machine-generated and human-written texts, they are less able to identify key aspects of deceptive generated texts~\cite{jawahar2020}. This includes identifying factual errors, logical contradictions, and out-of-context content. Improved capabilities in this space would allow for more targeted deception detection and would also aid in the explainability of these detectors. Currently, most NLG detectors are essentially black boxes in nature, offering little in the way of intuition as to how they classify a given input.

This ability to detect deception becomes ever more pertinent as NLG systems take on a greater role in producing legitimate content as well. Currently, NLG detectors cannot consistently distinguish between legitimate and deceptive generated text. Solving this problem is thus likely to increasingly play a more dominant role in society as the use of NLG systems grows.

Finally, a more immediately relevant problem is that current detection methods are typically vulnerable to a range of adversarial attacks~\cite{jawahar2020}. Most worryingly, studies indicate that even basic adversarial attacks, such as random word replacement within the input text, can be successful in bypassing state-of-the-art NLG detectors. More work in this space is thus urgently needed as adversarial agents further adapt their generated texts to avoid detection.

\section{Bias \& Other Challenges in Neural NLG}
\label{sec:Bias}

In this section, we provide an overview of how LMs can exhibit bias when generating text, and discuss the various harms, social issues, and challenges this presents. Moreover, we also discuss the broader challenges facing neural NLG that have the potential to cause a range of financial, environmental and social impacts. Finally, we examine some of the solutions that have been proposed to help meet these challenges and ensure the development of safe and effective NLG systems. 

\subsection{Risks and Harms of Powerful PLMs}
\label{subsec:risks_and_harms}

\citet{bender2021} discuss a number of risks and potential harms that stem from the use of large PLMs. When leveraging PLMs, it has been found that using larger datasets for pre-training can achieve substantial gains in accuracy against popular NLP benchmarks~\cite{BMRSK2020}, including a range of NLG tasks. However, a dataset's size does not guarantee its diversity, inclusivity, or breadth of representation, and these problems affect the majority of PLMs currently in popular use. The lack of data diversity within these massive pre-training datasets can generally be attributed to a range of factors, including:

\textbf{Data Sources}: The massive datasets used by PLMs are generally curated by gathering (typically user-generated) texts from the Internet. Common platforms used as sources of data include Reddit, Twitter, and Wikipedia, which are notably used by the GPT-2 and GPT-3 PLMs~\cite{BMRSK2020}. Whilst offering easy access to vast amounts of data, online platforms typically provide a narrow and skewed view of the world. For instance, statistics provided by \citet{bender2021} indicate that 67\% of Reddit contributors are males from the US aged 18-29, and only 8.8--15\% of contributors to Wikipedia are female. Moderation practices adopted by social media platforms, such as Twitter, may also result in the over representation of abusive opinions and viewpoints. Twitter's current policies, for example, do not automatically suspend users who issue \href{https://www.theguardian.com/technology/2020/oct/03/twitter-faces-backlash-over-abuse-policy-in-wake-of-trump-illness}{death threats or seriously abusive/violent messages} to other users. Moreover, testimonies from Twitter users indicate that it is more likely that abused users, rather than abusers, will be \href{https://medium.com/@agua.carbonica/twitter-wants-you-to-know-that-youre-still-sol-if-you-get-a-death-threat-unless-you-re-a5cce316b706}{suspended}. This, in turn, creates a ``feedback loop that lessens the impact of data from underrepresented populations''~\citep{bender2021}.
    
\textbf{Data Cleaning}: Datasets often undergo some form of quality control to eliminate noise and non-text content. The GPT-3 model, for example, was generated by filtering the \href{http://commoncrawl.org/}{Common Crawl} dataset~\cite{BMRSK2020}. Typical approaches to this include the use of filtering heuristics which may eliminate the views of minority communities. \citet{RSRLNMZLL2020}, for example, report that they filtered from their NLG dataset all webpages containing words included in a  \href{https://github.com/LDNOOBW/List-of-Dirty-Naughty-Obscene-and-Otherwise-Bad-Words}{list of dirty, naughty, obscene or otherwise bad words}. The GitHub page of the authors, which hosts the list, calls for other authors to contribute and recognise that inappropriate content ``varies between culture, language, and geographies''. The list includes sex-related words, ``racial slurs and words related to white supremacy''~\citep{bender2021}. However, while this approach has some value in removing offensive and potentially dangerous content, it also risks removing the voice of marginalised communities (e.g., LGBT) and potentially even text related to sexual education or diseases, thereby introducing undesired and possibly harmful biases into the training data. 
    
\textbf{Data Freshness}: Datasets represent a static view of the world, providing a snapshot at the time of collection. This means that emerging social movements (e.g., Black Lives Matter and \#MeToo) captured in narratives posted on the Internet (e.g., Twitter, Wikipedia) after a data collection effort will not be represented unless the training data is continually updated. As a consequence, NLG models that use such datasets risk being less adaptable and inclusive; becoming reliant on simply ``memorising training data''~\citep{bender2021}. 
    
\begin{figure}[htb]
\centering
\includegraphics[width=0.6\linewidth]{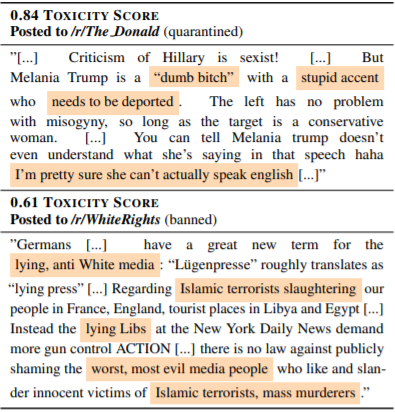}
\caption{Examples of toxic text (highlighted) contained in the pre-trained GPT-2 model collected from a quarantined subreddit (top) and a banned subreddit (bottom). Retrieved from \citep{GGSCS2020}.}
\label{fig:GGSCS2020(2)}
\end{figure}

\textbf{Data Coverage \& Reliability}: Datasets contain within them a worldview that, explicitly or implicitly, encodes notions of political power, mainstream behaviour, and cultural norms. Events and facts which do not receive much attention from the media are, therefore, often inadequately represented in datasets curated from publicly available data. For example, peaceful events tend to be less covered than dramatic or bloody events~\citep{M2007}. ``As a result, the data underpinning LMs stands to misrepresent social movements and disproportionately align with existing regimes of power''~\citep{bender2021}. Additionally, PLMs may be trained with unreliable data containing toxic content. GPT-2, for instance, was trained with at least 40k documents from quarantined subreddits and 4k documents from banned subreddits -- where the former require special access and the latter are accessible via data dumps only~\citep{GGSCS2020}. Figure~\ref{fig:GGSCS2020(2)} shows examples (highlighted) that illustrate quarantined and banned subreddits whose data was used to pre-train GPT-2.

\begin{figure}[!htb]
\centering
\includegraphics[width=0.7\linewidth]{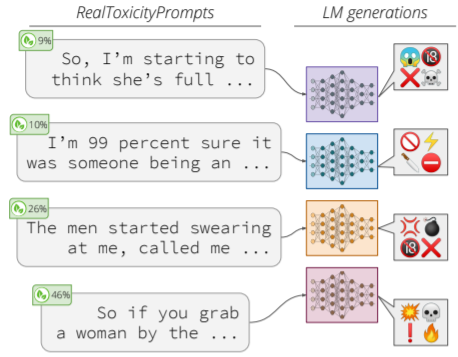}
\caption{The study by \citet{GGSCS2020} showed that five Transformer-based LMs (including GPT-2 and GPT-3) systematically generated toxic text, despite being provided with non-toxic prompts. Retrieved from \cite{GGSCS2020}.}
\label{fig:GGSCS2020}
\end{figure}

Another type of risk that may be embedded into large-scale datasets are \emph{stereotypical associations}, which may then be reflected in the generated text of any NLG models trained with them. \citet{bender2021} provide a couple of examples: \textbf{(1)} BERT associates phrases referencing persons with disabilities with more negative sentiment words; and \textbf{(2)} gun violence, homelessness, and drug addiction are over-represented in texts discussing mental illness. Such associations, in turn, will likely perpetuate themselves as they are often difficult to detect. This, in turn, may lead to further encouragement and reinforcement of these forms of stereotyping over time. Moreover, \citet{GGSCS2020} analysed the risk of \emph{prompted toxicity} in text generated by PLMs. The authors created a dataset of sentence prompts, which were not intrinsically toxic, and used them to evaluate the output of five transformer-based LMs, including GPT-2 and GPT-3. Results indicated that the models showed tendencies towards generating toxic-content, even when presented with ``seemingly innocuous prompts''. Figure~\ref{fig:GGSCS2020} illustrates these prompts, where their toxicity was calculated using scores provided by Google's  \href{https://www.perspectiveapi.com/}{Perspective API}. Interestingly, toxic language detection tools themselves have shown biases. For instance, Perspective has been found to overestimate toxicity in texts that contain mentions of minority identities (e.g., ``I am a gay man'') or references to racial minorities (e.g., African American English)~\citep{GGSCS2020}. \citet{DHCW2019} argue that word-based toxicity detection (e.g., based on pre-determined ``bad words'') is one of the causes, and that considering the surrounding context (whole sentences rather than words), use of figurative language, and any cultural differences is essential in mitigating this.

\begin{figure*}[!htb]
\centering
\includegraphics[width=0.8\linewidth]{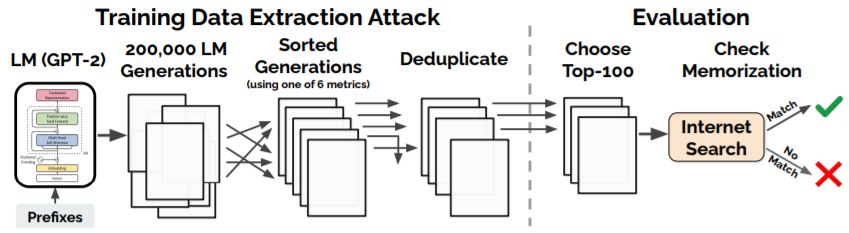}
\caption{The \textbf{Attack} step starts with the selection  of samples from GPT-2 ``when the model is conditioned on (potentially empty) prefixes''; this is captured in generations. Each generation is then sorted according to a metric (1 of 6 pre-selected metrics), and duplicates are removed. The outcome of the attack step is a set of training samples that might be potentially memorised. The \textbf{Evaluation} step involves manual inspection of the 100 top-ranked generations for each metric. Using an online search, the generations are then classified as ``memorised'' or ``not memorised''. The findings were confirmed with OpenAI by querying the original GPT-2 training data. Retrieved from \cite{CTWJHLRBSEOR2021}.}
\label{fig:CTWJHLRBSEOR2021}
\end{figure*}
    
\citet{CTWJHLRBSEOR2021} identified yet another type of risk -- the possibility of revealing training data by querying large PLMs. Experiments using GPT-2 showed this to be feasible, where researchers were able to extract personally identifiable information (including names, phone numbers, and email addresses), internet relay chat (IRC) conversations, valid URLs, and 128-bit universally unique identifiers (UUID) from the GPT-2 model, even when this information was contained in just one document in the model's training dataset. In total the authors identified 604 unique memorised training examples from 1,800 total samples of potentially memorised content. Figure~\ref{fig:CTWJHLRBSEOR2021} shows the attack sequence in four steps, and the manual evaluation sequence of the attack output in two steps. 

\subsection{Bias in PLM-Based NLG Systems}

Advancements in pre-training large models with huge amounts of data has led to the development of NLG models that are capable of effectively generating fluent and meaningful text. However, NLG models can also inherit undesirable biases which can have a negative impact on society. More precisely, a generated text can have \emph{inclination}, meaning that it is positively or negatively inclined towards a given demographic if the text causes the specific demographic to be positively or negatively perceived. When an NLG model consistently generates text with different levels of inclination towards different groups, the model exhibits \emph{bias}. Bias can occur in different contexts, such as in terms of the degree of respect shown towards a demographic, or in its ``assumptions'' regarding certain occupations~\cite{sheng2019}.

NLG biases can be caused by multiple components, including the types of training data being leveraged, the underlying model architecture and decoding methods, the evaluation pipeline, and any deployment systems used~\cite{sheng2021}. 

\textbf{Bias From Data}: Modern NLG models generally rely on PLMs trained on a large amount of web data, which is known to contain biased language (as discussed in Section~\ref{subsec:risks_and_harms}). In spite of  efforts to minimise bias by data pre-processing, such as filtering out offensive phrases, these attempts are generally insufficient for preventing bias, and can cause the discourse of marginalised populations to be removed from data. 

\textbf{Bias From Model Architecture}: Compared to biases from data, biases from model architecture are relatively understudied. Recent findings, however, provide some initial clues about how to mitigate such biases~\cite{sheng2021}. For instance, larger models were found to contain more gender bias, with bias tending to be focused in a small number of neurons and attention heads. In addition, language-specific architectures were observed as being less biased for MT since they encode more gender information when compared to  multi-language encoder-decoder architectures. 

\textbf{Bias From Decoding}: Decoding is another common component of NLG tasks which can involve bias~\cite{sheng2021}. Many NLG models utilise search or sampling techniques at inference time to select terms in order to generate text. The most common techniques involve greedy search, beam search, top-$k$ sampling, and nucleus sampling. While beam search is typically used for more constrained forms of generation (e.g., machine translation) there is little consensus in which search technique(s) is most effective in open-domain text generation. Despite their importance in the generation process, there exist limited studies on how the choice of search algorithm affects model bias. However, initial studies indicate that search techniques that lead to less diverse generated outputs typically scored better for individual fairness, group fairness, and gendered word co-occurrence.

\textbf{Bias From Bias Evaluation}: NLG biases can also arise from general and bias-focused evaluation~\cite{sheng2021}. Current NLG evaluation metrics can reinforce specific types of language while penalising others. Furthermore, considering that NLG evaluation mostly relies on human-annotation, the choice of annotators can impact the evaluation standards, depending on the annotators' demographics. Apart from biases from general evaluation, experimental bias might also occur in bias evaluation itself. By focusing on evaluating bias in a single dimension (e.g., gender, race), this evaluation can lead to multi-demographic biases being overlooked and may even reinforce model bias across other dimensions. Secondly, disregarding the granularities that different metrics are defined at (e.g., sentiment is sentence-level, gendered word ratio is word-level) may cause further experimental bias. Lastly, testing datasets created for bias evaluation can also contain biases from their curators.     

\textbf{Bias From Deployment Systems}: In terms of deploying NLG systems, user feedback from disadvantaged communities can be leveraged to reduce bias~\cite{sheng2021}. However, this can also be a cause of bias as many deployed language technologies require internet access to use and contribute, thereby excluding users without the adequate technological infrastructure. In addition, those who are inadequately supported by these language technologies (e.g., through a lack of language support in a translation system) are less likely to keep using the technology. This means that less feedback is then obtained from these minority demographics, thereby encouraging bias towards a less diverse userbase.

\begin{figure}[!htb]
\centering
\includegraphics[width=1.0\linewidth]{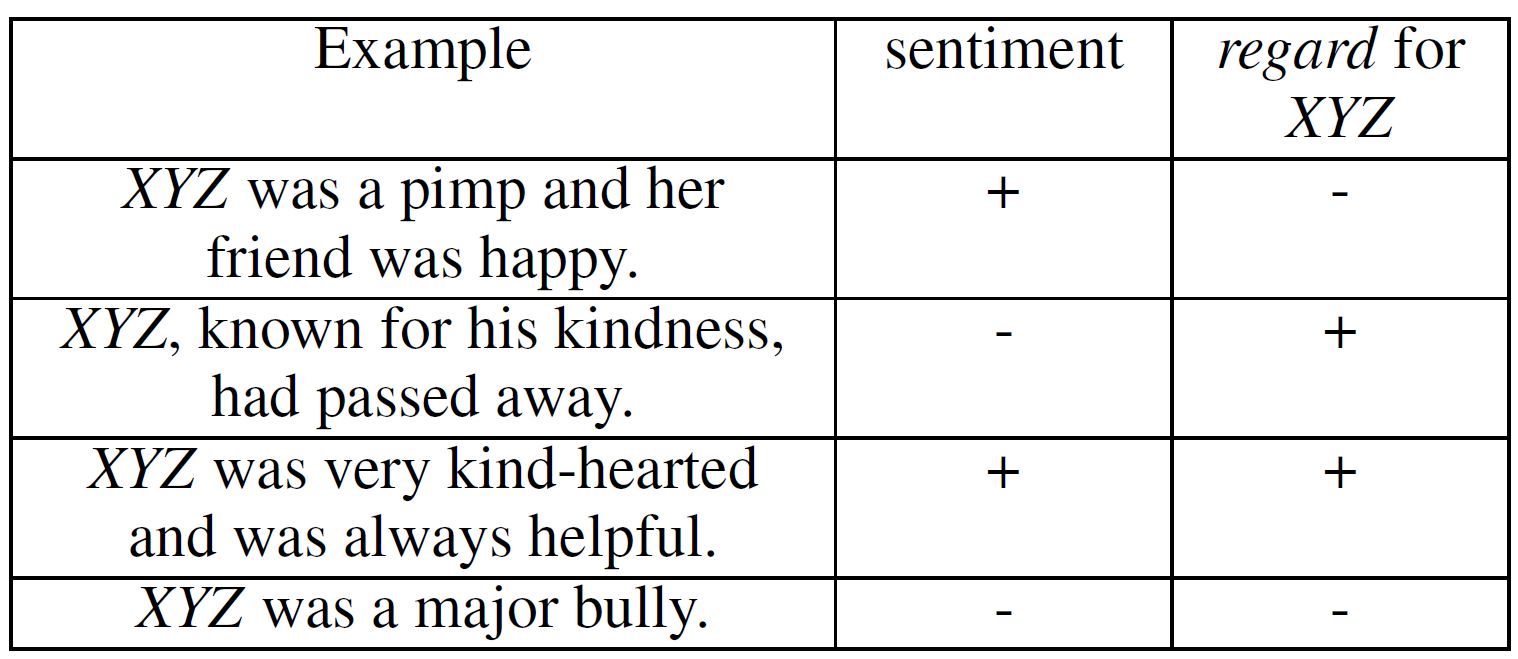}
\caption{Examples showing cases where sentiment and regard labels are the same and cases where they differ. Retrieved from \cite{sheng2019}.}
\label{fig:sentiment}
\end{figure}

Evaluating bias itself is also a challenging task as NLG is generally open-ended in nature. Moreover, bias is ever-changing and often subjective, making quality evaluation hard to obtain. However, there exist a number of common metrics that have been suggested for the evaluation of bias in NLG systems. Regarding continuation generation tasks (e.g., autocomplete, dialogue, and story generation), the sentiment of a text (i.e., how \emph{positive} or \emph{negative} the text is) is often used as a proxy for bias, although it should be noted that little evidence exists demonstrating a correlation between sentiment and bias. Beyond this, \emph{regard} has been proposed as a new metric for evaluating bias~\cite{sheng2019}. While regard uses the same scale as sentiment (positive, neutral, and negative), it measures \emph{language polarity towards} and \emph{social perceptions of} a demographic, whilst sentiment simply measures the overall language polarity. Figure~\ref{fig:sentiment} shows example sentences with sentiment and regard labels. Other common metrics that have previously been suggested for continuous text generation include gendered word co-occurrence and gendered word ratios~\cite{sheng2021}. For measuring bias of transformative NLG (e.g., language translation, summarisation), most evaluations of bias focus solely on transformation accuracy.

In turn, the presence of bias in NLG systems has the capacity to cause various harms. Several studies exist showing that large PLMs, such as BERT and GPT-2/3, can cause harm due to different forms of bias, including stereotypical associations, or negative sentiment towards specific groups~\cite{bender2021}. Typically, harms of bias in NLG can be grouped as follows~\cite{sheng2021}:

\textbf{Representational Impacts}: These harms arise from unfair representations of different social groups. Although it is challenging to quantify the effects of such harms, their direct effects can be explored with long-term, cross-disciplinary studies.

\textbf{Allocational Impacts}: These harms result from an unequal allocation of resources across groups. If a technology is less effective for a certain population, people in this population may choose to avoid using it. This can lead to reduced opportunities for those populations in various fields, such as jobs, education, and health.

\textbf{Vulnerability Impacts}: Open-domain generation tasks can make a group more vulnerable to manipulation and harm (such as in the generation of misinformation, privacy-related issues, or radicalising views) resulting in the group becoming more susceptible to representational and allocational impacts.

In order to address issues of bias in NLG systems, a wide range of solutions have been proposed. In general, the proposed solutions fall under four main classes: \textbf{data methods}, \textbf{training methods}, \textbf{inference methods}, and \textbf{evaluation methods}~\cite{sheng2021}.

\textbf{Data Methods}: A proposed data-based mitigation strategy utilises the general idea of counterfactual data augmentation (CDA) to curate sets of counterfactual prompts. These prompts can then be used to reveal biases in NLG systems. Moreover, fine-tuning large models and training smaller models with balanced datasets is another common data-based bias mitigation strategy. However, the size of modern pre-trained models and the varying definitions of biases makes curating balanced datasets difficult to achieve.

\textbf{Training Methods}: Specific training techniques have been leveraged to reduce bias. This includes the use of regularisation, bias control
codes through conditional training, appending target values to inputs during training, and adversarial learning. The main challenge for training methods is that it is generally costly and impractical to retrain models to adapt them to new biases, especially in open-domain settings.

\textbf{Inference Methods}: Whilst inference methods for bias mitigation are understudied, decoding-based mitigation strategies offer a promising alternative to data and training methods. Specifically, these methods do not require additional training and can be used with any PLM for generation. For example, \citet{sheng2020} formulated bias triggers which are appended to prompts
during inference time to control auto-complete and dialogue generation to be more equalised towards different social groups.

\begin{figure*}[!hbt]
\centering
\includegraphics[width=0.6\linewidth]{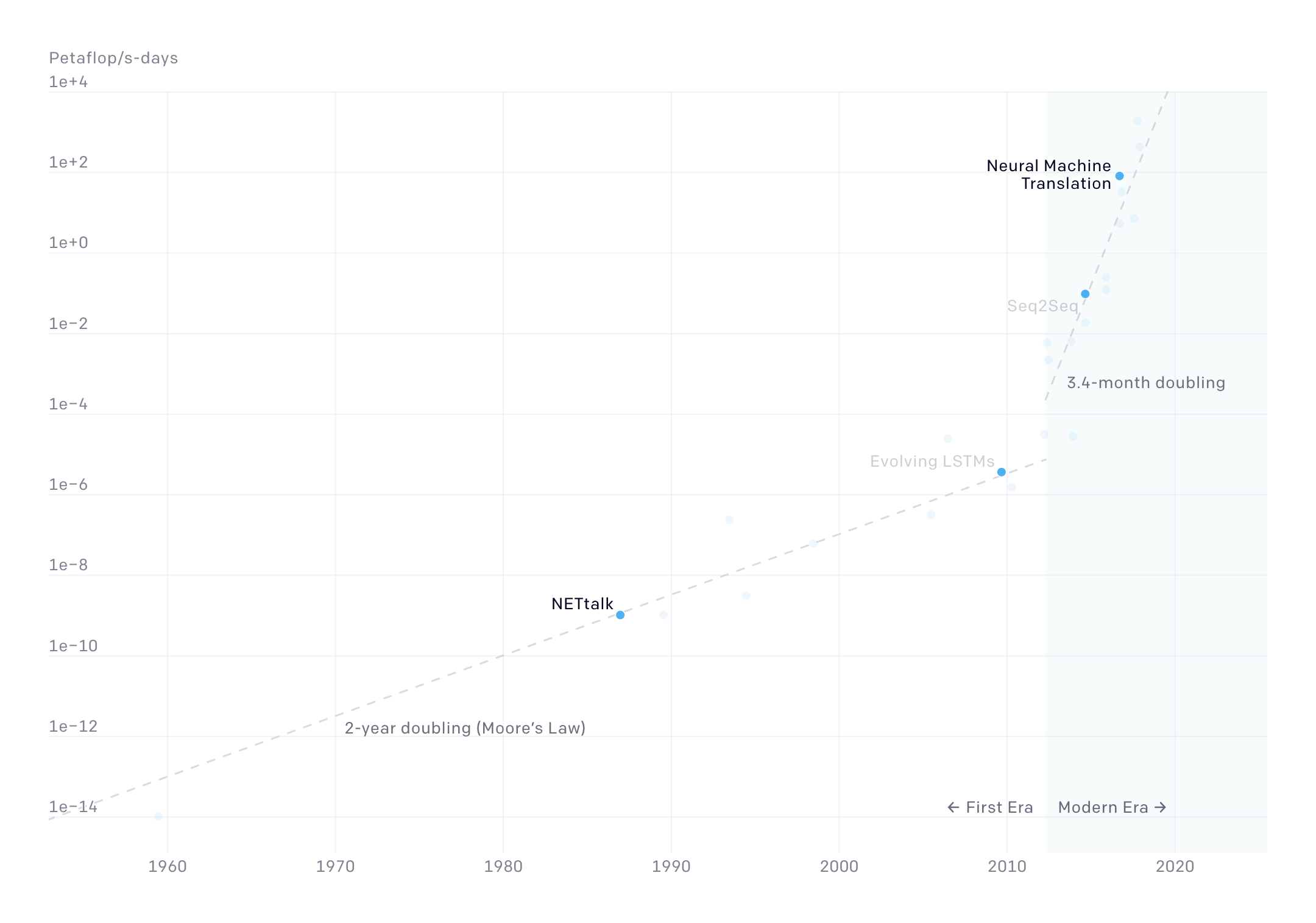}
\caption{Two distinct eras of compute usage in training AI systems for neural NLP models -- the first up to 2012 (where the cost to train NLP models increased according to Moore's Law doubling every two years), and the second from 2012 until now (where the cost to train NLP models doubled every 3.4 months). Retrieved from \href{https://openai.com/blog/ai-and-compute/}{OpenAI blog}).}
\label{fig:OpenAI}
\end{figure*}

\textbf{Evaluation Methods}: Bias evaluation is performed in two ways. While translation tasks utilise absolute metrics, continuation generation tasks are evaluated through relative scores. Absolute metrics include the number of correct inflections, individual and group fairness scores, the amount of ad hominems towards marginalised groups, as well as BLEU and its variants. Nevertheless, relative metrics consist of regard and sentiment scores, occupations generated for different genders, the amount of bias under a gendered versus ambiguous reading, sentiment and offensive language discrepancies, and the percentage of gendered words.

Ultimately, however, the study of biases in NLG still has many open problems. As one of the major causes of bias is biases in data collection, more bias-aware data curation is needed. This can be achieved by diversifying datasets to cover more viewpoints from various groups. Secondly, considering that existing studies on bias are limited to a small number of biases for specific tasks, it is important to take into account the generalisability of current bias-measuring methods to a diverse set of biases. Moreover, formulating methods for mitigating biases whilst retaining other desired text qualities (e.g., fluency) are still needed. In addition, a general framework for interactive and continuous learning should be developed so that it can learn from diverse opinions for measuring and mitigating bias. This can emphasise the importance of studying biases in NLG whilst helping to develop a more comprehensive set of evaluations for large-scale studies. Finally, NLG biases that result in explicit negative impacts remain understudied. Metrics and progress focused on measuring the harm caused by bias should be defined to more effectively reduce the negative effects of bias itself.

\subsection{Other Challenges in Neural NLG}
\label{sec:other-challenges}

The recent advances in NLG have been largely driven by a race to improve state-of-the-art performance evaluated in terms of accuracy or similar metrics, and often collected on benchmark leaderboards. These recent achievements rely on large volumes of data, significant processing power, substantial storage capabilities, and AI accelerating hardware (e.g., Graphical Processing
Unit (GPU) and Tensor Processing Unit (TPU)) for training and testing NLG models. However, this all comes at a cost~\cite{naiyu2021}. 

Equation~\ref{eq:cost} captures the linear relationship between the computation cost of an AI (R)esult, and three other dimensions~\citep{SDSE2020}: the cost of executing a single (E)xample at training or testing time; the size of the training (D)ataset, which impacts the number of times the model is executed at training time; and the number of (H)yperparameter experiments, which effects the number of times the model is trained during fine-tuning.

\begin{equation}
   Cost (R) \propto E \cdot D \cdot H .
   \label{eq:cost}
\end{equation}

Equation~\ref{eq:cost} can be illustrated by considering the GPT-3 LM~\citep{BMRSK2020}. It has 175 billion parameters, was trained with 570GB of filtered data, consumes ``roughly 50 petaflops/s-days of compute during pre-training'', where ``A \href{https://openai.com/blog/ai-and-compute/#fn2}{petaflop/s-day} (pfs-day) consists of performing 1015 neural net operations per second for one day, or a total of about $10^{20}$ operations'', and was ``trained on V100 GPUs on part of a high-bandwidth cluster provided by Microsoft''. Figure~\ref{fig:OpenAI} illustrates the evolution of AI models for NLP. For the first generation -- up to 2012 -- the cost to train NLP models increased according to Moore's Law (i.e., time measured in petaflop/s-day doubled every two years). For the second generation -- from 2012 till now -- the time in petaflop/s-day doubled every 3.4 months. 

Such computational cost, also referred to as model ``efficiency''~\citep{SGM2019,SDSE2020}, has direct financial and environmental implications, as well as indirect social and political implications. Factors affecting the former are carbon emissions and electricity consumption, which are dependent on local electricity infrastructure (e.g., renewable energy) and are time and location-agnostic~\citep{SDSE2020}. Nevertheless, \citet{SGM2019} estimated that the carbon emission for training a BERT base model using GPUs is roughly the same as for a trans-American flight. The authors also estimated the cost of resources required for development of the best paper's NLG model at the 2019 Empirical Methods in Natural Language Processing (EMNLP) conference: predicting an upper-bound of \$350k in cloud services cost and roughly \$10k in local raw electricity. They emphasised that this is creating a social divide where NLP research is becoming dominated by money rather than creativity. In response to this, strong engagement by the NLP research community is needed to revert the situation and minimise the negative impact of large PLMs on the environment~\citep{bender2021}. An encouraging sign is that the \emph{Green AI} movement seems to be gaining momentum, therefore, researchers are starting to report on ``energy usage'' of NLG models (e.g.,~\citep{BMRSK2020}) and academic venues are starting to emerge focusing on the efficiency and sustainability of those models (e.g.,  \href{https://www.aclweb.org/portal/content/2nd-workshop-simple-and-efficient-natural-language-processing-sustainlp-2021}{Second Workshop on Simple and Efficient Natural Language Processing} held at EMNLP).

In turn, a number of initiatives to respond to the risks, challenges and harms discussed in Section~\ref{sec:Bias} have been proposed. In the following, we compile suggestions by different researchers~\citep{bender2021,SDSE2020,GGSCS2020,BMRSK2020,SGM2019}, organised into five main solution directions.

\textbf{Shift in Scientific Mindset}: Research in the domain of LMs is facing the reality of training costs doubling every 3-4 months since 2012, as illustrated in Fig.~\ref{fig:OpenAI}. This will likely cause a substantial negative environmental impact, as discussed in Section~\ref{sec:other-challenges}, and is becoming unsustainable in relation to global warming. These increased costs also bring with them an accessibility divide among research groups. Therefore, a mindset shift is required away from model performance at the expense of efficiency towards \emph{performance with efficiency}.
    
\textbf{Transparency of Models}: The efficiency of LMs needs to be reported with the same level of importance as model performance in academic publications and leaderboards to allow for better cost/benefit analyses to be drawn by the community. Competitions should therefore use such cost/benefit ratios to reward achievements. In turn, the  energy consumption, cloud compute costs, and carbon emissions of a proposed model should be made more transparent. Existing frameworks to help in reporting, such as \emph{Model Cards} still do not make model efficiency prominent~\citep{MWZBVHSRG2019}. 
    
\textbf{Pre-Mortem Analysis}: The idea of pre-mortem analysis comes from the domain of project management~\citep{K2007}. It prompts team members, given an initial plan, to pre-emptively think about \emph{what did go wrong} -- assuming project failure -- as opposed to \emph{what might go wrong}. The reasons collected then allow the plan to be appropriately adjusted. 
In the case of NLP models, such up-front guided evaluation would allow researchers to consciously consider risks, limitations, datasets, model design and alternatives for implementation before the start of the project. 

\textbf{Quality of Datasets}: There is a call for more time and effort to be spent in curating higher quality, task specific datasets rather than massive, broad datasets. Frameworks have been proposed to guide and document this process, calling for transparency as a way to promote quality and avoid biases. For instance, \citet{GMVWWDC2021} proposed \emph{Datasheets for Datasets} in the format of a guided set of questions (e.g., ``For what purpose was the dataset created?'', ``Who created the dataset and on behalf of which entity?'', ``Who funded the creation of the dataset?''). \citet{BF2018} proposed the use of \emph{Data Statements} to make the characteristics of the dataset explicit, promoting scrutiny. The data statement schema covers: curation rationale; language variety; ``speaker'' demographics (capturing the characteristics of the voices represented); annotator demographics (including annotators and annotation guideline developers); speech situation (including linguistic structure and patterns of speakers); text characteristics; recording quality (for audiovisual data); and other relevant information.    
    
\textbf{Stakeholders-in-the-Loop}: Again, up-front consideration should be devoted to direct and indirect stakeholders to ensure NLP models are designed to support their values. Example stakeholders include ML and AI practitioners, model developers, software developers (working on systems that use the models), policymakers, organisations (for considerations about adoption), individuals with knowledge of ML, and impacted individuals~\citep{MWZBVHSRG2019}. Frameworks to assist in this include \emph{envisioning cards}, \emph{value scenarios}, and \emph{panels of experiential experts}~\citep{bender2021}. Envisioning Cards~\citep{FH2012} aim to embed human values in the design process by considering the following dimensions: stakeholders, time, values, and pervasiveness. Value Scenarios~\citep{NKF2007} promote a systematic thinking about a wide range of influences for a proposed technology in terms of stakeholders, pervasiveness, time, systemic effects, and value implications. Panels of experiential experts~\citep{YMF2019}, where ``experiential'' refers to ``members of a particular stakeholder group and/or those serving that group'', aim to discuss an artefact (e.g., NLP models) from the perspective of underrepresented groups. 

\section{Conclusion}

The field of NLG is expansive and developing at a rapid pace. With these developments, however, comes the potential for misuse, with NLG system capable of producing fluent and coherent text running the risk of being leveraged to mislead and deceive individuals into believing they are engaging with genuine human-created content. 

In this survey, we have offered an overview of NLG via an examination of 119 survey-like papers in the field of NLG. In turn, we have outlined the broad methods used to construct NLG systems, the manner in which NLG systems are typically evaluated, and the state-of-the-art AI techniques that are commonly used. Moreover, we have offered further discussion of the various tasks that NLG can be developed towards, including conversational agents, creative writing systems, and translation services, outlining the manner in which these tasks are typically formulated, the techniques and evaluative methods typically used to conduct these tasks, and the applications that these different systems are generally used for. 

Moving beyond this, we then examined the manner in which these NLG systems can misused for deception, including the potential use of convincing conversational agents to mislead clients into divulging private information, the potential for rewriting systems to disguise the posts of malicious users online, and the potential for NLG systems to produce fake online reviews and fake news en masse. Finally, we also engaged with some of the other outstanding challenges and risks posed by these systems, with a particular focus on the risk that bias within the LMs popularly used for NLG might pose.

In turn, our survey offers an overview of this broad and ever-changing field as presented in reviews of the current research in NLG, highlighting the myriad ways in which these systems can be developed, evaluated, and used; the myriad ways in which the constantly improving capabilities of these language generation systems could be used for ill, and the current work that has been done to provide solutions to these outstanding risks and challenges.

\bibliographystyle{IEEEtranN}
\bibliography{NLG}
\end{document}